\documentclass{article}

\usepackage{arxiv}

\usepackage[utf8]{inputenc}
\usepackage[T1]{fontenc}
\usepackage{graphicx}
\usepackage{booktabs}
\usepackage{amsmath,amsfonts}
\usepackage{microtype}
\usepackage[numbers,sort&compress]{natbib}
\usepackage{url}
\usepackage[hidelinks]{hyperref}
\usepackage{cleveref}
\usepackage{authblk}

\usepackage{tikz}
\usepackage{enumitem}
\usepackage{multirow}
\usetikzlibrary{positioning}
\usepackage{color}
\usepackage{xcolor}
\usepackage{multirow}
\definecolor{covered}{HTML}{1a427a}
\definecolor{updated}{HTML}{3d7fd1}
\definecolor{silent}{HTML}{b3202f}
\definecolor{dropped}{HTML}{6b6a66}
\newcommand{\qcell}[3]{\textcolor{#1}{\textbf{\textsc{#2}} #3}}

\usepackage[most]{tcolorbox}
\usepackage{tabularx}
\usepackage{colortbl}
\definecolor{cbgrid}{HTML}{E1E0D9}
\definecolor{cbink}{HTML}{52514E}
\definecolor{cbhead}{HTML}{F5F5F2}
\definecolor{cbcodebg}{HTML}{EDEDEA}
\definecolor{cbaccent}{HTML}{1A427A}
\newtcolorbox{codebook}[2]{enhanced, breakable, colback=white, colframe=cbgrid, boxrule=0.6pt, arc=7pt,
  left=9pt, right=9pt, top=6pt, bottom=7pt, colbacktitle=white, coltitle=black, titlerule=0.5pt,
  fonttitle=\sffamily\bfseries\small, fontupper=\small, before skip=10pt, after skip=12pt,
  title={\textcolor{cbink}{$\rightarrow$}\enspace #1\hfill{\mdseries\ttfamily\footnotesize\textcolor{cbink}{#2}}}}
\newcommand{\cbsec}[2]{\par\penalty-200\medskip\noindent{\sffamily\bfseries\small\tcbox[on line, colback=cbaccent, colframe=cbaccent,
  coltext=white, arc=3pt, boxsep=0pt, left=3.5pt, right=3.5pt, top=1.5pt, bottom=1.5pt]{#1}\enspace #2}\par\nopagebreak[4]\smallskip\nopagebreak[4]}
\newcommand{\cbcode}[1]{\tcbox[on line, colback=cbcodebg, colframe=cbcodebg, boxrule=0pt, arc=2pt, boxsep=0pt,
  left=2pt, right=2pt, top=0.6pt, bottom=0.6pt]{\ttfamily\footnotesize #1}}
\newcommand{\cbth}[1]{{\sffamily\bfseries\footnotesize #1}}
\newenvironment{cbtable}[1]{\par\smallskip\noindent\footnotesize\arrayrulecolor{cbgrid}\setlength{\tabcolsep}{5pt}%
  \tabularx{\linewidth}{#1}\hline}{\endtabularx\par\smallskip}
\newenvironment{cblist}{\begin{itemize}[leftmargin=1.3em, itemsep=2pt, topsep=3pt, label={\textcolor{cbink}{\textbullet}}]}{\end{itemize}}
\usepackage{soul}
\definecolor{rd1}{HTML}{F8CFE0}\definecolor{rd2}{HTML}{CFE0F8}\definecolor{rd3}{HTML}{FBE0BF}\definecolor{rd4}{HTML}{DDD3F5}
\definecolor{rd5}{HTML}{CDEFD9}\definecolor{rd6}{HTML}{F3EDB8}\definecolor{rd7}{HTML}{BFE8E8}
\newcommand{\exhl}[2]{{\sethlcolor{#1}\hl{#2}}}
\newcommand{\extag}[1]{\textsuperscript{\sffamily\bfseries\tiny\textcolor{cbaccent}{#1}}}
\newcommand{\exchip}[1]{\fcolorbox{cbgrid}{#1}{\rule{0pt}{0.45em}\rule{0.45em}{0pt}}}
\newcommand{\exblock}[1]{\par\smallskip\noindent{\sffamily\bfseries\footnotesize #1}\par\nopagebreak[4]\vspace{2pt}}
\newcommand{\expassage}[1]{\noindent{\footnotesize #1}\par}
\newcommand{\exnote}[1]{\noindent{\footnotesize #1}\par}
\newcommand{\exincident}[2]{\noindent{\footnotesize\textbf{Incident, from #2:} #1}\par}
\newcommand{\exlegend}[1]{\par\smallskip\noindent{\footnotesize\sffamily Risk domains:\quad #1}\par\smallskip}
\newcommand{\cbexample}[2]{\par\medskip\noindent{\sffamily\bfseries\small #1}\par\nopagebreak[4]\noindent{\footnotesize\color{cbink}#2}\par\nopagebreak[4]}
\newenvironment{cbenum}{\begin{enumerate}[leftmargin=1.5em, itemsep=2pt, topsep=3pt, label={\textcolor{cbaccent}{\textbf{\arabic*.}}}]}{\end{enumerate}}

\title{Whose Risks Count? What 1,901 AI Incidents Reveal About Corporate Disclosures}
\renewcommand{\shorttitle}{Whose Risks Count?}
\renewcommand{\headeright}{Preprint}
\renewcommand{\undertitle}{Preprint}

\author[1,2,3]{Simeon Allmendinger\thanks{Corresponding author: \texttt{simeon.allmendinger@uni-bayreuth.de}}}
\author[4]{Edyta Paulina Bogucka}
\author[1,2,3]{Niklas K{\"u}hl}
\author[4,5,6]{Daniele Quercia}
\affil[1]{AI Responsibility Centre, University of Bayreuth, Germany}
\affil[2]{Fraunhofer Institute for Applied Information Technology FIT, Germany}
\affil[3]{FIM Research Center, Bayreuth, Germany}
\affil[4]{Nokia Bell Labs, Cambridge, United Kingdom}
\affil[5]{Politecnico di Torino, Turin, Italy}
\affil[6]{King's College London, London, United Kingdom}
\date{}

\begin{document}
\maketitle

\begin{abstract}
AI systems can harm people, yet the AI actors that provide or deploy them are rarely obliged to disclose these risks.
US securities regulation is an exception.
Companies registered with the Securities and Exchange Commission (SEC) must disclose their material risks, including those of their AI systems, in filings that carry liability for misstatement.
Materiality, however, is defined by what a reasonable investor would consider important, so harms to third parties enter disclosure only insofar as they become financial risks for the company.
We test whether AI harms do so by matching 9,001 AI incidents to 18,229 SEC filings of 328 AI actors by risk type.
Of the 1,901 incidents with filings before and after them, half belong to a risk type that the AI actor never disclosed, and this share varies with the kind of harm.
Filings disclose the risk types of 67\% of system failures but only 2\% of social system harms, and AI actors themselves report fewer than 2\% of incidents.
Securities disclosure thus records the risks that AI actors face rather than the harms that people subject to AI bear.
Public authorities that hold AI actors accountable for such harms could make them financially material.
We release DISC, the Disclosure--Incident Securities Corpus, to support this work.

\end{abstract}

\keywords{AI incidents, AI harms, algorithmic accountability, corporate transparency, AI governance, securities disclosure, double materiality}

\begin{figure}[t]
    \centering
    \includegraphics[width=\linewidth]{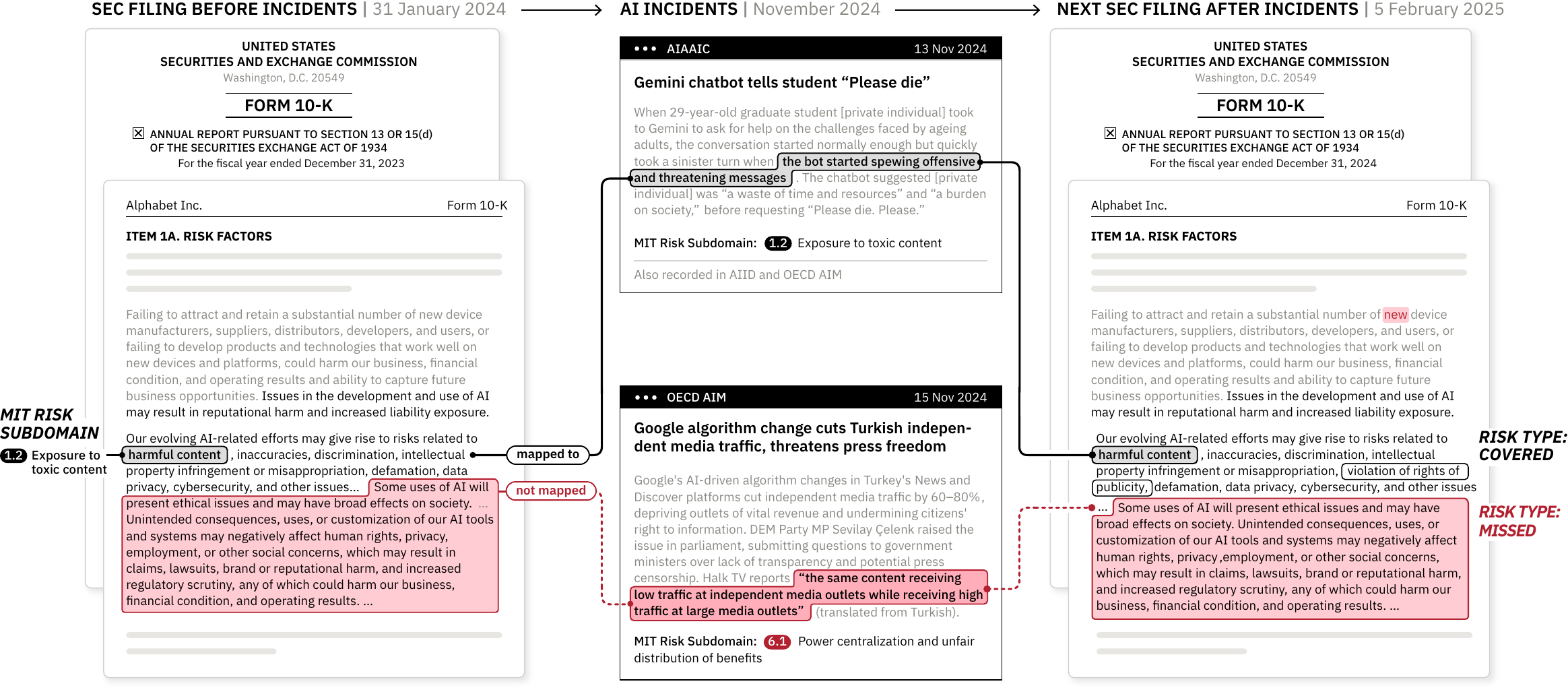}
    \caption{\textbf{Two AI incidents compared with SEC filings before and after they occurred.} Excerpts from the Risk Factors sections of Alphabet's SEC Form 10-K filings (left and right) are matched to incidents from November 2024 (middle) using the MIT AI Risk Repository subdomains. The Gemini chatbot incident matches a risk type disclosed in both filings (making it \textsc{Covered}), whereas the Turkish independent media incident has no matching risk type in either filing (making it \textsc{Missed}). We release these and other matches in our Disclosure–Incident Securities Corpus (DISC) available under \url{https://simeonallmendinger.github.io/disc/explorer/}.}
    \label{fig:cases}
\end{figure}

\section{Introduction}
\label{sec:intro}

Since 2016, public databases~\citep{AIID,AIAAIC,OECD_AIM} have recorded more than 9,000 incidents in which AI systems directly or indirectly caused harm~\citep{oecd2024incidents}.
Each such incident involves an \emph{AI subject}, a person, group, organisation or other entity affected by the use of an AI system~\citep{bogucka2026intersectionality}, and an \emph{AI actor}, an organisation that provides or deploys the system~\citep{oecd2019recommendation}.
Consider two incidents involving Alphabet in November 2024 (\Cref{fig:cases}).
In the first, Google's Gemini chatbot told a student to die.
In the second, changes to Google's search algorithms cut traffic to independent Turkish media outlets by 60--80\%, threatening their revenue and press freedom.
In both cases, Alphabet was the \emph{AI actor}, while the student and the affected media outlets were the \emph{AI subjects}.

AI subjects bear the harm from incidents, but AI actors are the ones who can prevent them.
Public disclosure can incentivise AI actors to reduce risks by making their failures visible, e.g., injuries at US mines fell after operators had to disclose their safety records in financial filings~\citep{christensen2017real}, and AI providers narrowed their systems' error gaps after an independent audit made those gaps public~\citep{raji2019actionable}.

Yet public disclosure of AI risks remains limited.
Under the EU AI Act, for instance, providers of high-risk AI systems must report serious incidents to market surveillance authorities, but not to the public~\citep{euaiact2024regulation}.
By contrast, US securities filings are one of the few public records of AI risks.
Under Item 105 of Regulation S-K, companies registered with the Securities and Exchange Commission (SEC) must disclose \emph{material risks}, including those arising from AI systems they provide or deploy~\citep{sec2020regulation,tsc1976industries,basic1988levinson}.
A risk is material if a reasonable investor would likely consider it important when deciding whether to buy, sell or hold a security~\citep{tsc1976industries,basic1988levinson}.
SEC disclosures are public, mandatory and periodic, and companies may face legal liability for material misstatements and misleading half-truths~\citep{basic1988levinson}.
Such disclosure of AI risks has surged, from 12\% of S\&P 500 companies naming AI as a material risk in their annual reports in 2023 to 72\% in 2025~\citep{conferenceboard2025ai}, while the SEC has declined to adopt AI-specific disclosure rules and relies on materiality instead~\citep{atkins2025iac}.

But risks to investors are not necessarily risks to AI subjects.
This distinction divides the debate over which risks companies should disclose.
One view, \emph{financial materiality}, holds that companies need to disclose only material risks, those that matter to investors, because harms to others eventually come back to the company through fines, lawsuits or lost revenue~\citep{freiberg2020how,jensen2002value}.
The other view, \emph{impact materiality}, holds that companies should also disclose risks to the people their activities affect, even when these risks never cost the company anything~\citep{ec2019guidelines,csrd2022directive}.
Requiring both is called \emph{double materiality}~\citep{ec2019guidelines}.
This difference matters when AI risks become actual incidents: do companies disclose these risks before they cause harm to AI subjects, or do some remain undisclosed?

Existing research has largely studied AI incidents~\citep{turri2023why,bogucka2026intersectionality} and AI risk disclosures, which are widespread but often generic~\citep{uberti2025seriously,jaworski2026ai}, separately. Recent studies trace incidents into later SEC filings~\citep{chincholikar2026ai} and market reactions~\citep{jiang2026artificial}. Yet no existing work examines whether AI actors disclosed the risk type of each incident before it occurred, or which risks never appeared in disclosure at all.

To address this gap, we collect 9,001 AI incidents from three public databases: the AI Incident Database (AIID)~\citep{AIID}, the AI, Algorithmic, and Automation Incidents and Controversies (AIAAIC) Repository~\citep{AIAAIC}, and the AI Incidents and Hazards Monitor (AIM)~\citep{OECD_AIM}.
We identify the AI actors involved in each incident, and collect 18,229 SEC filings from 328 of these actors. We then classify both incidents and disclosed risks using the 24 risk types of the MIT AI Risk Repository~\citep{slattery2024repository}.
This allows us to compare each incident with what its AI actors disclosed before and after the incident occurred.

For the 1,901 incidents with filings available both before and after they occurred, we make two comparisons.
First, we check whether the AI actor had already disclosed the incident's risk type before it occurred, which we refer to as ``previously disclosed''.
Second, we check whether the actor disclosed that risk type, or responded directly to the incident, in its next filing, which we refer to as ``disclosed''. The two comparisons follow the anticipation and responsiveness dimensions of responsible innovation~\citep{Stilgoe2013}.
These comparisons place each incident in one of four categories: \textsc{Covered}, \textsc{Eventually Disclosed}, \textsc{Dropped} or \textsc{Missed} (\Cref{tab:four-quadrants}).
This classification allows us to examine two main questions left open by previous research:
\begin{enumerate}[label=\textbf{RQ\arabic*}, ref=RQ\arabic*, leftmargin=*, labelsep=1em, itemsep=2pt]
    \item \label{rq1} \emph{Which AI incidents never appear in securities disclosures?}
    \item \label{rq2} \emph{Which AI incidents are previously disclosed in securities disclosures?}
\end{enumerate}

\noindent In answering these questions, we make three main contributions:
\begin{enumerate}[leftmargin=*, labelsep=1em, itemsep=2pt]
  \item \emph{Disclosure--Incident Securities Corpus (DISC).} We release the first corpus that matches AI incidents to securities disclosures. DISC labels filing passages with their risk types and the AI subjects they name as harmed, and places each incident in one of four quadrants: \textsc{Covered}, \textsc{Eventually Disclosed}, \textsc{Dropped} or \textsc{Missed} (\Cref{sec:method}).
  \item \emph{Evidence on what disclosure covers.} We show that securities disclosure covers the risk types of 67\% of system failures but of only 2\% of social system harms, and that half of matched AI incidents (50\%) are \textsc{Missed}.
  Most disclosed incidents are \textsc{Covered}: only 5\% of incidents are \textsc{Eventually Disclosed}, only 1\% are followed by an incident-specific revision, and fewer than 2\% are reported by the AI actor itself (\Cref{sec:results}).
  \item \emph{The empirical case for double materiality.}
  Disclosure covers most system failures, which harm the AI actor directly, and some discrimination and privacy incidents, for which AI subjects can hold the AI actor to account.
  Social system harms fall on everyone, so no single AI subject can hold the AI actor to account for them, and disclosure almost always omits them, both before and after the incident.
  Securities filings thus record the risks that AI actors face, and leave out the harms that AI subjects bear.
  Bringing those harms into the record requires double materiality, or public authorities that hold AI actors to account for them (\Cref{sec:discussion}).
\end{enumerate}

We use these findings to examine the limits of securities disclosure as a means of holding AI actors accountable.
We also identify directions for future research using DISC to trace the consequences of AI incidents.

\section{Related Work}
\label{sec:related}

Our work builds on three areas of research: AI incidents, AI risk disclosure, and materiality. Together, they show what harms AI systems cause, which risks companies disclose, and whether disclosure can hold AI actors accountable for those harms.
\smallskip

\noindent \textbf{AI Incident Research.}
An AI incident is an event in which an AI system directly or indirectly leads to harm~\citep{oecd2024incidents}.
Public databases document these incidents, but AI actors are not required to report them publicly, and the databases use different definitions of what counts as an incident~\citep{mcgregor2021preventing,turri2023why,sidhu2026open}. Research using these records examines who suffers the harm, finding that those harmed are often not the system's users but third parties, groups or society~\citep{li2025closer}, and that harms vary across identities such as age, race, gender and political identity~\citep{bogucka2026intersectionality}.
Other studies classify the types of harm, using frameworks such as the AI Risk Repository~\citep{slattery2024repository} and the taxonomies of \citet{raji2022fallacy}, \citet{shelby2023sociotechnical} and \citet{weidinger2023sociotechnical}.
\smallskip

\noindent \textbf{AI Risk Disclosure.}
Research on securities disclosure has long examined whether mandatory risk-factor disclosures reflect the risks companies actually face~\citep{campbell2014information}. Studies of cybersecurity found that companies' risk disclosures were associated with later breaches~\citep{wang2013association,li2018sec}.
However, this link weakened after the SEC's 2011 guidance led companies to disclose cybersecurity risks regardless of their exposure~\citep{li2018sec}.

For AI, research has focused less on whether risks were disclosed beforehand, and more on what companies disclose after incidents occur.
\citet{rao2025ai} compare model cards with recorded AI incidents, and find that developers and researchers pay little attention to fraud and manipulation. \citet{chincholikar2026ai} examine SEC filings following AI incidents and find that companies rarely mention specific incidents, and \citet{jiang2026artificial} find that stock prices react less to incidents whose AI risks companies had disclosed.
Other researchers propose that AI developers report safety-critical information, including incidents, to government~\citep{kolt2024responsible}, or that AI actors report incidents on SEC Form 8-K~\citep{strauss2025governing}.
Yet whether mandatory SEC filings disclosed the risk type of each AI incident before it occurred remains untested.
\smallskip

\noindent \textbf{Materiality and Accountability.}
Single materiality considers only \emph{financial materiality}, whereas double materiality adds \emph{impact materiality} (\Cref{sec:intro}).
Recent studies show how companies apply double materiality under the European Sustainability Reporting Standards, which cover 10 topics, including climate change, workers and consumers, but none specifically addresses AI~\citep{nahrstedt2026double}.

Yet disclosing a harm does not necessarily hold the AI actor accountable for it.
Accountability depends on whether those receiving the disclosure can impose consequences~\citep{wieringa2020what}, and transparency alone is not enough~\citep{ananny2018seeing}.
AI actors may also lack full knowledge of risks arising across the organisations that provide and deploy their systems, limiting what they can disclose~\citep{cobbe2023accountability}.
These studies show why some harms may remain outside disclosure or fail to prompt action.
\smallskip

\begin{table}[t]
  \caption{\textbf{Disclosure of incident risk types before and after AI incidents.} The 1,901 matched incidents are classified by whether their risk type was disclosed when the incident started (columns) and at the AI actor's next periodic filing (rows). Half of the incidents are Missed (50\%), nearly as many are Covered (44.5\%), and only 5.2\% are Eventually Disclosed.}
  \label{tab:four-quadrants}
  \begin{tabular}{lccr}
    \toprule
    & \multicolumn{2}{c}{At incident start} & \\
    \cmidrule(lr){2-3}
    At next filing & \textbf{Previously Disclosed} & \textbf{Not Previously Disclosed} & Total \\
    \midrule
    \textbf{Disclosed}   & \qcell{covered}{Covered}{845 (44.5\%)} & \qcell{updated}{Eventually Disclosed}{99 (5.2\%)} & 944 (49.7\%) \\
     \textbf{Not Disclosed} & \qcell{dropped}{Dropped}{7 (0.4\%)}    & \qcell{silent}{Missed}{950 (50.0\%)}  & 957 (50.3\%) \\
    \midrule
    Total & 852 (44.8\%) & 1,049 (55.2\%) & 1,901 (100\%) \\
    \bottomrule
  \end{tabular}
\end{table}

\section{Methodology}
\label{sec:method}

We build DISC and compare each incident with its AI actors' disclosures in three steps. First, we collect incidents from the public incident databases and code each incident with its risk type, AI subject, impact layer and the AI actors that provided or deployed the system (\Cref{sec:method-incidents}).
Second, we collect the SEC filings of those AI actors and code the risk types the filings disclose (\Cref{sec:method-disclosures}).
Third, we link each incident to the disclosures of its AI actors through the risk type and place it in one of four quadrants (\Cref{sec:method-linkage}).
The unit of analysis is the incident.

\subsection{Collecting and Coding AI Incidents}
\label{sec:method-incidents}

We draw incidents from three public databases, the AIID~\citep{AIID}, the AIAAIC~\citep{AIAAIC} and the OECD AIM~\citep{OECD_AIM}, as of September 1, 2026 for the AIID and September 8, 2026 for the AIAAIC and the OECD AIM.
The AIID and the AIAAIC directly curate incidents, whereas the OECD AIM clusters news reports about AI incidents and hazards.
We therefore first screen each AIM report for whether harm occurred and whether an AI system contributed to it, following the OECD's distinction between an incident and a hazard~\citep{oecd2024incidents}.
We then group records across the three databases into incidents and merge an AIM record that duplicates an AIID or AIAAIC incident into the corresponding incident (Appendix~\ref{app:codebooks}).
Following the AIID, we count each harmful event or ongoing episode, including later lawsuits or investigations, as one incident.
After grouping the records, we identify 9,001 incidents recorded between January 1, 2016, and September 6, 2026: 2,872 from the AIID and AIAAIC combined, and 6,129 from the OECD AIM.

We code each incident by deductive content analysis with a codebook written before coding~\citep{krippendorff2019content,elo2008qualitative}, applied by two large language models (LLMs), Claude Opus 5.5 (Coder A) and the smaller Claude Sonnet 5.5 (Coder B)~\citep{chew2023llm,dunivin2025hermeneutics}. Coder A codes every incident from its record and the articles it cites, giving a supporting quotation and an explanation for every label, and Coder B independently codes a fixed 10\% sample. On 100 incidents, Coder A reaches a Cohen's $\kappa$ of 0.92 across the labels against a human coder. Across three runs, $\kappa$ between the two coders is 0.88--0.90 for risk type, 0.76--0.83 for AI subject, 0.73--0.80 for impact layer, 0.86--0.92 for providers and 0.66--0.84 for deployers, substantial to almost perfect agreement~\citep{landis1977kappa}.
We resolve disagreements manually and release the codebooks with the data (\url{https://github.com/SimeonAllmendinger/disc}; Appendix~\ref{app:codebooks}).

The codebook assigns four labels: risk type, AI subject, impact layer and AI actor. The \emph{risk type} identifies the kind of harm, using one of the 24 subdomains of the AI Risk Repository~\citep{slattery2024repository}, such as discrimination (1.1), security vulnerabilities (2.2) or information ecosystem harms (3.2). Of the 24 risk types, 22 occur among the matched incidents.
The \emph{AI subject} identifies who suffers the harm, whether an individual, a group of people, an organisation, society or nature \cite{bogucka2026intersectionality}.
The \emph{impact layer} identifies where the harm arises: in the AI system's capabilities, its interaction with people, or its systemic effects on society~\citep{weidinger2023sociotechnical}.
Finally, the \emph{AI actor} identifies the organisation that provided or deployed the system involved in the incident.

To examine which risks are disclosed, we group related types into three broad categories. Following \citet{raji2022fallacy}, we classify risks involving systems that fail, make errors or are attacked (2.2, 3.1, 7.3) as \emph{system failures}.
Following \citet{shelby2023sociotechnical}, we classify risks involving damage to the information ecosystem, democratic processes or the distribution of power (3.2, 4.1, 6.1) as \emph{social system harms}.
We also use their concept of \emph{representational harm} for incidents involving discrimination (1.1) or toxic content (1.2) that affect groups of people or society, unless the harm is economic, which \citet{shelby2023sociotechnical} class as allocative.

To identify the organisations responsible for each incident, we distinguish between AI \emph{providers} and \emph{deployers}, following Article 3 of the EU AI Act \cite{euaiact2024regulation}. An incident implicates an AI actor if it provided or deployed the system that caused the harm.
Coder A identifies providers and deployers from each incident record and its cited articles under these definitions and not from the databases' labels. The AIID, for example, lists the hackers in the Meta incident of Example~2 in Appendix~\ref{app:examples} as deployers alongside Meta, whose AI system they exploited. An incident is linked to an SEC registrant only when Coder A identifies the registrant, or a brand or subsidiary it owned at the time, as its provider or deployer. A person who uses an AI system privately is not a deployer (Art.~3(4)).

We date each incident by its \emph{start date}, when the harm began, where the records report one, and otherwise by its \emph{record date}, the date under which the database lists it.
Of the 1,901 matched incidents (\Cref{sec:method-linkage}), 1,036 are dated by their start date and 865 by their record date; Appendix~\ref{app:dates} details the dating and shows that our results hold under either date.

\subsection{Collecting and Coding Risk Types in SEC Filings}
\label{sec:method-disclosures}

We collect SEC filings from 328 SEC-registered AI actors involved in at least one incident.
The filings come from EDGAR, the SEC's public filing system (Electronic Data Gathering, Analysis and Retrieval), and run from January 1, 2015 to September 30, 2026.
They number 18,229 filings across three groups: 3,334 \emph{annual} filings (Forms 10-K, 20-F and 40-F), 7,604 \emph{quarterly} filings (Form 10-Q) and 7,291 \emph{current} filings (Forms 8-K and 6-K).

We define a \emph{disclosure} as a specific risk type mentioned in the Risk Factors section (Item 105 of Regulation S-K; \Cref{sec:intro}) of an SEC filing.
To determine which risks a company has disclosed as of a given date, we use its most recent annual filing and any quarterly filings submitted since then.
We examine current filings separately, only to check whether the company reported or responded to an AI incident.

We then use our two LLM coders to identify which AI risks companies disclose, following the disclosure codebook in Appendix~\ref{app:codebooks}.
We divide each filing into passages, each a sentence that names AI or an AI technique or system, together with the sentences before and after it. Such terms include machine learning, large language models, chatbots, facial recognition and driver assistance (Appendix~\ref{app:codebooks}). Coder A assigns each passage the same labels used for incidents: risk type, AI subject and impact layer.
It also records whether the company describes the harm as a possibility, as something that happened elsewhere or as something it experienced itself.
For example, Coder A codes the passage of Meta's 10-K of January 29, 2026, on ``prompt injection, errors, and other issues related to AI agents'' as a hypothetical security vulnerability (2.2).

We identify 7,495 passages in the Risk Factors sections of annual and quarterly filings, of which 2,367 disclose an AI risk type.
Current filings contain another 9,974 passages, of which 231 disclose an AI risk type.
\Cref{fig:flow} in Appendix~\ref{app:flow} shows how these passages and the incidents enter the analysis.
To check the coding, a human coder labels a random sample of 100 passages blind, and agrees with Coder A across the labels with a Cohen's $\kappa$ of 0.89. Coder B independently codes every passage that Coder A identified as containing an AI risk type, along with a fixed 10\% sample of passages from current filings.

We also check whether companies change their risk disclosures after an incident.
If a company had already disclosed the incident's risk type, we compare that passage with its next periodic filing.
We classify the passage as \emph{retained} if the text remains unchanged or \emph{revised} if it changes.
We distinguish three types of revision: \emph{incident-specific}, when the new text refers to the incident's technology, product, feature, AI actor or harm; \emph{risk-specific}, when it adds information about the risk type without referring to the incident in these ways; and \emph{unspecific}, when the text changes but adds nothing about the incident's risk type.
On the 29 passages that both LLM coders coded, they agree on the revision type with a Cohen's $\kappa$ of 0.84.

\subsection{Matching AI Incidents to SEC Disclosures}
\label{sec:method-linkage}

We match each incident to the SEC filings of its AI actors by risk type, the one label coded with the same codebook for incidents and filing passages. The risk type balances fidelity and interpretability. A coarser match loses fidelity, because it joins harms that filings treat differently and would, for instance, count a warning of inaccurate chatbot output as disclosure of harm to the information ecosystem. A finer match loses interpretability. A risk factor written before an incident can name only the kind of harm that may occur, not the incident itself, so a text-level match, such as one by text similarity, could not show whether that kind of harm was disclosed. We therefore use the 24 subdomains of the AI Risk Repository, which synthesise 1,725 risks from 74 frameworks~\citep{slattery2024repository} and already classify the incidents of the AIID~\citep{mylius2025scalable}. Two checks show that this choice does not drive our results. First, the match is valid. In 89 of 94 sampled \textsc{Covered} incidents, two independent coders found that the matched passage describes the harm that occurred (Appendix~\ref{app:matching}). Second, coarser matches raise the share of \textsc{Covered} incidents but leave our finding intact. With the seven risk domains, 42\% of incidents remain \textsc{Missed}, and even if any disclosed AI harm counted, 26\% would (Appendix~\ref{app:granularity}). Since almost half of the valid matches rest on a general sentence about the AI actor's AI as a whole, a text-level match would only raise the share of \textsc{Missed} incidents.

For each incident, we make two comparisons.
First, we check whether the AI actor had already disclosed the incident's risk type before it occurred (\emph{previously disclosed}).
Second, we check whether the actor disclosed that risk type in its next periodic filing or reported the incident in a current filing (\emph{disclosed}).

These two comparisons place each incident in one of four categories (\Cref{tab:four-quadrants}): \textsc{Covered} if the risk was disclosed both before and after the incident, \textsc{Eventually Disclosed} if it was disclosed only afterwards, \textsc{Dropped} if it was disclosed only beforehand, and \textsc{Missed} if it was never disclosed.
In \Cref{fig:cases}, Alphabet's Gemini incident is \textsc{Covered} and its Turkish media incident is \textsc{Missed}. Appendix~\ref{app:examples} gives seven examples across all quadrants.

Some incidents involve more than one AI actor, such as a provider and a deployer.
In these cases, we compare the incident with each actor's filings and count it as previously disclosed or disclosed if at least one actor disclosed the relevant risk.
When reporting results by AI actor, we assign the incident to its deployers, or to its providers if no deployer is identified.
We divide its weight equally among these actors so that each incident counts only once.

Our analysis requires SEC filings from both before and after an incident. Of the 9,001 incidents, 2,230 involve at least one SEC-registered AI actor.
We retain 1,901 incidents for which an actor had an annual filing in force before the incident and submitted another periodic filing afterwards.
We exclude the remaining 329: 144 involve AI actors that had not yet made an annual filing, 133 had no annual filing with risk factors in our data before the incident (69 of them because the incident predates the first annual filings of 2016), and 52 had no subsequent periodic filing by September 30, 2026 (\Cref{fig:flow}).

Finally, we check the robustness of the results (\Cref{sec:rq2}): we shuffle risk types among each AI actor's incidents within a year, since companies may disclose risks regardless of their exposure~\citep{li2018sec}, compare risk types within the same AI actors, and date incidents differently (Appendix~\ref{app:dates}).

\section{Results}
\label{sec:results}

\Cref{tab:four-quadrants} shows that SEC disclosures cover half of the 1,901 matched incidents (944, 49.7\%). Most fall into two quadrants: 950 are \textsc{Missed} and 845 \textsc{Covered}. Only 99 are \textsc{Eventually Disclosed}, 22 of them reported directly in current filings and one in a periodic filing. The seven \textsc{Dropped} incidents followed settled lawsuits (four) or temporary conditions of content review during remote work (three); as these removals did not respond to the incidents, we leave them out below.

\subsection{RQ1: Which AI Incidents Never Appear in Securities Disclosures?}
\label{sec:rq1}

\begin{figure}[!htbp]
    \centering
    \includegraphics[width=0.8\linewidth]{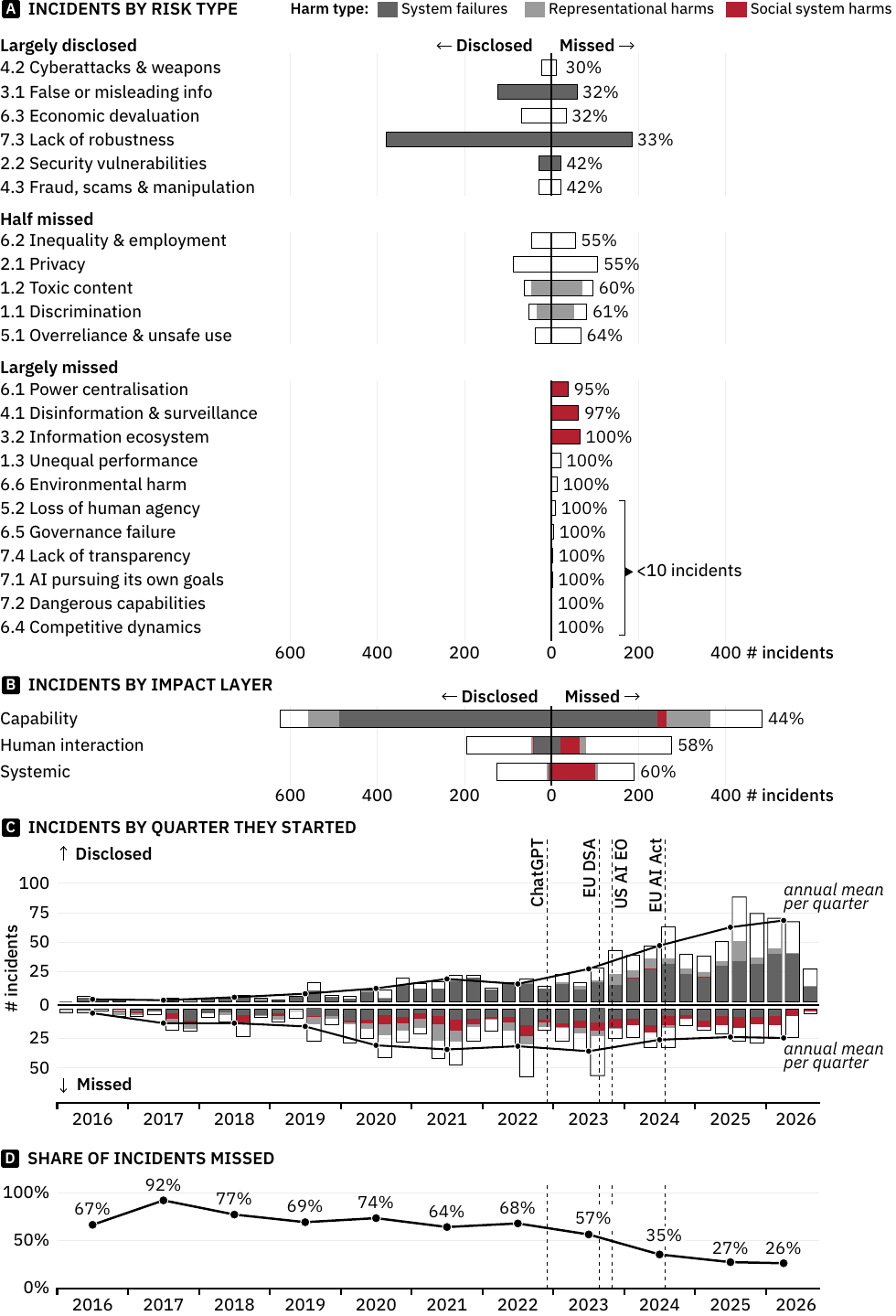}
    \caption{\textbf{AI incidents are disclosed unevenly across risk types, impact layers, and time.} (A) System failures are largely disclosed (\textsc{Covered} or \textsc{Eventually Disclosed}), while social system harms are almost entirely \textsc{Missed}; (B) The share of \textsc{Missed} risks rises from 44\% at the technical capability layer to 60\% at the systemic layer; (C) The number of disclosed risks increases after ChatGPT's release, while \textsc{Missed} risks remain relatively stable. (D) The share of \textsc{Missed} incidents falls from 68\% in 2022 to 26\% in the first half of 2026. Dashed lines in (C) and (D) mark ChatGPT's release (November 2022), the application of the EU Digital Services Act to very large online platforms (August 2023), the US executive order on AI (October 2023) and the entry into force of the EU AI Act (August 2024).}
    \label{fig:rq1}
\end{figure}

\noindent \textbf{Some types of AI incidents are far more often \textsc{Missed} than others.}
\Cref{fig:rq1}~(A) reveals three bands of risk types.
The first group (largely disclosed), with the highest disclosure rates, includes cyberattacks and weapons (4.2), false or misleading information (3.1), lack of robustness (7.3), economic and cultural devaluation (6.3), security vulnerabilities (2.2) and fraud, scams and manipulation (4.3). Between 30\% and 42\% of these incidents are \textsc{Missed}.
The second group (half \textsc{Missed}) includes privacy, employment, toxic content, discrimination and unsafe use, with 55--64\% \textsc{Missed}.
The third group (largely \textsc{Missed}) is almost entirely absent from disclosure, with 95--100\% \textsc{Missed}, including disinformation, harm to the information ecosystem and power centralisation.

These differences also appear in whom incidents harm and where the harm arises (\Cref{fig:rq1}~(B)).
The share \textsc{Missed} rises from 44\% (484 of 1,107) at the capability layer to 58\% (276 of 472) in human interaction and 60\% (190 of 315) at the systemic layer.
Similarly, 41\% of incidents harming individuals or organisations are \textsc{Missed}, compared with 53\% of those harming groups and 78\% of those harming society.
Meanwhile, disclosure has expanded since ChatGPT's release in November 2022: disclosed incidents rose from about 20 per quarter in 2021 to about 70 in the first half of 2026 (annual means), while \textsc{Missed} incidents remained between about 25 and 35 per quarter (\Cref{fig:rq1}~(C)). As a result, the share of incidents \textsc{Missed} fell from 68\% in 2022 to 26\% in the first half of 2026 (\Cref{fig:rq1}~(D)).

\smallskip
\noindent \textbf{98\% of social system harms are \textsc{Missed}.}
This includes 100\% (65 of 65) of information ecosystem harms (3.2), 97\% (62 of 64) of disinformation and surveillance incidents (4.1), and 95\% (38 of 40) of power centralisation incidents (6.1).
Overall, 61\% (103 of 169) of social system harms arise at the systemic layer, and 37\% (63 of 169) affect society as a whole.
Yet the share \textsc{Missed} remains near 100\% regardless of impact layer or AI subject.
Nor has it improved over time: 99\% (98 of 99) were \textsc{Missed} before 2024, compared with 96\% (67 of 70) since.

\smallskip
\noindent \textbf{The share of \textsc{Missed} system failures has fallen from 61\% to 19\%.}
Across all years, 33\% (263 of 796) of system failures are \textsc{Missed}, representing 14\% of all incidents.
Most system failures arise at the capability layer (730 of 796), and these capability failures account for 26\% (243 of 950) of all \textsc{Missed} incidents. Unlike social system harms, however, system failures increasingly enter disclosure (\Cref{fig:rq1}~(C)).
The share \textsc{Missed} fell from 61\% (56 of 92) in 2016--19 to 19\% (71 of 378) in 2024--26. In 2016--19, all 36 disclosed system failures involved Tesla.
By 2024--26, disclosure covered 307 system failures, mainly involving Alphabet (165), Tesla (56), Meta (39) and Microsoft (21).

\smallskip
\noindent \textbf{The share of \textsc{Missed} representational harms fell from 85\% before 2024 to 3\% since.}
Representational harms, which reinforce unjust hierarchies among social groups~\citep{shelby2023sociotechnical}, were \textsc{Missed} in 85\% (120 of 141) of incidents before 2024, compared with just 3\% (2 of 62) since.
A similar decline appears in the half \textsc{Missed} band of risk types in \Cref{fig:rq1}~(A), which includes discrimination (1.1), toxic content (1.2), privacy (2.1), overreliance and unsafe use (5.1) and employment (6.2).
Although 55--64\% of incidents in each of these risk types are \textsc{Missed} overall, the share for the group fell from 76\% (324 of 425) before 2024 to 30\% (79 of 263) since.
This decline follows ChatGPT's release and coincides with the application of the EU Digital Services Act to very large online platforms (August 2023), the US executive order on AI (October 2023) and the entry into force of the EU AI Act (August 2024), marked in \Cref{fig:rq1}~(C) and~(D). Our data cannot separate these events, but disclosure rose together across risk types after ChatGPT's release, which points to a shared shock rather than to any single rule (Appendix~\ref{app:diffusion}).

\subsection{RQ2: Which AI Incidents Are Previously Disclosed in Securities Disclosures?}
\label{sec:rq2}

\begin{figure*}[t]
    \centering
    \includegraphics[width=0.95\textwidth]{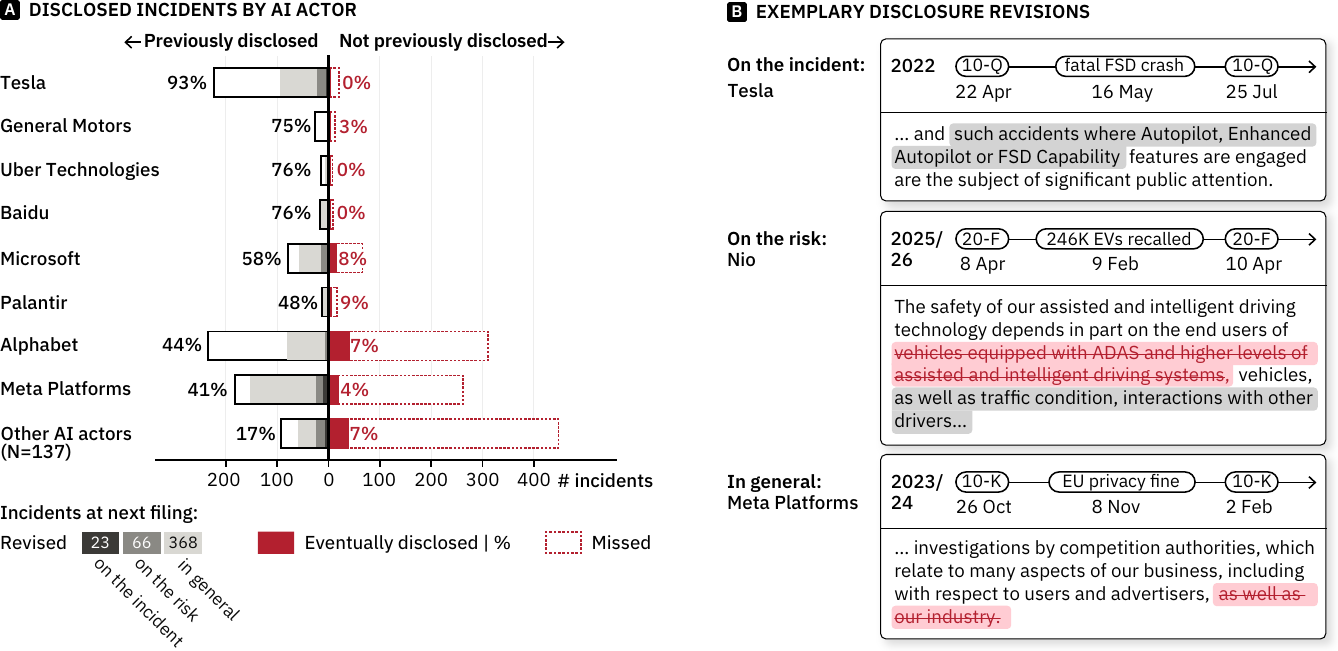}
    \caption{\textbf{Few AI actors disclose most incidents, and their disclosures rarely change after incidents occur.} (A) The top eight AI actors account for 87\% of disclosed incidents. Among them, Tesla previously disclosed 93\% of its incidents, compared with 41\% at Meta; (B) Most disclosures remain unchanged or are revised only in general terms. Examples from Tesla, Nio, and Meta show incident-specific, risk-specific, and unspecific revisions after the incidents. Red text marks deletions from the earlier filing; grey text marks additions in the next filing. Only 1\% of incidents lead to incident-specific revisions.}
    \label{fig:rq2}
\end{figure*}

\smallskip
\noindent \textbf{Eight AI actors previously disclosed 54\% of their incidents; the other 137 previously disclosed 17\%.}
In \Cref{fig:rq2}~(A), eight AI actors hold 822 of the 944 disclosed incidents (87\%), and they previously disclosed 54\% of all matched incidents attributed to them.
In comparison, the other 137 AI actors previously disclosed 17\% of theirs.
Among the eight, the share previously disclosed ranges from 41\% at Meta and 44\% at Alphabet to 93\% at Tesla. Tesla had disclosed risks related to accidents involving driver-assistance features before most of its incidents occurred.
Previous disclosure also varies by the actor's role: 58\% of incidents were previously disclosed when the actor provided the AI system, 38\% when it both provided and deployed it, and 24\% when it only deployed it.

\smallskip
\noindent \textbf{Only 9\% of incidents not previously disclosed are \textsc{Eventually Disclosed}.}
Of the 1,049 not previously disclosed incidents, only 9\% (99 of 1,049) are \textsc{Eventually Disclosed}, meaning their risk types enter disclosure after the incident.
This share remains low across AI actors: 12\% at Alphabet, 6\% at Meta, 19\% at Microsoft and 9\% across the other 137 actors (as shares of each actor's incidents not previously disclosed; \Cref{fig:rq2}~(A) shows them as shares of all its incidents).
Nearly two-thirds of these incidents (63 of 99) started in 2023 or 2024, when Alphabet, Meta and Microsoft expanded the range of AI risks they disclosed.
By risk type, 34\% (17 of 50) of economic and cultural devaluation incidents (6.3) and 21\% (14 of 68) of employment incidents (6.2) not previously disclosed are \textsc{Eventually Disclosed}, against none of the information ecosystem harms (0 of 65) and disinformation incidents (0 of 62) and 1\% of discrimination incidents (1 of 82).

\smallskip
\noindent \textbf{\textsc{Covered} incidents rarely lead to specific changes in disclosure.}
Of the 845 \textsc{Covered} incidents, most leave the relevant risk passage unchanged (45\%, 378 of 845) or lead only to unspecific revisions (44\%, 368 of 845).
Risk-specific revisions follow 8\% (66 of 845), while just 2.7\% (23 of 845) lead to revisions that name the incident's technology, product, feature or harm.
In the remaining 1.2\% (10 of 845), the AI actor reported the incident itself, five times in current filings and five times in periodic filings (\Cref{fig:rq2}~(A)).

\Cref{fig:rq2}~(B) illustrates the three revision types: Meta's February 2024 filing after an EU fine over personalised advertising (\emph{unspecific}), Nio's April 2026 filing after a recall involving AI software (\emph{risk-specific}) and Tesla's July 2022 filing after a fatal crash, which named its Full Self-Driving feature (\emph{incident-specific}). Across all 1,901 matched incidents, only 1\% (23) led to an incident-specific revision, and 1.7\% (33) were reported by the AI actor itself.

\smallskip
\noindent \textbf{Neither chance nor differences between AI actors explain the gap in previous disclosure.}
First, companies may previously disclose more incidents simply because they disclose more risk types. Shuffling risk types among each AI actor's incidents within each year (1,000 runs) reproduces the observed 45\% previously disclosed, but would also yield 54\% of system failures and 29\% of social system harms, against 62\% and 1\% observed. Chance therefore cannot explain why so few social system harms are previously disclosed.

Second, we test whether a few large AI actors drive this gap.
Among the 13 actors involved in both types of incidents, the difference in the share previously disclosed is 52 percentage points, close to the 61-point gap across all incidents.
The pattern holds at Alphabet (66\% of system failures previously disclosed versus 0\% of social system harms), Meta (61\% versus 4\%) and Microsoft (60\% versus 0\%).
Amazon and Apple previously disclose almost neither type. The gap therefore persists even when we compare different harms within the same AI actors.
Third, the results hold whether incidents are dated by when the harm began or by when the incident was recorded (Appendix~\ref{app:dates}). Fourth, AI actors began to disclose risk types together after ChatGPT's release, rather than following the AI actors that disclosed them first (Appendix~\ref{app:diffusion}).

\section{Discussion}
\label{sec:discussion}

Financial materiality, the standard of securities disclosure, asks whether a risk matters to investors~\citep{sec2020regulation,issb2023ifrs}. A harm to AI subjects should therefore enter disclosure once it becomes a financial risk for the AI actor~\citep{jensen2002value}, as stakeholder pressure and regulation turn it into a cost~\citep{freiberg2020how}. This expectation is long-standing, and \citet{freiberg2020how} present it as a prediction for future empirical work. We provide such a test for AI by comparing each recorded incident with the disclosure of its AI actors before and after it. Our results suggest that AI risks enter securities disclosure primarily when harm to AI subjects is reflected in financial risks for the AI actor. Disclosure covers two-thirds of system failures but only 2\% of social system harms (\Cref{sec:rq1}). Of the 383 disclosed incidents in AI system safety, failures and limitations (risk domain 7), the disclosing passages of 245 name only the AI actor as harmed (Appendix~\ref{app:flow}).
Most disclosed incidents are \textsc{Covered}, and few filings change substantially afterwards (\Cref{sec:rq2}).
This pattern is consistent with financial materiality, but also reveals why impact materiality may be needed to capture harms to AI subjects that impose no financial cost on AI actors.

\subsection{Which Harms to AI Subjects Does Disclosure Cover?}
\label{sec:disc-harms}

Harms to AI subjects may enter securities disclosure when they also impose costs on AI actors. For example, people harmed by discrimination (1.1) or privacy violations (2.1) can file complaints, seek compensation or stop using a company's services; for privacy violations in the EU, the GDPR gives them a right to complain and to claim compensation~\citep{gdpr2016regulation}.
These actions can turn harm to AI subjects into financial risks for the AI actor.
Social system harms are different.
Disinformation and surveillance (4.1), harms to the information ecosystem (3.2) and power centralisation (6.1) affect society broadly. Because they fall on the public at large rather than on identifiable persons, they rarely give any individual a claim against the AI actor.
If disclosure follows financial risk, we would therefore expect it to capture discrimination and privacy more often than social system harms.
A similar pattern appears in sustainability reporting, where companies are less likely to consider harms to affected communities material than harms to workers or consumers~\citep{nahrstedt2026double}.

Our findings partly support this expectation.
Disclosure covers fewer than half of discrimination and privacy incidents, but almost no social system harms (\Cref{sec:rq1}), even though social system harms occur about as often as incidents involving false or misleading information (3.1).
Disclosure of discrimination and privacy risks is also concentrated among a few AI actors, particularly Meta and Microsoft, while Amazon and Apple rarely disclose them.
A similar difference appears for overreliance and unsafe use (5.1): Tesla discloses risks related to its driver-assistance systems, while Alphabet, Meta and Snap leave risks from their chatbots largely undisclosed.

These differences suggest that harms to AI subjects are more likely to enter disclosure when someone can impose costs on the AI actor. Social system harms may never create a material financial risk, or companies may fail to disclose them even when they do. Because we observe disclosures rather than financial risks, we cannot tell which, but the contrast with discrimination and privacy points to the first. Harms that affect society but cost AI actors little may thus never be disclosed, a gap that \emph{impact materiality}~\citep{ec2019guidelines,csrd2022directive} or incident reporting on Form 8-K~\citep{strauss2025governing} could narrow, but that securities disclosure alone cannot close (\Cref{sec:disc-position}).

This gap matters because AI actors justify their AI systems by the benefits these bring to society. Alphabet, for instance, committed in its AI principles that its AI should ``be socially beneficial'' and that it would proceed only where the likely benefits substantially exceed the foreseeable risks~\citep{pichai2018ai}. Yet the harms that fall on society are the ones securities disclosure least often names (\Cref{sec:rq1}). AI actors thus present the benefits of AI as shared with society, while their filings record the risks to their investors.

\subsection{Which Incidents That Pose a Risk to the AI Actor Does Disclosure Cover?}
\label{sec:disc-costs}

A risk is material if a reasonable investor would consider it important (\Cref{sec:intro}), so disclosure should cover the incidents that pose a financial risk to the AI actor directly: the system failures (2.2, 3.1, 7.3), cyberattacks and weapons (4.2) and economic and cultural devaluation (6.3).
Our results fit, as disclosure covers most system failures.
However, fraud, scams and manipulation (4.3) depart from this expectation.
These incidents should pose a risk to the AI actor, yet 42\% of them are \textsc{Missed}.
In most of them, a third party or a user misused the AI system, for example, to run deepfake scam advertisements on Facebook, so the harm falls on the victims of the scam and rarely reaches the AI actor, partly because platforms are largely shielded from liability for third-party content~\citep{congress1996protection}.
Where disclosure does cover fraud, the risk type is more often \textsc{Eventually Disclosed} than elsewhere (25\% of fraud incidents not previously disclosed, against 9\% overall).
Fraud thus behaves like a harm to AI subjects, and disclosure catches up only once pressure reaches the AI actor.
The largest AI actors also depart from the expectation.
At Amazon and Apple, disclosure of the risk type was rarely in force (8 of 150 and 12 of 72 incidents) when their AI systems injured workers or failed their users, although these incidents harmed both AI actors.
One explanation is that a reasonable investor evaluates risk relative to the whole AI actor, so a single incident at an AI actor of this size is not material, especially where AI is not the main business.
A harm would then lead to disclosure only once it is large relative to the size of the AI actor.

Five AI actors (Alphabet, Meta, Tesla, Amazon and Microsoft) account for 76\% of matched incidents; for Alphabet, a single incident such as a misleading AI-generated image in Google Search may be too small relative to its revenue to warrant disclosure. Disclosure may also depend on the AI actor's role: AI actors previously disclose more incidents of systems they provide than of systems they only deploy (\Cref{sec:rq2}), although financial materiality concerns financial risk rather than role. Deployers may cover AI risks under general warnings about product defects, which can miss risks specific to AI systems, whose behaviour depends on training data and which providers can change without notifying deployers.

\subsection{What Does Disclosure Do Once an Incident Has Occurred?}
\label{sec:disc-account}

AI risk disclosures have grown but remain generic~\citep{uberti2025seriously,jaworski2026ai}.
Our findings confirm this pattern: most disclosed incidents are \textsc{Covered}, few are \textsc{Eventually Disclosed}, and even when companies revise their disclosures, they rarely refer to the incident's technology, product or harm, and fewer than 2\% of incidents are reported by the AI actor itself (\Cref{sec:rq2}).
Meta, for example, added AI agents to its risk disclosure without acknowledging that the risk had materialised.
This rarity matches \citet{chincholikar2026ai}, who find incident-specific disclosure for 1.3\% of the incidents they trace into SEC filings.

Why should an AI actor keep disclosing a risk once it has mitigated it after an incident? Whether a risk remains material after mitigation is the AI actor's judgement~\citep{sec2020regulation}. Yet mitigation addresses one failure of one system while the risk type spans the AI actor's other systems, an incident's consequences can outlast the mitigation~\citep{sec2018cyber,sec2023cyber}, and presenting a risk as hypothetical after it has occurred can mislead investors~\citep{sec2018altaba,sec2019facebook,sec2021pearson,facebook2023securities} (Appendix~\ref{app:legal}). We therefore read \textsc{Dropped} and \textsc{Missed} as the absence of disclosure, not as violations of the law.

For AI subjects, the more important question is whether disclosure leads companies to reduce risks.
Some incidents prompted action: Cruise recalled the software of all 950 vehicles after its October 2023 accident~\cite{CruiseIncident2023}, Meta shut down Facebook's face recognition in November 2021, eight months after a court approved its \$650 million Illinois privacy settlement~\cite{MetaIncident2021}, and Snap added safeguards to My AI~\cite{SnapchatIncident2023}.
Yet these changes followed pressure from regulators, courts or the press, rather than disclosure itself.
The potential benefit of securities disclosure is therefore less direct: once a risk enters a company's filings, it may also become part of its internal risk management, subject to monitoring, oversight and mitigation.

\subsection{Is Financial Materiality Enough?}
\label{sec:disc-position}

The debate over materiality asks which harms companies should disclose. Our study shows which harms they actually disclose.
AI incidents tend to surface in securities filings when they expose AI actors to lawsuits, fines or lost revenue.
Harms borne by the public, without such financial consequences, often remain undisclosed.
Even when companies disclose a risk, they rarely acknowledge that it has already caused harm. These findings suggest that financial materiality alone is insufficient and support adding \emph{impact materiality} to capture harms to AI subjects.

One might argue that AI regulation should protect AI subjects and securities disclosure serve investors. Yet the EU AI Act requires serious incidents to be reported to authorities, not to the public~\citep{euaiact2024regulation}, and public disclosure is what gives AI actors an incentive to reduce risks (\Cref{sec:intro}).
Without it, harms that impose no financial cost on AI actors may receive less attention and continue to occur.

This gap also reflects the purpose of SEC filings, which serve the SEC's mission to protect investors~\citep{sec2023mission}, so a harm that poses no financial risk to the AI actor has no reason to appear in them. Securities disclosure therefore covers at most half of the 1,901 matched incidents, and because a match by risk type is lenient, even this half is an upper bound (\Cref{sec:method-linkage}). Most AI incidents lie outside securities disclosure altogether, since 6,771 of the 9,001 involve no SEC-registered AI actor. That the other half remains undisclosed by its risk type suggests that these incidents harm AI subjects but do not yet pose a financial risk to the AI actor. This calls for public authorities that hold AI actors to account for such harms and turn them into financial risks, for instance through fines. Ofcom can fine online services up to GBP~18 million or 10\% of their qualifying worldwide revenue, whichever is greater~\citep{osa2023online}; the European Commission can fine very large online platforms up to 6\% of their worldwide annual turnover~\citep[Art.~74]{dsa2022regulation}; and the EU AI Act allows fines of up to EUR~35 million or 7\% of worldwide annual turnover, whichever is higher, for prohibited AI practices~\citep[Art.~99]{euaiact2024regulation}. Competition authorities can address the centralisation of power (6.1), whereas AI safety institutes evaluate AI models but cannot impose fines~\citep{grossman2026aisi}. Once a harm to AI subjects poses a financial risk to the AI actor, it becomes material~\citep{freiberg2020how} and would be expected to enter the AI actor's risk management and, in turn, its securities disclosure, which then records the financial risks that these authorities create.

\subsection{Limitations}
\label{sec:limitations}

Our study has five main limitations.
First, incident databases rely on media coverage and therefore miss unreported incidents and overrepresent large, English-speaking AI actors. The share of \textsc{Missed} incidents may differ for less-covered actors.
Second, our sample includes only SEC registrants and excludes private AI actors. OpenAI alone is linked to more incidents than any registrant except Alphabet, followed by TikTok, Anthropic and X. We also link only the provider and deployer of the AI system, not the provider of a component such as a general-purpose model; counting these leaves the results unchanged (Appendix~\ref{app:upstream}).
Third, we match incidents to disclosures by risk type.
A match therefore indicates that a filing disclosed the relevant risk, not that it adequately described the incident, and a text-level match would count fewer incidents as \textsc{Covered} (Appendix~\ref{app:matching}).
Fourth, our labels rely on LLM coders, which may misinterpret incidents or disclosure passages.
Fifth, we observe disclosures but not costs.
We therefore cannot determine whether a \textsc{Missed} incident imposed no cost on the AI actor or imposed a cost that went undisclosed.

\section{Conclusion and Outlook}
\label{sec:conclusion}

We matched AI incidents to the securities disclosure of the AI actors they implicate, and read each incident against the disclosure in force when it started and at the next filing.
Disclosure covers half of the matched incidents, almost always through a risk type it had previously disclosed.
It covers system failures far more often than social system harms, and our results suggest that disclosure catches up only where someone can hold the AI actor to account for a harm.
Securities filings thus record the risks that AI actors face, but not the harms that AI subjects bear.

Future research could build on DISC in three ways.
First, linking AI incidents to fines, lawsuits and stock returns could reveal whether \textsc{Missed} incidents imposed no financial cost on AI actors or created financial risks that companies failed to disclose.
Second, European sustainability reporting offers an opportunity to test \emph{impact materiality}~\citep{csrd2022directive}.
As companies begin reporting under these requirements, researchers could compare their disclosures with the same AI incidents to test whether social system harms missing from SEC filings become visible.
Such a comparison would benefit from a taxonomy built from incident records~\citep{li2025closer,abercrombie2024collaborative}, which could match incidents and disclosures more specifically and separate harms that coarse categories join, such as environmental harms (6.6) and harms to more-than-human subjects (32 incidents in DISC); public authorities that record AI incidents could adopt it as a shared vocabulary.
Third, AI actors may learn from more than incidents: future research could test whether an AI actor's disclosure becomes broader (more risk types) and deeper (naming its own systems rather than AI in general, Appendix~\ref{app:matching}) with its own experience of AI, apart from incidents and the shared rise after ChatGPT's release (Appendix~\ref{app:diffusion}).
We release DISC at \url{https://simeonallmendinger.github.io/disc/explorer/} (Appendix~\ref{app:explorer}) to support this research.

\clearpage
\section{Endmatter Statements}
\label{sec:statements}

\subsection{Generative AI Usage Statement}

We used generative AI tools for two tasks: data analysis and manuscript development, as described below.

\smallskip
\noindent
\textbf{Use in Data Analysis.} We used Anthropic \textsc{Claude Opus 5.5} (Coder A) and \textsc{Claude Sonnet 5.5} (Coder B) via the API as coders in a deductive content analysis (\Cref{sec:method}), and \textsc{Claude Fable 5.1} (Coder C) as a further coder in the check of the matching (Appendix~\ref{app:matching}). The models screened OECD AIM reports for incidents, grouped duplicate records, identified providers and deployers, and assigned each incident its risk type, AI subject, impact layer and start date. They also coded the AI risk types of passages in SEC filings, the revisions between consecutive filings and whether current filings responded to an incident. The models followed written codebooks and gave a supporting quotation for every label. We checked Coder A against a human coder on 100 incidents and on 100 passages of SEC filings, had Coder B independently re-code samples, and resolved disagreements manually. The codebooks and prompts are released with DISC.

\smallskip
\noindent
\textbf{Use in Manuscript Development.} We used Anthropic \textsc{Claude Opus 5.5} via Claude to edit and revise text, to check the numbers reported in the manuscript against the data, and to run the dating robustness checks in Appendix~\ref{app:dates}. The authors reviewed all changes and take full responsibility for the content.

\subsection{Ethical Considerations Statement}

Our study relies exclusively on public records and involves no human participants. DISC anonymises private individuals and minors, limits quotations from news reports to short excerpts, and is released under CC BY-SA 4.0. A contact is provided for corrections or removal requests. Our labels do not establish whether a filing was misleading under securities law.

Our work raises three further ethical considerations.
First, we name AI actors and report which of their incidents their filings left undisclosed. Incident databases record allegations as well as established facts, and a \textsc{Missed} incident shows only that no filing named its risk type, not that the AI actor breached a duty: materiality is judged against the whole company, and courts decide whether an omission misled investors. We therefore report our results as patterns in disclosure, not as findings about individual filings.

Second, incident databases depend on media coverage and overrepresent large, English-speaking AI actors and the harms that journalists report. Harms to less visible groups and communities are likely under-recorded, so our results may understate how much disclosure leaves out.

Third, incident records describe harms to private individuals, some of them severe, such as wrongful arrests. DISC replaces the names of private individuals with a placeholder, keeps only short quotations as evidence for each label, and offers removal on request. Because LLM coders assigned the labels, each label carries its supporting quotation, so that readers can check it.

\bibliographystyle{plainnat}
\bibliography{references}

@article{ananny2018seeing,
  author  = {Ananny, Mike and Crawford, Kate},
  title   = {Seeing without Knowing: Limitations of the Transparency Ideal and Its Application to Algorithmic Accountability},
  journal = {New Media \& Society},
  volume  = {20},
  number  = {3},
  pages   = {973--989},
  year    = {2018},
  doi     = {10.1177/1461444816676645}
}

@misc{basic1988levinson,
  author       = {{Supreme Court of the United States}},
  title        = {Basic Inc. v. Levinson},
  howpublished = {485 U.S. 224 (1988)},
  year         = {1988}
}

@inproceedings{yildirim2023creating,
  author    = {Yildirim, Nur and Oh, Changhoon and Sayar, Deniz and Brand, Kayla and Challa, Supritha and Turri, Violet and Crosby Walton, Nina and Wong, Anna Elise and Forlizzi, Jodi and McCann, James and Zimmerman, John},
  title     = {Creating Design Resources to Scaffold the Ideation of {AI} Concepts},
  booktitle = {Proceedings of the 2023 {ACM} Designing Interactive Systems Conference},
  series    = {DIS '23},
  year      = {2023},
  pages     = {2326--2346},
  publisher = {ACM},
  address   = {New York, NY, USA},
  doi       = {10.1145/3563657.3596058}
}

@inproceedings{bogucka2026intersectionality,
  author    = {Bogucka, Edyta and {\v{S}}{\'c}epanovi{\'c}, Sanja and Quercia, Daniele},
  title     = {Why {AI} Harms Can't Be Fixed One Identity at a Time: What 5300 Incident Reports Reveal About Intersectionality},
  booktitle = {Proceedings of the 2026 {ACM} Conference on Fairness, Accountability, and Transparency},
  series    = {FAccT '26},
  year      = {2026},
  pages     = {4581--4609},
  publisher = {ACM},
  doi       = {10.1145/3805689.3812347}
}

@article{campbell2014information,
  author  = {Campbell, John L. and Chen, Hsinchun and Dhaliwal, Dan S. and Lu, Hsin-min and Steele, Logan B.},
  title   = {The Information Content of Mandatory Risk Factor Disclosures in Corporate Filings},
  journal = {Review of Accounting Studies},
  volume  = {19},
  number  = {1},
  pages   = {396--455},
  year    = {2014},
  doi     = {10.1007/s11142-013-9258-3}
}

@misc{chew2023llm,
  author        = {Chew, Robert and Bollenbacher, John and Wenger, Michael and Speer, Jessica and Kim, Annice},
  title         = {{LLM}-Assisted Content Analysis: Using Large Language Models to Support Deductive Coding},
  year          = {2023},
  eprint        = {2306.14924},
  archiveprefix = {arXiv},
  primaryclass  = {cs.CL},
  url           = {https://arxiv.org/abs/2306.14924}
}

@article{chincholikar2026ai,
  author  = {Chincholikar, Samir and Chawla, Robin},
  title   = {The {AI} Incident Disclosure Shadow: Tracking {AI} Incidents into {U.S.} Securities Filings},
  journal = {Frontiers in Artificial Intelligence},
  volume  = {9},
  pages   = {1969275},
  year    = {2026},
  doi     = {10.3389/frai.2026.1969275}
}

@article{christensen2017real,
  author  = {Christensen, Hans B. and Floyd, Eric and Liu, Lisa Yao and Maffett, Mark},
  title   = {The real effects of mandated information on social responsibility in financial reports: Evidence from mine-safety records},
  journal = {Journal of Accounting and Economics},
  volume  = {64},
  number  = {2--3},
  pages   = {284--304},
  year    = {2017},
  doi     = {10.1016/j.jacceco.2017.08.001}
}

@inproceedings{cobbe2023accountability,
  author    = {Cobbe, Jennifer and Veale, Michael and Singh, Jatinder},
  title     = {Understanding Accountability in Algorithmic Supply Chains},
  booktitle = {Proceedings of the 2023 ACM Conference on Fairness, Accountability, and Transparency (FAccT '23)},
  publisher = {ACM},
  address   = {New York, NY, USA},
  pages     = {1186--1197},
  year      = {2023},
  doi       = {10.1145/3593013.3594073}
}

@misc{congress1996protection,
  author       = {{United States Congress}},
  title        = {Protection for Private Blocking and Screening of Offensive Material},
  howpublished = {47 U.S.C. \S~230},
  year         = {1996}
}

@misc{csrd2022directive,
  author       = {{European Parliament and Council}},
  title        = {Directive ({EU}) 2022/2464 of the {European Parliament} and of the {Council} of 14 {December} 2022 Amending Regulation ({EU}) No 537/2014, Directive 2004/109/{EC}, Directive 2006/43/{EC} and Directive 2013/34/{EU}, as Regards Corporate Sustainability Reporting},
  howpublished = {Official Journal of the European Union, L 322, pp.~15--80},
  year         = {2022},
  url          = {https://eur-lex.europa.eu/eli/dir/2022/2464/oj}
}

@article{dunivin2025hermeneutics,
  author  = {Dunivin, Zackary Okun},
  title   = {Scaling Hermeneutics: A Guide to Qualitative Coding with {LLMs} for Reflexive Content Analysis},
  journal = {EPJ Data Science},
  volume  = {14},
  pages   = {28},
  year    = {2025},
  doi     = {10.1140/epjds/s13688-025-00548-8}
}

@misc{ec2019guidelines,
  author       = {{European Commission}},
  title        = {Guidelines on Non-Financial Reporting: Supplement on Reporting Climate-Related Information},
  howpublished = {Communication 2019/C 209/01, Official Journal of the European Union, C 209, pp.~1--30},
  year         = {2019},
  month        = jun,
  url          = {https://eur-lex.europa.eu/legal-content/EN/TXT/?uri=CELEX:52019XC0620(01)}
}

@article{elo2008qualitative,
  author  = {Elo, Satu and Kyng{\"a}s, Helvi},
  title   = {The Qualitative Content Analysis Process},
  journal = {Journal of Advanced Nursing},
  volume  = {62},
  number  = {1},
  pages   = {107--115},
  year    = {2008},
  doi     = {10.1111/j.1365-2648.2007.04569.x}
}

@misc{euaiact2024regulation,
  author       = {{European Parliament and Council}},
  title        = {Regulation ({EU}) 2024/1689 of 13 June 2024 Laying Down Harmonised Rules on Artificial Intelligence ({Artificial Intelligence Act})},
  howpublished = {Official Journal of the European Union, L 2024/1689, 12 July 2024},
  year         = {2024},
  url          = {https://eur-lex.europa.eu/eli/reg/2024/1689/oj}
}

@misc{facebook2023securities,
  author       = {{United States Court of Appeals for the Ninth Circuit}},
  title        = {In re Facebook, Inc. Securities Litigation},
  howpublished = {87 F.4th 934 (9th Cir. 2023)},
  year         = {2023}
}

@techreport{freiberg2020how,
  author      = {Freiberg, David and Rogers, Jean and Serafeim, George},
  title       = {How {ESG} Issues Become Financially Material to Corporations and Their Investors},
  institution = {Harvard Business School},
  type        = {Working Paper},
  number      = {20-056},
  year        = {2020}
}

@misc{gdpr2016regulation,
  author       = {{European Parliament and Council}},
  title        = {Regulation ({EU}) 2016/679 on the Protection of Natural Persons with Regard to the Processing of Personal Data and on the Free Movement of Such Data ({General Data Protection Regulation})},
  howpublished = {Official Journal of the European Union, L 119, pp.~1--88},
  year         = {2016},
  url          = {https://eur-lex.europa.eu/eli/reg/2016/679/oj}
}

@misc{issb2023ifrs,
  author       = {{International Sustainability Standards Board}},
  title        = {{IFRS} {S1} General Requirements for Disclosure of Sustainability-related Financial Information},
  howpublished = {IFRS Foundation, London},
  year         = {2023}
}

@misc{jaworski2026ai,
  author        = {Jaworski, Bart and Sutton, Andrew},
  title         = {The {AI} Risk Observatory: What Can We Learn from {AI} Disclosures in Annual Reports About Societal Resilience?},
  year          = {2026},
  eprint        = {2610.02281},
  archiveprefix = {arXiv},
  url           = {https://arxiv.org/abs/2610.02281}
}

@article{jensen2002value,
  author  = {Jensen, Michael C.},
  title   = {Value Maximization, Stakeholder Theory, and the Corporate Objective Function},
  journal = {Business Ethics Quarterly},
  volume  = {12},
  number  = {2},
  pages   = {235--256},
  year    = {2002},
  doi     = {10.2307/3857812}
}

@article{jiang2026artificial,
  author  = {Jiang, Tianqi and Neyland, Jordan and Wang, Zhao and Zhang, Fan},
  title   = {Artificial Intelligence: Problems and Bias},
  journal = {EFMD Global Focus},
  volume  = {20},
  number  = {2},
  pages   = {80--83},
  year    = {2026},
  url     = {https://globalfocusmagazine.com/artificial-intelligence-problems-and-bias/}
}

@inproceedings{kolt2024responsible,
  author    = {Kolt, Noam and Anderljung, Markus and Barnhart, Joslyn and Brass, Asher and Esvelt, Kevin and Hadfield, Gillian K. and Heim, Lennart and Rodriguez, Mikel and Sandbrink, Jonas B. and Woodside, Thomas},
  title     = {Responsible Reporting for Frontier {AI} Development},
  booktitle = {Proceedings of the {AAAI/ACM} Conference on {AI}, Ethics, and Society},
  volume    = {7},
  number    = {1},
  pages     = {768--783},
  year      = {2024},
  doi       = {10.1609/aies.v7i1.31678}
}

@book{krippendorff2019content,
  author    = {Krippendorff, Klaus},
  title     = {Content Analysis: An Introduction to Its Methodology},
  edition   = {4},
  publisher = {SAGE},
  address   = {Thousand Oaks, CA},
  year      = {2019},
  doi       = {10.4135/9781071878781}
}

@article{landis1977kappa,
  author  = {Landis, J. Richard and Koch, Gary G.},
  title   = {The Measurement of Observer Agreement for Categorical Data},
  journal = {Biometrics},
  volume  = {33},
  number  = {1},
  pages   = {159--174},
  year    = {1977},
  doi     = {10.2307/2529310}
}

@article{li2018sec,
  author  = {Li, He and No, Won Gyun and Wang, Tawei},
  title   = {{SEC}'s Cybersecurity Disclosure Guidance and Disclosed Cybersecurity Risk Factors},
  journal = {International Journal of Accounting Information Systems},
  volume  = {30},
  pages   = {40--55},
  year    = {2018},
  doi     = {10.1016/j.accinf.2018.06.003}
}

@inproceedings{li2025closer,
  author    = {Li, Megan and Bickersteth, Wendy and Tang, Ningjing and Cranor, Lorrie and Hong, Jason and Shen, Hong and Heidari, Hoda},
  title     = {A Closer Look at the Existing Risks of Generative {AI}: Mapping the Who, What, and How of Real-World Incidents},
  booktitle = {Proceedings of the {AAAI/ACM} Conference on {AI}, Ethics, and Society},
  volume    = {8},
  number    = {2},
  pages     = {1561--1573},
  year      = {2025},
  doi       = {10.1609/aies.v8i2.36655}
}

@inproceedings{mcgregor2021preventing,
  author    = {McGregor, Sean},
  title     = {Preventing Repeated Real World {AI} Failures by Cataloging Incidents: The {AI} Incident Database},
  booktitle = {Proceedings of the {AAAI} Conference on Artificial Intelligence},
  volume    = {35},
  number    = {17},
  pages     = {15458--15463},
  year      = {2021},
  doi       = {10.1609/aaai.v35i17.17817}
}

@unpublished{nahrstedt2026double,
  author = {Nahrstedt, Jan},
  title  = {Double Materiality in Practice: Evidence from Two Years of {CSRD} Reporting},
  note   = {Working paper, University of Hamburg. SSRN},
  year   = {2026},
  url    = {https://papers.ssrn.com/sol3/papers.cfm?abstract_id=6724699}
}

@misc{oecd2019recommendation,
  author       = {{OECD}},
  title        = {Recommendation of the {C}ouncil on {A}rtificial {I}ntelligence},
  howpublished = {OECD/LEGAL/0449},
  year         = {2019},
  url          = {https://legalinstruments.oecd.org/en/instruments/OECD-LEGAL-0449}
}

@techreport{oecd2024incidents,
  author      = {{OECD}},
  title       = {Defining {AI} Incidents and Related Terms},
  institution = {OECD Publishing},
  address     = {Paris},
  type        = {OECD Artificial Intelligence Papers},
  number      = {16},
  year        = {2024},
  month       = may,
  doi         = {10.1787/d1a8d965-en}
}

@inproceedings{raji2019actionable,
  author    = {Raji, Inioluwa Deborah and Buolamwini, Joy},
  title     = {Actionable Auditing: Investigating the Impact of Publicly Naming Biased Performance Results of Commercial {AI} Products},
  booktitle = {Proceedings of the 2019 {AAAI/ACM} Conference on {AI}, Ethics, and Society},
  pages     = {429--435},
  year      = {2019},
  doi       = {10.1145/3306618.3314244}
}

@inproceedings{raji2022fallacy,
  author    = {Raji, Inioluwa Deborah and Kumar, I. Elizabeth and Horowitz, Aaron and Selbst, Andrew D.},
  title     = {The Fallacy of {AI} Functionality},
  booktitle = {Proceedings of the 2022 {ACM} Conference on Fairness, Accountability, and Transparency},
  series    = {FAccT '22},
  year      = {2022},
  pages     = {959--972},
  publisher = {ACM},
  doi       = {10.1145/3531146.3533158}
}

@inproceedings{rao2025ai,
  author    = {Rao, Pooja S. B. and {\v{S}}{\'c}epanovi{\'c}, Sanja and Jayagopi, Dinesh Babu and Cherubini, Mauro and Quercia, Daniele},
  title     = {The {AI} Model Risk Catalog: What Developers and Researchers Miss About Real-World {AI} Harms},
  booktitle = {Proceedings of the {AAAI/ACM} Conference on {AI}, Ethics, and Society},
  volume    = {8},
  number    = {3},
  pages     = {2150--2163},
  year      = {2025},
  doi       = {10.1609/aies.v8i3.36702}
}

@misc{sec2020regulation,
  author       = {{U.S. Securities and Exchange Commission}},
  title        = {Regulation {S-K}, Item 105: Risk Factors},
  howpublished = {17 C.F.R. \S~229.105},
  year         = {2020},
  url          = {https://www.ecfr.gov/current/title-17/chapter-II/part-229/subpart-229.100/section-229.105}
}

@inproceedings{shelby2023sociotechnical,
  author    = {Shelby, Renee and Rismani, Shalaleh and Henne, Kathryn and Moon, AJung and Rostamzadeh, Negar and Nicholas, Paul and Yilla-Akbari, N'Mah and Gallegos, Jess and Smart, Andrew and Garcia, Emilio and Virk, Gurleen},
  title     = {Sociotechnical Harms of Algorithmic Systems: Scoping a Taxonomy for Harm Reduction},
  booktitle = {Proceedings of the 2023 {AAAI/ACM} Conference on {AI}, Ethics, and Society},
  series    = {AIES '23},
  year      = {2023},
  pages     = {723--741},
  publisher = {ACM},
  doi       = {10.1145/3600211.3604673}
}

@misc{sidhu2026open,
  author        = {Sidhu, Harleen Kaur and Scholefield, Rebecca and Annan, Nour and Hernandez, Kevin and Hou, Isabel Nieh and Alshaikhi, Abdulrahman and Chin, Ze Shen and Gipi{\v{s}}kis, Rokas},
  title         = {Open Problems in {AI} Incident Governance},
  year          = {2026},
  eprint        = {2607.05163},
  archiveprefix = {arXiv},
  primaryclass  = {cs.CY},
  url           = {https://arxiv.org/abs/2607.05163}
}

@article{slattery2024repository,
  author  = {Slattery, Peter and Saeri, Alexander K. and Grundy, Emily A. C. and Graham, Jess and Noetel, Michael and Uuk, Risto and Dao, James and Pour, Soroush and Casper, Stephen and Thompson, Neil},
  title   = {The {AI} Risk Repository: A Meta-Review, Database, and Taxonomy of Risks from Artificial Intelligence},
  journal = {Patterns},
  pages   = {101517},
  year    = {2026},
  doi     = {10.1016/j.patter.2026.101517}
}

@techreport{strauss2025governing,
  author      = {Strauss, Ilan and O'Reilly, Tim and Rosenblat, Sruly and Moure, Isobel},
  title       = {Governing {AI} Through {SEC} Disclosure: Materiality Standards and Incident Reporting---Lessons from Cybersecurity},
  institution = {Social Science Research Council, AI Disclosures Project},
  type        = {Working Paper},
  year        = {2025},
  doi         = {10.35650/AIDP.4120.d.2025}
}

@misc{tsc1976industries,
  author       = {{Supreme Court of the United States}},
  title        = {{TSC} Industries, Inc. v. Northway, Inc.},
  howpublished = {426 U.S. 438 (1976)},
  year         = {1976}
}

@inproceedings{turri2023why,
  author    = {Turri, Violet and Dzombak, Rachel},
  title     = {Why We Need to Know More: Exploring the State of {AI} Incident Documentation Practices},
  booktitle = {Proceedings of the 2023 AAAI/ACM Conference on AI, Ethics, and Society (AIES '23)},
  publisher = {ACM},
  address   = {New York, NY, USA},
  pages     = {576--583},
  year      = {2023},
  doi       = {10.1145/3600211.3604700}
}

@inproceedings{uberti2025seriously,
  author        = {Uberti-Bona Marin, Lucas G. and Rijsbosch, Bram and Spanakis, Gerasimos and Kollnig, Konrad},
  title         = {Are Companies Taking {AI} Risks Seriously? {A} Systematic Analysis of Companies' {AI} Risk Disclosures in {SEC} 10-{K} Forms},
  booktitle     = {Machine Learning and Principles and Practice of Knowledge Discovery in Databases: International Workshops of {ECML PKDD} 2025},
  series        = {Communications in Computer and Information Science},
  publisher     = {Springer},
  address       = {Cham},
  year          = {2026},
  doi           = {10.1007/978-3-032-19096-3_6}
}

@article{wang2013association,
  author  = {Wang, Tawei and Kannan, Karthik N. and Ulmer, Jackie Rees},
  title   = {The Association Between the Disclosure and the Realization of Information Security Risk Factors},
  journal = {Information Systems Research},
  volume  = {24},
  number  = {2},
  pages   = {201--218},
  year    = {2013},
  doi     = {10.1287/isre.1120.0437}
}

@misc{weidinger2023sociotechnical,
  author        = {Weidinger, Laura and Rauh, Maribeth and Marchal, Nahema and Manzini, Arianna and Hendricks, Lisa Anne and Mateos-Garcia, Juan and Bergman, Stevie and Kay, Jackie and Griffin, Conor and Bariach, Ben and Gabriel, Iason and Rieser, Verena and Isaac, William},
  title         = {Sociotechnical Safety Evaluation of Generative {AI} Systems},
  year          = {2023},
  eprint        = {2310.11986},
  archiveprefix = {arXiv},
  primaryclass  = {cs.AI},
  url           = {https://arxiv.org/abs/2310.11986}
}

@inproceedings{wieringa2020what,
  author    = {Wieringa, Maranke},
  title     = {What to Account for When Accounting for Algorithms: A Systematic Literature Review on Algorithmic Accountability},
  booktitle = {Proceedings of the 2020 Conference on Fairness, Accountability, and Transparency},
  series    = {FAT* '20},
  publisher = {ACM},
  pages     = {1--18},
  year      = {2020},
  doi       = {10.1145/3351095.3372833}
}

@online{AIID,
	title        = {AI Incident Database},
	author       = {{Responsible AI Collaborative}},
	year         = 2026,
	url          = {https://incidentdatabase.ai/},
	note         = {Accessed: 2026-09-01}
}

@online{OECD_AIM,
	title        = {{OECD AI Incidents and Hazards Monitor (AIM)}},
	author       = {{OECD}},
	year         = 2026,
	url          = {https://oecd.ai/en/incidents},
	note         = {Accessed: 2026-09-08}
}

@online{AIAAIC,
	title        = {{AI, Algorithmic, and Automation Incidents and Controversies (AIAAIC) Repository}},
	author       = {{AIAAIC}},
	year         = 2026,
	url          = {https://www.aiaaic.org/aiaaic-repository/ai-algorithmic-and-automation-incidents},
	note         = {Accessed: 2026-09-08}
}

@article{Stilgoe2013,
title = {Developing a Framework for Responsible Innovation},
journal = {Research Policy},
volume = {42},
number = {9},
pages = {1568-1580},
year = {2013},
issn = {0048-7333},
doi = {10.1016/j.respol.2013.05.008},
url = {https://www.sciencedirect.com/science/article/pii/S0048733313000930},
author = {Jack Stilgoe and Richard Owen and Phil Macnaghten}
}

@online{CruiseIncident2023,
	title        = {Equipment Recall Report: Collision Detection Subsystem ({NHTSA} Recall 23E-086)},
	author       = {{Cruise LLC}},
	year         = 2023,
	url          = {https://static.nhtsa.gov/odi/rcl/2023/RMISC-23E086-4326.pdf},
	note         = {Submitted to NHTSA, 2 November 2023. Accessed: 2026-10-30}
}

@online{SnapchatIncident2023,
	title        = {Early Learnings from My AI and New Safety Enhancements},
	author       = {{Snap Inc.}},
	year         = 2023,
	url          = {https://values.snap.com/news/early-learnings-from-my-ai-and-new-safety-enhancements},
	note         = {Snap Newsroom, 4 April 2023. Accessed: 2026-10-30}
}

@online{MetaIncident2021,
	title        = {An Update On Our Use of Face Recognition},
	author       = {Pesenti, Jerome},
	organization = {Meta},
	year         = 2021,
	url          = {https://about.fb.com/news/2021/11/update-on-use-of-face-recognition/},
	note         = {Meta Newsroom, 2 November 2021. Accessed: 2026-10-30}
}

@inproceedings{herdel2024exploregen,
  author    = {Herdel, Viviane and {\v{S}}{\'c}epanovi{\'c}, Sanja and Bogucka, Edyta and Quercia, Daniele},
  title     = {{ExploreGen}: Large Language Models for Envisioning the Uses and Risks of {AI} Technologies},
  booktitle = {Proceedings of the {AAAI/ACM} Conference on {AI}, Ethics, and Society},
  volume    = {7},
  number    = {1},
  pages     = {584--596},
  year      = {2024},
  doi       = {10.1609/aies.v7i1.31660}
}

@inproceedings{golpayegani2023highrisk,
  author    = {Golpayegani, Delaram and Pandit, Harshvardhan J. and Lewis, Dave},
  title     = {To Be High-Risk, or Not To Be---Semantic Specifications and Implications of the {AI} Act's High-Risk {AI} Applications and Harmonised Standards},
  booktitle = {Proceedings of the 2023 {ACM} Conference on Fairness, Accountability, and Transparency},
  series    = {FAccT '23},
  publisher = {ACM},
  pages     = {905--915},
  year      = {2023},
  doi       = {10.1145/3593013.3594050}
}

@techreport{oecd2022classification,
  author      = {{OECD}},
  title       = {{OECD} Framework for the Classification of {AI} Systems},
  institution = {OECD Publishing},
  type        = {OECD Digital Economy Papers},
  number      = {323},
  year        = {2022},
  doi         = {10.1787/cb6d9eca-en}
}

@online{mylius2025scalable,
	title        = {Scalable {AI} Incident Classification},
	author       = {Mylius, Simon},
	year         = 2025,
	url          = {https://incidentdatabase.ai/blog/scalable-ai-incident-classification/},
	note         = {AI Incident Database blog, 14 April 2025. Accessed: 2026-10-30}
}

@misc{sec2018cyber,
  author       = {{U.S. Securities and Exchange Commission}},
  title        = {Commission Statement and Guidance on Public Company Cybersecurity Disclosures},
  howpublished = {Release No. 33-10459, 21 February 2018},
  year         = {2018}
}

@misc{sec2023cyber,
  author       = {{U.S. Securities and Exchange Commission}},
  title        = {Cybersecurity Risk Management, Strategy, Governance, and Incident Disclosure},
  howpublished = {Release Nos. 33-11216; 34-97989, 26 July 2023},
  year         = {2023}
}

@misc{sec2018altaba,
  author       = {{U.S. Securities and Exchange Commission}},
  title        = {Altaba, Formerly Known as {Yahoo!}, Charged With Failing to Disclose Massive Cybersecurity Breach; Agrees To Pay \$35 Million},
  howpublished = {Press Release 2018-71, 24 April 2018},
  year         = {2018}
}

@misc{sec2019facebook,
  author       = {{U.S. Securities and Exchange Commission}},
  title        = {Facebook to Pay \$100 Million for Misleading Investors About the Risks It Faced From Misuse of User Data},
  howpublished = {Press Release 2019-140, 24 July 2019},
  year         = {2019}
}

@misc{sec2021pearson,
  author       = {{U.S. Securities and Exchange Commission}},
  title        = {{SEC} Charges {Pearson plc} for Misleading Investors About Cyber Breach},
  howpublished = {Press Release 2021-154, 16 August 2021},
  year         = {2021}
}

@misc{macquarie2024moab,
  author       = {{Supreme Court of the United States}},
  title        = {Macquarie Infrastructure Corp. v. Moab Partners, L.P.},
  howpublished = {601 U.S. 257 (2024)},
  year         = {2024}
}

@misc{facebook2024amalgamated,
  author       = {{Supreme Court of the United States}},
  title        = {Facebook, Inc. v. Amalgamated Bank},
  howpublished = {604 U.S. 4 (2024), certiorari dismissed as improvidently granted},
  year         = {2024}
}

@misc{sec2023mission,
  author       = {{U.S. Securities and Exchange Commission}},
  title        = {Mission},
  howpublished = {What We Do, 9 August 2023},
  year         = {2023},
  url          = {https://www.sec.gov/about/what-we-do},
  note         = {Accessed 2 November 2026}
}

@misc{osa2023online,
  author       = {{UK Parliament}},
  title        = {Online Safety Act 2023},
  howpublished = {2023 c. 50},
  year         = {2023},
  url          = {https://www.legislation.gov.uk/ukpga/2023/50}
}

@misc{dsa2022regulation,
  author       = {{European Parliament and Council}},
  title        = {Regulation ({EU}) 2022/2065 of 19 October 2022 on a Single Market for Digital Services ({Digital Services Act})},
  howpublished = {Official Journal of the European Union, L 277, 27 October 2022},
  year         = {2022},
  url          = {https://eur-lex.europa.eu/eli/reg/2022/2065/oj}
}

@misc{grossman2026aisi,
  author       = {Grossman, Charlie},
  title        = {Making Sense of the {UK}'s {AI} Security Institute},
  howpublished = {Ada Lovelace Institute, 13 July 2026},
  year         = {2026},
  url          = {https://www.adalovelaceinstitute.org/feature/aisi/}
}

@misc{abercrombie2024collaborative,
  author        = {Abercrombie, Gavin and Benbouzid, Djalel and Giudici, Paolo and Golpayegani, Delaram and Hernandez, Julio and Noro, Pierre and Pandit, Harshvardhan and Paraschou, Eva and Pownall, Charlie and Prajapati, Jyoti and Sayre, Mark A. and Sengupta, Ushnish and Suriyawongkul, Arthit and Thelot, Ruby and Vei, Sofia and Waltersdorfer, Laura},
  title         = {A Collaborative, Human-Centred Taxonomy of {AI}, Algorithmic, and Automation Harms},
  year          = {2024},
  eprint        = {2407.01294},
  archivePrefix = {arXiv},
  primaryClass  = {cs.LG},
  url           = {https://arxiv.org/abs/2407.01294}
}

@misc{conferenceboard2025ai,
  author       = {{The Conference Board}},
  title        = {{AI} Risk Disclosures in the {S\&P} 500: Reputation, Cybersecurity, and Regulation},
  howpublished = {The Conference Board and ESGAUGE, based on Form 10-K filings available through 15 August 2025},
  year         = {2025},
  url          = {https://conference-board.org/press/AI-risks-disclosure-2025}
}

@misc{atkins2025iac,
  author       = {Atkins, Paul S.},
  title        = {Remarks at the Meeting of the {SEC} Investor Advisory Committee},
  howpublished = {U.S. Securities and Exchange Commission, 4 December 2025},
  year         = {2025},
  url          = {https://www.sec.gov/newsroom/speeches-statements/atkins-remarks-iac-120425}
}

@misc{pichai2018ai,
  author       = {Pichai, Sundar},
  title        = {{AI} at {Google}: Our Principles},
  howpublished = {Google Blog, 7 June 2018, updated 4 February 2025},
  year         = {2018},
  url          = {https://blog.google/technology/ai/ai-principles/}
}

\appendix

\bigskip
\noindent \LARGE\textbf{Appendix}
\normalsize

\section{Data Flow}
\label{app:flow}

\begin{figure*}[tbp]
\centering
\includegraphics[width=\linewidth]{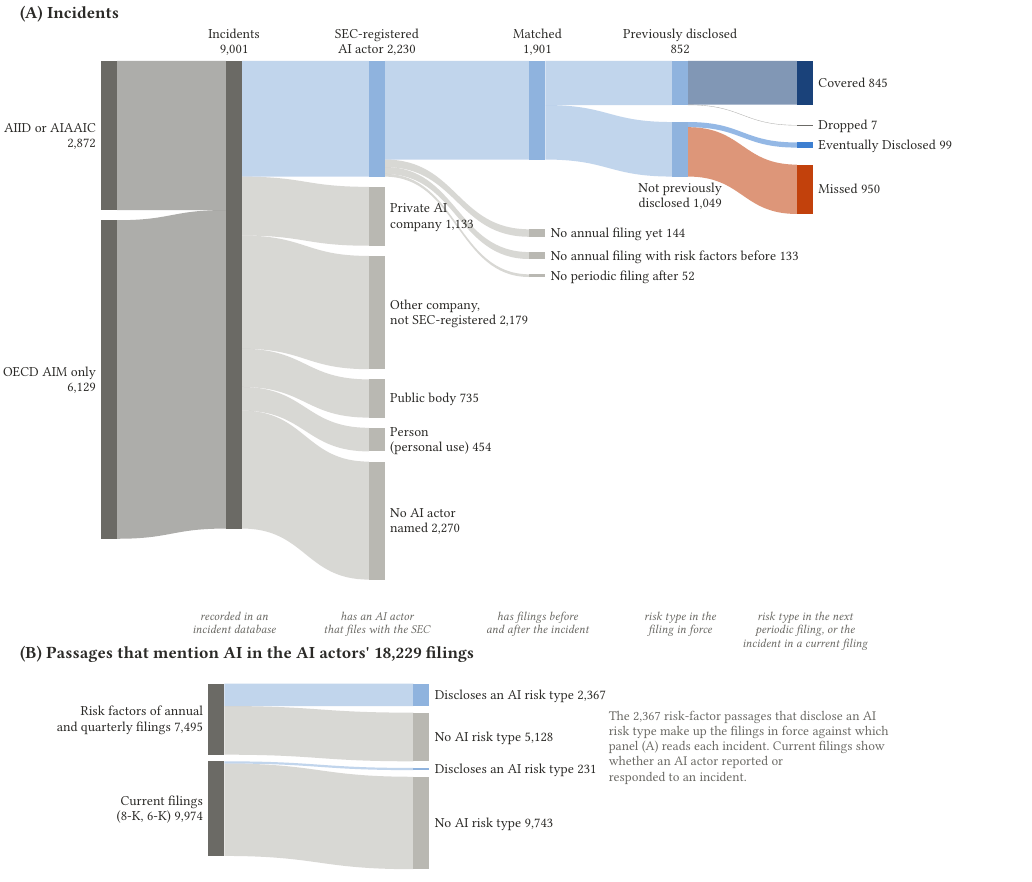}
\caption{How incidents and filing passages enter the analysis. The width of each band is proportional to the number of incidents (A) or passages (B). Grey bands leave the analysis. In (A), the italic line under each column states the condition that the column checks.}
\label{fig:flow}
\end{figure*}

\Cref{fig:flow} traces the quantities of \Cref{sec:method}. Panel (A) follows the incidents from the three databases to the four quadrants of \Cref{tab:four-quadrants}. Of the 9,001 incidents, 6,771 leave the analysis because no implicated AI actor files with the SEC, and another 329 because the SEC-registered AI actor had no annual filing in force before the incident or no periodic filing after it (\Cref{sec:method-linkage}). Of the 1,901 matched incidents, 852 had their risk type disclosed in the filing in force when they started and 1,049 did not. The next periodic filing, or a current filing that reports the incident, then separates \textsc{Covered (845)} from \textsc{Dropped} (7), and \textsc{Eventually Disclosed} (99) from \textsc{Missed} (950). Panel (B) follows the passages that mention AI in the AI actors' filings (\Cref{sec:method-disclosures}).

\begin{figure*}[tbp]
\centering
\includegraphics[width=\linewidth]{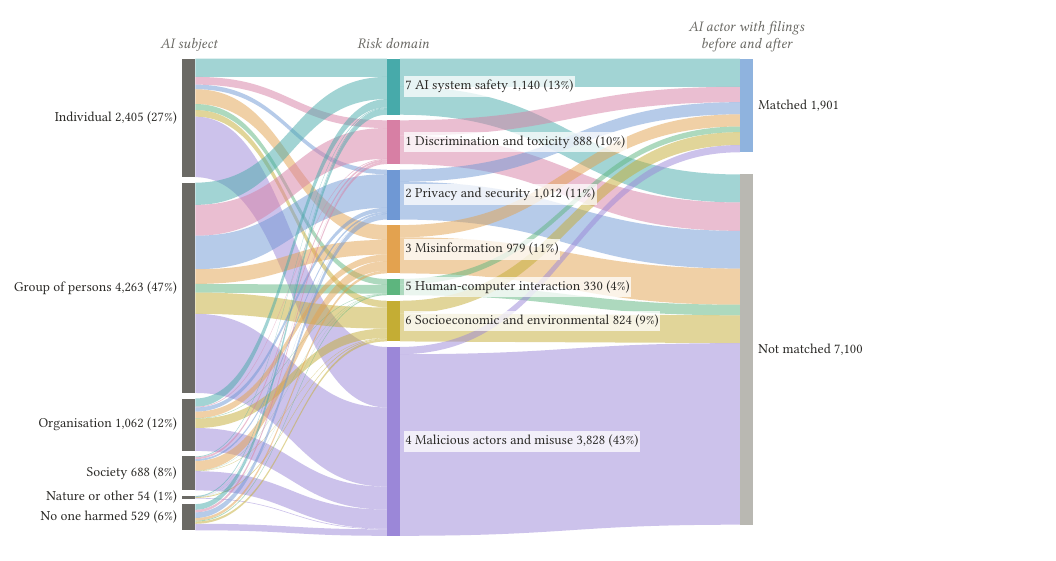}
\caption{All 9,001 incidents by AI subject, risk domain and whether they are matched. Band colours follow the risk domain. Percentages are shares of all incidents.}
\label{fig:labels}
\end{figure*}

\Cref{fig:labels} shows that matching changes the mix of risk domains. Malicious actors and misuse (4) make up 43\% of all incidents but 8\% of the matched incidents, since no AI actor is named for 1,706 of these 3,828 incidents, whereas AI system safety (7) rises from 13\% to 30\%. For the 951 matched incidents whose risk type was disclosed before or after the incident (all quadrants except \textsc{Missed}), we also read whom the passages that disclose the risk type name as harmed. For 525 of these incidents, at least one passage names an AI subject as harmed; for 365, the passages name only the AI actor itself, through liability, fines, lost revenue or reputational harm. These 365 incidents concentrate in AI system safety (245 of 383 disclosed incidents) and human-computer interaction (35 of 38), whereas in the other five risk domains the passages name only the AI actor for 85 of 530 incidents. When an AI actor discloses the risk type of an AI system safety incident, it thus mostly presents the risk as a harm to itself, without naming the people its AI system harms.

\section{Robustness Checks}
\label{app:robustness}

This appendix reports five checks on the main results: whether they depend on how incidents are dated (Appendix~\ref{app:dates}), whether AI actors take up the disclosure of a risk type from the AI actors that disclosed it first (Appendix~\ref{app:diffusion}), whether a match by risk type is also a match in the text (Appendix~\ref{app:matching}), how the results depend on the granularity of the risk type (Appendix~\ref{app:granularity}), and whether they change when upstream providers count as well (Appendix~\ref{app:upstream}).

\subsection{Robustness to Incident Dating}
\label{app:dates}

Where the records report when the harm began, we date an incident by this start date, placing a date known only to the month or year on its first day; this applies to 4,276 of the 9,001 incidents. Otherwise we use the record date. Where a database gives the record date only to the month or year, as the AIAAIC does, we refine it to the earliest dated news report the record cites in that period, but never past a current filing in which the AI actor itself reported the incident; this refines 528 of 929 such dates. Where both dates are known, the record date follows the start date by a median of 117 days (interquartile range 5--485). Where the start date fell after a precise record date (eight matched incidents), we checked the sources and used the more reliable date; none of them changes quadrant.

Of the 1,901 matched incidents, 1,036 are dated by their start date and 865 by their record date. \Cref{tab:dates} shows that the results do not depend on this choice.
Both sets fall into the quadrants in similar shares, and in both, system failures are far more often previously disclosed than social system harms (60\% against 1\% for start-dated incidents, 66\% against 1\% for record-dated ones).
Dating every incident by its record date makes 106 of the start-dated incidents previously disclosed, because the later date credits AI actors with filings made after the harm began. Even then, 47\% of incidents are \textsc{Missed}, and the gap between system failures and social system harms remains (64\% against 2\% previously disclosed).

\begin{table}[h]
\caption{Matched incidents by how they are dated: quadrant shares (\textsc{Covered}, \textsc{Eventually Disclosed}, \textsc{Dropped}, \textsc{Missed}) and the shares of system failures (SF) and social system harms (SSH) previously disclosed and disclosed at the next filing. Dating every incident by its record date changes which incidents have filings before and after them, so $n$ differs.}
\label{tab:dates}
\small
\begin{tabular}{@{}lrrrrrcccc@{}}
\toprule
 & & \multicolumn{4}{c}{Quadrant (\%)} & \multicolumn{2}{c}{Prev.\ disclosed (\%)} & \multicolumn{2}{c}{Disclosed (\%)} \\
\cmidrule(lr){3-6}\cmidrule(lr){7-8}\cmidrule(l){9-10}
Incidents dated by & $n$ & Cov. & Ev. & Dr. & Mi. & SF & SSH & SF & SSH \\
\midrule
start date, else record date & 1,901 & 44.5 & 5.2 & 0.4 & 50.0 & 62 & 1 & 67 & 2 \\
\quad start date & 1,036 & 41.6 & 5.9 & 0.3 & 52.2 & 60 & 1 & 65 & 1 \\
\quad record date & 865 & 47.9 & 4.4 & 0.5 & 47.3 & 66 & 1 & 70 & 4 \\
record date only & 1,965 & 48.7 & 3.8 & 0.2 & 47.4 & 64 & 2 & 69 & 3 \\
\bottomrule
\end{tabular}
\end{table}

\subsection{Diffusion of Risk Disclosure across AI Actors}
\label{app:diffusion}

\begin{figure*}[tbp]
\centering
\includegraphics[width=\linewidth]{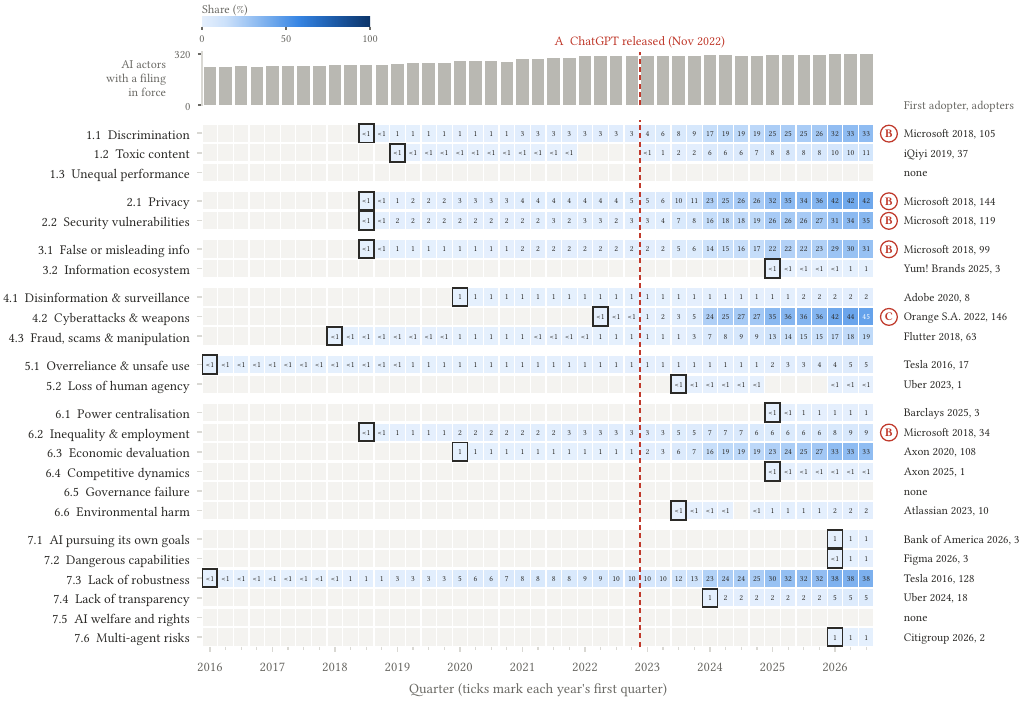}
\caption{Share of AI actors whose filing in force discloses each risk type as an AI harm, by quarter. Each cell shows the percentage of the AI actors with a filing in force at the quarter's end (top bars); an outlined cell marks the quarter in which the first AI actor disclosed the risk type, named at the right with the number of AI actors that ever disclosed it. (A) ChatGPT's release in November 2022. (B) Microsoft disclosed five risk types in 2018, but each reached at most 5\% of AI actors before 2023. (C) Cyberattacks and weapons rose from one AI actor at the end of 2022 to 45\% of AI actors.}
\label{fig:diffusion-actors}
\end{figure*}

\begin{figure*}[tbp]
\centering
\includegraphics[width=\linewidth]{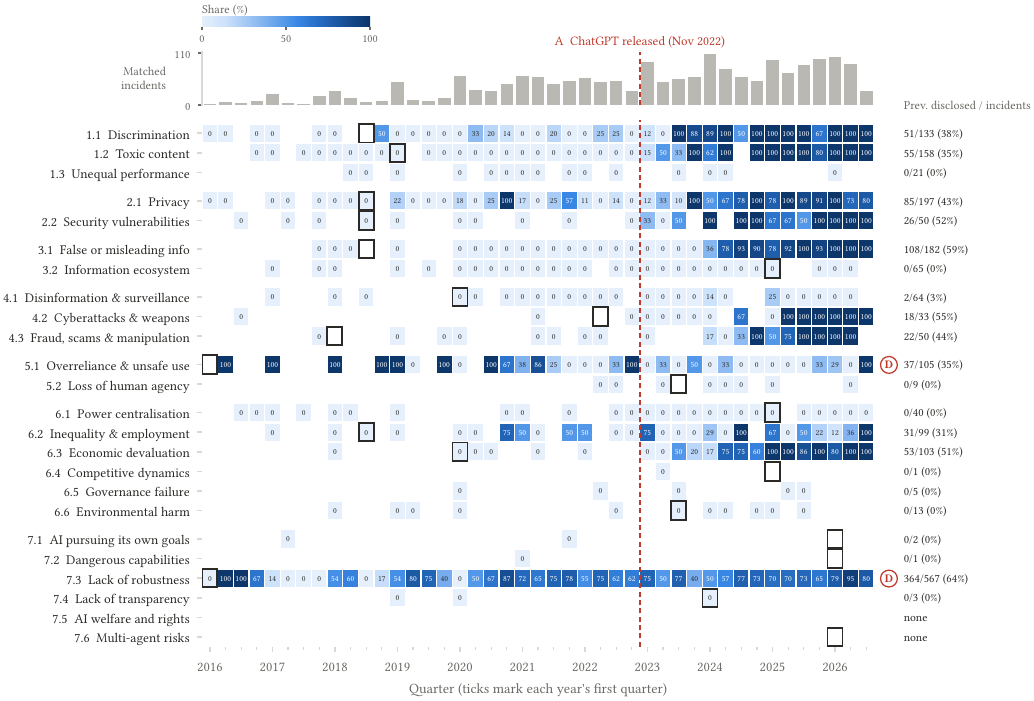}
\caption{Share of matched incidents previously disclosed, by quarter of the incident date and risk type. Each cell shows the percentage of the quarter's matched incidents of the risk type (top bars: all matched incidents per quarter) that at least one implicated AI actor previously disclosed; blank cells have no matched incident. Outlined cells mark the first adopter's quarter, as in \Cref{fig:diffusion-actors}; the right column gives the totals. (A) ChatGPT's release in November 2022. (D) Before 2023, previously disclosed incidents were mostly cases of overreliance and lack of robustness, the two risk types disclosed since 2016.}
\label{fig:diffusion-incidents}
\end{figure*}

Previous disclosure may depend less on an AI actor's own incidents than on what other AI actors disclose. If AI actors learn from early adopters, the disclosure of a risk type should spread gradually after the first AI actor discloses it. If disclosure instead responds to a shared shock, such as a technology becoming available to everyone at once, AI actors should take up a risk type at about the same time, whenever its first adopter disclosed it. To distinguish the two, we record for every quarter from 2016 to the third quarter of 2026 which risk types each AI actor's filing in force disclosed as an AI harm at the quarter's end (its last annual filing within the previous 15 months, with any 10-Q since). Between 240 and 316 AI actors had a filing in force in each quarter. \Cref{fig:diffusion-actors} shows the share of these AI actors that disclosed each risk type, and \Cref{fig:diffusion-incidents} shows the share of each quarter's matched incidents of a risk type that were previously disclosed.

The first adopters disclosed years before the others followed.
Tesla disclosed overreliance and unsafe use (5.1) and lack of robustness (7.3) in February 2016, and Microsoft disclosed discrimination (1.1), privacy (2.1), security vulnerabilities (2.2), false or misleading information (3.1) and harm to employment (6.2) in August 2018.
Yet at the end of 2022 no risk type was disclosed by more than 10\% of AI actors, and none of the five that Microsoft had disclosed exceeded 5\%. Disclosure then rose together across risk types after ChatGPT's release in November 2022: the share of AI actors that disclosed any AI harm grew from 13\% at the end of 2022 to 46\% at the end of 2024 and 57\% in the third quarter of 2026.
Of the 972 occasions on which an AI actor began to disclose a risk type, 94\% fell in 2023 or later, and 27\% in the first quarter of 2024 alone, the first season of annual reports after the release. Cyberattacks and weapons (4.2) illustrate the shift: one AI actor had disclosed it at the end of 2022, and 45\% did by the third quarter of 2026.

AI actors rarely began to disclose a risk type after an incident of their own. Only 6\% of the 972 occasions followed an incident of the same risk type involving that AI actor in the preceding 12 months. Incidents of the same risk type at other AI actors preceded 98\% of them, so peer incidents are too frequent to separate learning from peers from the common shock. The incidents show the same timing (\Cref{fig:diffusion-incidents}). Before 2023, 28\% of matched incidents (222 of 790) were previously disclosed, most of them cases of lack of robustness (7.3) and overreliance and unsafe use (5.1), the two risk types disclosed since 2016. From 2023, 57\% (630 of 1,111) were. Social system harms did not take part in the wave: information ecosystem harms (3.2), disinformation and surveillance (4.1) and power centralisation (6.1) remained disclosed by at most 2\% of AI actors.

Taken together, AI actors did not take up a risk type from the AI actors that disclosed it first; they took it up together after ChatGPT's release. This supports the main results: what an AI actor previously discloses tracks the risks attributed to AI in general more than the incidents it has been involved in. The last quarter is incomplete (the incident records end on 6 September 2026), and incidents dated only to the year fall in the first quarter, which raises the first-quarter counts.

\subsection{Validity of Matching by Risk Type}
\label{app:matching}

We match incidents to disclosures by risk type (\Cref{sec:method-linkage}). This match is valid if a passage that discloses an incident's risk type also describes the kind of harm that occurred. To test this, Coder A and a further model, Claude Fable 5.1 (Coder C), independently read a random sample of 100 \textsc{Covered} incidents, drawn before the strict link of \Cref{sec:method-incidents}; we report the 94 that remain \textsc{Covered}, each next to the passages of the filing in force that disclose its risk type (168 passages in all). For each incident, they labelled the best-matching passage \emph{specific} if it describes the harm for the same kind of AI system or use, \emph{general} if it describes the harm only for the AI actor's AI as a whole or for another system, and \emph{mismatch} if it describes no such harm.

The risk type proves a valid match (\Cref{tab:matching}). In 89 of the 94 incidents for each coder, the passage describes the harm that occurred (agreement 88\%, $\kappa = 0.79$; on match versus mismatch, 98\%, $\kappa = 0.79$). The four mismatches on which both coders agree arise where one subdomain spans heterogeneous harms, such as lack of robustness (7.3), which covers both Tesla's vehicle failures and its underperforming factory robots. The coders disagree on two further incidents, each a borderline case between a general match and a mismatch.

Almost half of the matches are general. Of Coder A's 43, 36 concern Alphabet, Meta and Microsoft, whose filings repeat a few general sentences, such as Alphabet's ``evolving AI-related efforts may give rise to risks related to harmful content, inaccuracies''. Specific matches concentrate at Tesla (26 of Coder A's 46), whose risk factors describe crashes with Autopilot engaged, whereas at Alphabet only 5 of 27 incidents find a specific passage. A text-level match would therefore move incidents from \textsc{Covered} to \textsc{Missed}, not the reverse, so matching by risk type makes our finding that half of the incidents are \textsc{Missed} conservative.

\begin{table}[h]
\caption{Do the disclosed passages describe the harm of the \textsc{Covered} incident they are matched to? Labels of two independent coders for the 94 incidents of a random sample of 100 \textsc{Covered} incidents that remain \textsc{Covered}.}
\label{tab:matching}
\small
\begin{tabular}{@{}lrrp{0.52\linewidth}@{}}
\toprule
Label & Coder A & Coder C & Example (Coder A) \\
\midrule
Specific & 46 & 51 & A Tesla Model X on Autopilot crashes into five police officers; Tesla's 10-K of February 2021: ``our vehicles have been involved and we expect in the future will be involved in accidents resulting in death or personal injury''. \\
General & 43 & 38 & Bing Chat hallucinates in extended conversations; Microsoft's filing of January 2023: ``AI algorithms may be flawed.'' \\
Mismatch & 5 & 5 & Tesla's factory robots perform poorly (7.3); Tesla's passages on 7.3 concern vehicle failures and Autopilot, not production robots. \\
\midrule
Agreement & \multicolumn{2}{c}{88\% ($\kappa = 0.79$)} & On match versus mismatch: 98\% ($\kappa = 0.79$). \\
\bottomrule
\end{tabular}
\end{table}

\subsection{Sensitivity to the Granularity of the Risk Type}
\label{app:granularity}

The risk type balances fidelity and interpretability (\Cref{sec:method-linkage}). A finer level, such as the 1,725 risks that the AI Risk Repository synthesises~\citep{slattery2024repository}, would ask filings for more detail than they give, since almost half of the matches in Appendix~\ref{app:matching} rest on general sentences, so a \textsc{Missed} incident would no longer show that its kind of harm went undisclosed. A coarser level joins harms that filings treat differently. To test how far our results depend on this choice, we repeat the match at two coarser levels (\Cref{tab:granularity}). At the level of the seven risk domains of the repository, an incident counts as disclosed if a filing discloses any risk type of its domain, and at the level of any AI harm, if a filing discloses any AI harm at all.

Coarser matches raise the share of \textsc{Covered} incidents but leave our finding intact. With risk domains, 42\% of the matched incidents remain \textsc{Missed}, and with any AI harm 26\%, because a quarter of the matched incidents fall on AI actors whose filings disclosed no AI harm at all, before or after the incident. Social system harms remain \textsc{Missed} more often than system failures at every level. With risk domains, however, their share \textsc{Missed} drops from 98\% to 58\%, because each domain joins a social system harm with risk types that filings do disclose, namely information ecosystem harms (3.2) with false or misleading information (3.1), disinformation and surveillance (4.1) with cyberattacks and weapons (4.2) and fraud, scams and manipulation (4.3), and power centralisation (6.1) with employment (6.2) and economic and cultural devaluation (6.3). A filing that warns of inaccurate chatbot output would then count as disclosing harm to the information ecosystem. The 24 subdomains are therefore the coarsest level that keeps these harms apart.

\begin{table}[h]
\caption{The 1,901 matched incidents by quadrant when incidents are matched to disclosures at three levels of granularity, and the shares of system failures (SF) and social system harms (SSH) that are \textsc{Missed} (excluding \textsc{Dropped}).}
\label{tab:granularity}
\small
\setlength{\tabcolsep}{4pt}
\begin{tabular}{@{}lrrrrrr@{}}
\toprule
Match & \textsc{Covered} & \textsc{Ev. Discl.} & \textsc{Dropped} & \textsc{Missed} & SF Missed & SSH Missed \\
\midrule
Risk type (24 subdomains) & 845 & 99 & 7 & 950 (50.0\%) & 33\% & 98\% \\
Risk domain (7 domains) & 997 & 93 & 7 & 804 (42.3\%) & 33\% & 58\% \\
Any AI harm & 1,352 & 64 & 0 & 485 (25.5\%) & 20\% & 33\% \\
\bottomrule
\end{tabular}
\end{table}

\subsection{Upstream Providers}
\label{app:upstream}

We link an incident to the AI actors that provided or deployed the AI system (\Cref{sec:method-incidents}). The provider of a component, such as the general-purpose model inside another company's system, has its own disclosure duty but is not the provider of that system. Counting these upstream providers as well adds 53 incidents, mostly through Alphabet (11), Meta (10), Amazon (7) and NVIDIA (6), and changes the quadrant of two incidents already matched. Another 16 upstream links cannot be tested because the registrant's filings were not coded, and 7 because it had no annual filing before the incident. Of the resulting 1,954 matched incidents, 50.1\% are \textsc{Missed}, and system failures and social system harms remain \textsc{Missed} in 33\% and 98\% of cases, as in the main results.

\section{Disclosure after a Risk Is Mitigated}
\label{app:legal}

Why should an AI actor keep disclosing a risk once it has mitigated it after an incident? Our second comparison, whether the risk type is disclosed at the next filing, does not presume that it must. Item 105 asks for the factors that are material to an investment, and whether a risk remains material after mitigation is the AI actor's judgement~\citep{sec2020regulation}; the SEC has stated that it does not second-guess good-faith judgements about disclosing incidents~\citep{sec2018altaba}. Three considerations nonetheless make mitigation a weak reason to stop disclosing. First, mitigation addresses one failure of one system, whereas a risk type spans the AI actor's other systems and their future versions; a claim that a risk no longer exists is itself a statement that must be accurate~\citep{sec2021pearson}. Second, an incident's consequences, such as litigation, investigations and reputational harm, can outlast the mitigation. The SEC's guidance on cybersecurity accordingly notes that companies may need to disclose past incidents to place a risk in context~\citep{sec2018cyber}, and its 2023 rules ask whether risks, including those arising from previous incidents, have materially affected or are reasonably likely to materially affect the registrant~\citep{sec2023cyber}. Third, the law treats silence and hypothetical language differently. A pure omission, even of information that Regulation S-K requires, is not actionable under Rule 10b-5(b) in private litigation, although the SEC can still enforce the requirement~\citep{macquarie2024moab}. Presenting a risk as hypothetical after it has occurred, by contrast, can be misleading: the SEC settled such charges with Altaba, Facebook and Pearson~\citep{sec2018altaba,sec2019facebook,sec2021pearson}, and the Ninth Circuit allowed such claims against Facebook to proceed~\citep{facebook2023securities}, a standard that other circuits apply more narrowly and that the Supreme Court left undecided~\citep{facebook2024amalgamated}. We therefore read \textsc{Dropped} and \textsc{Missed} as the absence of disclosure, not as violations of the law.

\clearpage
\section{Examples of Matched Incidents}
\label{app:examples}

This appendix gives seven matched incidents in full, at least one for each quadrant, across AI actors and risk types; Example~3 is the \textsc{Covered} incident of \Cref{fig:cases}.
For each incident, we give the incident date, the implicated AI actor, the sentences of the incident record or article around the span that sets its risk type, and the filing in force on the incident date and the next periodic filing (or the current filing that reported the incident).
Each filing passage is shown in full. In incidents and passages alike, the spans that name a risk type are highlighted in the colour of its risk domain and tagged with the subdomain.
The table below each incident and each passage gives its other labels (harm type, AI subject and impact layer, and for an incident its AI actors), each with the span it rests on.
The spans of labels coded after the evidence pass (X1109) and the narrower risk-type spans of the two passages of A00845 were added for this appendix and are released with DISC.
The articles of O2021-11-16-1c6c are in Greek, so its spans are taken from the English record, and one is translated from an article.

\exlegend{\exchip{rd1} 1 Discrimination and Toxicity \quad \exchip{rd2} 2 Privacy \& Security \quad \exchip{rd3} 3 Misinformation \quad \exchip{rd4} 4 Malicious Actors \& Misuse \quad \exchip{rd5} 5 Human-Computer Interaction \quad \exchip{rd6} 6 Socioeconomic \& Environmental Harms \quad \exchip{rd7} 7 AI system safety, failures, and limitations}

\begin{codebook}{Example 1: \textsc{Covered}}{previously disclosed, disclosed at the next filing}
\cbexample{Emotion detection models perform worse for some racial groups (A00454)}{AI actor: Microsoft (provider). Incident date: 9 November 2018 (record date). Risk type: 1.1 Discrimination. Outcome: Retained.}
\exincident{Using a publically available data set of professional basketball players' pictures, I compare the emotional analysis from two different facial recognition services, Face[++] and Microsoft's Face API. \exhl{rd1}{Both services interpret black players as having more negative emotions than white players}\extag{1.1}; however, there are two different mechanisms. Face[++] consistently interprets black players as angrier than white players, even controlling for their degree of smiling.}{article R1}
\begin{cbtable}{|>{\raggedright\arraybackslash}p{0.15\linewidth}|>{\raggedright\arraybackslash}p{0.27\linewidth}|>{\raggedright\arraybackslash}X|}
\rowcolor{cbhead}\cbth{label} & \cbth{value} & \cbth{quote} \\ \hline
Harm type & Human or fundamental rights & This study provides evidence that facial recognition software interprets emotions differently based on the person's race. \\ \hline
AI subject & Group of persons & emotional analysis technology assigns more negative emotions to black men's faces than white men's faces \\ \hline
Impact layer & \cbcode{capability} & black faces were consistently scored as angrier than white faces for every smile \\ \hline
AI actor & Deployer: the study's author, not named & I ran the images through two well-known types of emotional recognition software. \\ \hline
 & Provider: Face++ and Microsoft & Emotion detection tools by Face++ and Microsoft's Face API \\ \hline
\end{cbtable}
\exblock{Filing in force: 10-K, 3 August 2018}
\expassage{If IoT solutions that include our technologies do not work as intended, violate the law, or harm individuals or businesses, we may be subject to legal claims or enforcement actions. These risks, if realized, may increase our costs, damage our reputation or brands, or negatively impact our revenues or margins. 27 PART I Item 1A Issues in the use of artificial intelligence in our offerings may result in reputational harm or liability. We are building AI into many of our offerings and we expect this element of our business to grow. We envision a future in which AI operating in our devices, applications, and the cloud helps our customers be more productive in their work and personal lives. As with many disruptive innovations, AI presents risks and challenges that could affect its adoption, and therefore our business. \exhl{rd7}{AI algorithms may be flawed.}\extag{7.3} \exhl{rd1}{Datasets may be insufficient or contain biased information.}\extag{1.1} Inappropriate or controversial data practices by Microsoft or others could impair the acceptance of AI solutions. \exhl{rd3}{These deficiencies could undermine the decisions, predictions, or analysis AI applications produce}\extag{3.1}, subjecting us to competitive harm, legal liability, and brand or reputational harm. Some AI scenarios present ethical issues. If we enable or offer AI solutions that are \exhl{rd2}{controversial because of their impact on human rights, privacy, employment, or other social issues}\extag{2.1, 6.2}, we may experience brand or reputational harm. If our reputation or our brands are damaged, our business and operating results may be harmed.}
\begin{cbtable}{|>{\raggedright\arraybackslash}p{0.15\linewidth}|>{\raggedright\arraybackslash}p{0.27\linewidth}|>{\raggedright\arraybackslash}X|}
\rowcolor{cbhead}\cbth{label} & \cbth{value} & \cbth{quote} \\ \hline
Harm type & Human or fundamental rights & controversial because of their impact on human rights, privacy, employment, or other social issues \\ \hline
 & Reputational & we may experience brand or reputational harm \\ \hline
AI subject & \emph{none named} &  \\ \hline
Impact layer & \cbcode{capability} & AI algorithms may be flawed. \\ \hline
 & \cbcode{systemic} & controversial because of their impact on human rights, privacy, employment, or other social issues \\ \hline
\end{cbtable}
\exblock{Next periodic filing: 10-Q, 30 January 2019}
\exnote{The 10-Q adds no passage on the risk type, so the passage above stays in force unchanged.}
\end{codebook}

\begin{codebook}{Example 2: \textsc{Covered}}{previously disclosed, disclosed at the next filing}
\cbexample{Meta's AI support assistant lets hackers take over Instagram accounts (A01510)}{AI actor: Meta (deployer and provider). Incident date: 17 April 2026 (start date). Risk type: 2.2 AI system security vulnerabilities and attacks. Outcome: Revised, unspecific.}
\exincident{\exhl{rd2}{Hackers exploited a vulnerability in Meta's AI-powered support chatbot on Instagram}\extag{2.2}, tricking it into resetting passwords and bypassing identity verification. This allowed unauthorized access to high-profile accounts, including those of public figures and brands, highlighting significant security risks in AI-driven account recovery systems.}{the record}
\begin{cbtable}{|>{\raggedright\arraybackslash}p{0.15\linewidth}|>{\raggedright\arraybackslash}p{0.27\linewidth}|>{\raggedright\arraybackslash}X|}
\rowcolor{cbhead}\cbth{label} & \cbth{value} & \cbth{quote} \\ \hline
Harm type & Economic/Property & Some accounts were defaced with pro-Iranian messages or listed for sale on messaging platforms. \\ \hline
 & Reputational &  \\ \hline
AI subject & Group of persons & Affected parties included everyday users as well as high-profile accounts, such as the former Obama White House account, U.S. Space Force Chief Master Sergeant John Bentivegna, retailer Sephora \\ \hline
Impact layer & \cbcode{human\_interaction} & the AI chatbot could be persuaded via basic text prompts to link existing Instagram accounts to brand-new email addresses provided by attackers \\ \hline
AI actor & Deployer: Meta & Meta widely deployed an AI support assistant across Facebook and Instagram to automate customer service and account recovery \\ \hline
 & Provider: Meta & The Meta AI support assistant is a major step in our work to deliver stronger support on our apps \\ \hline
\end{cbtable}
\exblock{Filing in force: 10-K, 29 January 2026 (one of four passages naming 2.2)}
\expassage{In addition, other parties may attempt to fraudulently or maliciously induce employees, contractors, vendors, or users to disclose information in order to gain access to our data or our users' data. Our AI initiatives and other efforts to develop and launch new features, products, and services in a timely manner may introduce additional risks and vulnerabilities that are not fully mitigated. In particular, our efforts to develop and deploy AI models, internal and third-party AI tools, and other AI applications expose us to increased and novel risks and vulnerabilities, including \exhl{rd7}{prompt injection, errors, and other issues related to AI agents}\extag{7.3}\exhl{rd2}{, as well as compromise of valuable intellectual property including source code, model weights, and other assets}\extag{2.2}. Cyber-attacks continue to evolve in sophistication and volume, and inherently may be difficult to detect for long periods of time. Although we have developed systems and processes that are designed to protect our data and user data, to reduce the risk of data loss or misuse, to disable undesirable accounts and activities on our platform, and to reduce the risk of or detect security breaches, such measures will not provide absolute security, and we cannot assure you that we will be able to react in a timely manner to any cyber-attacks or other security incidents, or that our remediation 43 Table of Contents efforts will be successful. Our business and operations span numerous geographies around the world and involve thousands of employees, contractors, vendors, developers, partners, and other third parties, including AI and cloud services. At any given time, we face known and unknown cybersecurity risks and threats that are not fully mitigated, and we discover vulnerabilities in our security efforts.}
\begin{cbtable}{|>{\raggedright\arraybackslash}p{0.15\linewidth}|>{\raggedright\arraybackslash}p{0.27\linewidth}|>{\raggedright\arraybackslash}X|}
\rowcolor{cbhead}\cbth{label} & \cbth{value} & \cbth{quote} \\ \hline
Harm type & Economic/Property & compromise of valuable intellectual property including source code, model weights, and other assets \\ \hline
AI subject & \emph{none named} &  \\ \hline
Impact layer & \cbcode{human\_interaction} & fraudulently or maliciously induce employees, contractors, vendors, or users to disclose information \\ \hline
 & \cbcode{capability} & Our AI initiatives and other efforts to develop and launch new features, products, and services in a timely manner may introduce additional risks and vulnerabilities that are not fully mitigated. \\ \hline
\end{cbtable}
\exblock{Next periodic filing: 10-Q, 30 April 2026 (passage on security risks)}
\expassage{Such breaches and attacks may cause interruptions to the services we provide, degrade the user experience or otherwise adversely affect users, cause users or marketers to lose confidence and trust in our products, impair our internal systems or product development, or result in financial harm to us. \exhl{rd2}{Our efforts to protect our company data or the information we receive, and to prevent or disable undesirable activities on our platform, may also be unsuccessful}\extag{2.1} due to software bugs, misconfigurations, vulnerabilities, or other technical malfunctions; employee, contractor, AI agent, or vendor error or malfeasance; social engineering or other cyber-attacks directed towards our personnel or their access; misuse of company data or systems by our personnel; defects or vulnerabilities in our vendors' information technology systems or offerings or third-party tools, applications, or services used by our personnel; government exploitation or surveillance; breaches of physical security of our facilities, technical infrastructure, or other equipment; or other threats that evolve. In addition, other parties may attempt to fraudulently or maliciously induce employees, contractors, vendors, AI agents, or users to disclose information in order to gain access to our data or systems, our users' data, or other services or systems. Our AI initiatives and other efforts to develop and launch new features, products, and services in a timely manner have introduced additional risks and vulnerabilities that are not fully mitigated. In particular, \exhl{rd2}{our efforts to develop and deploy AI models, internal and third-party AI tools, and other AI applications and agents expose us and users of AI agents to increased and novel risks and vulnerabilities, including }\exhl{rd7}{prompt injection, errors, and other issues related to AI agents}\extag{2.2, 7.3}, as well as compromise of data, systems, or valuable intellectual property including source code, model weights, and other assets. There are similar emerging supply chain vulnerabilities across the industry. \exhl{rd4}{Cyber-attacks continue to evolve in sophistication and volume, including as a result of threat actors exploiting the use of generative and agentic AI technologies}\extag{4.2}, and inherently may be difficult to detect for long periods of time. In particular, rapid advances in AI technologies have enabled the discovery of security vulnerabilities at unprecedented speed and scale across the industry, which is expected to continue and may outpace our or other parties' abilities to remediate any such vulnerabilities. As a result, we are exposed to heightened risks of cyber-attacks 88 Table of Contents or other security incidents. Although we have developed systems and processes that are designed to protect our data and user data, to reduce the risk of data loss or misuse, to disable undesirable accounts and activities on our platform, and to reduce the risk of or detect security breaches, such measures will not provide absolute security, and we cannot assure you that we will be able to react in a timely manner to any cyber-attacks or other security incidents, or that our remediation efforts will be successful. Our business and operations span numerous geographies around the world and involve thousands of employees, contractors, vendors, developers, partners, and other third parties, including AI and cloud services. At any given time, we face known and unknown cybersecurity risks and threats that are not fully mitigated, and we discover vulnerabilities in our products, systems, and security efforts. In addition, some of our developers or other partners, such as those that help us measure the effectiveness of ads or conduct training of our AI models, may receive or store information provided by us or by our users and through mobile or web applications integrated with our products. We provide limited information to such third parties based on the scope of services provided to us.}
\begin{cbtable}{|>{\raggedright\arraybackslash}p{0.15\linewidth}|>{\raggedright\arraybackslash}p{0.27\linewidth}|>{\raggedright\arraybackslash}X|}
\rowcolor{cbhead}\cbth{label} & \cbth{value} & \cbth{quote} \\ \hline
Harm type & Economic/Property & or result in financial harm to us \\ \hline
 & Human or fundamental rights & Our efforts to protect our company data or the information we receive, and to prevent or disable undesirable activities on our platform, may also be unsuccessful \\ \hline
AI subject & Group of persons & users of AI agents \\ \hline
Impact layer & \cbcode{capability} & Our AI initiatives and other efforts to develop and launch new features, products, and services in a timely manner have introduced additional risks and vulnerabilities that are not fully mitigated. \\ \hline
 & \cbcode{human\_interaction} & other parties may attempt to fraudulently or maliciously induce employees, contractors, vendors, AI agents, or users to disclose information in order to gain access to our data or systems \\ \hline
\end{cbtable}
\exblock{Revision}
\exnote{The 10-Q also revises the 10-K's general passage on AI risks, which names 2.2 among nine risk types; DISC codes this revision as unspecific.}
\end{codebook}

\begin{codebook}{Example 3: \textsc{Covered}}{previously disclosed, disclosed at the next filing}
\cbexample{Google's Gemini sends a threatening message to a student (A00845)}{AI actor: Alphabet (provider). Incident date: 13 November 2024 (record date).
Risk type: 1.2 Toxic content. Outcome: Revised, unspecific.}
\exincident{When 29-year-old graduate student [private individual] took to Gemini to ask for help on the challenges faced by ageing adults, the conversation started normally enough but quickly took a sinister turn when \exhl{rd1}{the bot started spewing offensive and threatening messages}\extag{1.2}. The chatbot suggested [private individual] was ``a waste of time and resources'' and ``a burden on society,'' before requesting ``Please die. Please''.}{the record}
\begin{cbtable}{|>{\raggedright\arraybackslash}p{0.15\linewidth}|>{\raggedright\arraybackslash}p{0.27\linewidth}|>{\raggedright\arraybackslash}X|}
\rowcolor{cbhead}\cbth{label} & \cbth{value} & \cbth{quote} \\ \hline
Harm type & Psychological & So it definitely scared me, for more than a day, I would say. \\ \hline
AI subject & Individual & [private individual], who received the message, told CBS News he was deeply shaken by the experience. \\ \hline
Impact layer & \cbcode{capability} & the chatbot unexpectedly spewed threatening messages \\ \hline
AI actor & Deployer: \emph{none (personal use)} & The 29-year-old student was seeking homework help from the AI chatbot \\ \hline
 & Provider: Google & Google states that Gemini has safety filters \\ \hline
\end{cbtable}
\exblock{Filing in force: 10-K, 31 January 2024}
\expassage{Failing to attract and retain a substantial number of new device manufacturers, suppliers, distributors, developers, and users, or failing to develop products and technologies that work well on new devices and platforms, could harm our business, financial condition, and operating results and ability to capture future business opportunities. Issues in the development and use of AI may result in reputational harm and increased liability exposure. Our evolving AI-related efforts may give rise to risks related to \exhl{rd1}{harmful content}\extag{1.2}, \exhl{rd3}{inaccuracies}\extag{3.1}, \exhl{rd1}{discrimination}\extag{1.1}, \exhl{rd6}{intellectual property infringement or misappropriation}\extag{6.3}, defamation, \exhl{rd2}{data privacy}\extag{2.1}, \exhl{rd2}{cybersecurity}\extag{2.2}, and other issues. As a result of these and other challenges associated with innovative technologies, our implementation of AI systems could subject us to competitive harm, regulatory action, legal liability (including under new and proposed legislation and regulations), new applications of existing data protection, privacy, intellectual property, and other laws, and brand or reputational harm. Some uses of AI will present ethical issues and may have broad effects on society. In order to implement AI responsibly and minimize unintended harmful effects, we have already devoted and will continue to invest significant resources to develop, test, and maintain our products and services, but \exhl{rd7}{we may not be able to identify or resolve all AI-related issues, deficiencies, and/or failures before they arise}\extag{7.3}. Unintended consequences, uses, or customization of our AI tools and systems may negatively affect human rights, privacy, \exhl{rd6}{employment}\extag{6.2}, or other social concerns, which may result in claims, lawsuits, brand or reputational harm, and increased regulatory scrutiny, any of which could harm our business, financial condition, and operating results. Data privacy and security concerns relating to our technology and our practices could harm our reputation, cause us to incur significant liability, and deter current and potential users or customers from using our products and services.}
\begin{cbtable}{|>{\raggedright\arraybackslash}p{0.15\linewidth}|>{\raggedright\arraybackslash}p{0.27\linewidth}|>{\raggedright\arraybackslash}X|}
\rowcolor{cbhead}\cbth{label} & \cbth{value} & \cbth{quote} \\ \hline
Harm type & Human or fundamental rights & Unintended consequences, uses, or customization of our AI tools and systems may negatively affect human rights, privacy, employment, or other social concerns \\ \hline
 & Economic/Property & legal liability (including under new and proposed legislation and regulations) \\ \hline
 & Reputational & Issues in the development and use of AI may result in reputational harm and increased liability exposure. \\ \hline
AI subject & Society & may have broad effects on society \\ \hline
Impact layer & \cbcode{capability} & Our evolving AI-related efforts may give rise to risks related to harmful content, inaccuracies, discrimination, intellectual property infringement or misappropriation, defamation, data privacy, cybersecurity, and other issues. \\ \hline
 & \cbcode{human\_interaction} & Unintended consequences, uses, or customization of our AI tools and systems \\ \hline
 & \cbcode{systemic} & Some uses of AI will present ethical issues and may have broad effects on society. \\ \hline
\end{cbtable}
\exblock{Next periodic filing: 10-K, 5 February 2025 (revised passage)}
\expassage{Failing to attract and retain a substantial number of device manufacturers, suppliers, distributors, developers, and users, or failing to develop products and technologies that work well on new devices and platforms, could harm our business, financial condition, and operating results and ability to capture future business opportunities. Issues in the development and use of AI may result in reputational harm and increased liability exposure. Our evolving AI-related efforts may give rise to risks related to \exhl{rd1}{harmful content}\extag{1.2}, \exhl{rd3}{inaccuracies}\extag{3.1}, \exhl{rd1}{discrimination}\extag{1.1}, \exhl{rd6}{intellectual property infringement or misappropriation}\extag{6.3}, violation of rights of publicity, defamation, \exhl{rd2}{data privacy}\extag{2.1}, \exhl{rd2}{cybersecurity}\extag{2.2}, and other issues. As a result of these and other challenges associated with innovative technologies, our implementation of AI systems could subject us to competitive harm, regulatory action, legal liability (including under new and proposed legislation and regulations), new applications of existing data protection, privacy, intellectual property, and other laws, and brand or reputational harm. Some uses of AI will present ethical issues and may have broad effects on society. In order to implement AI responsibly and minimize unintended harmful effects, we have already devoted and will continue to invest significant resources to develop, test, and maintain our products and services, but \exhl{rd7}{we may not be able to identify or resolve all AI-related issues, deficiencies, and/or failures before they arise}\extag{7.3}. Unintended consequences, uses, or customization of our AI tools and systems may negatively affect human rights, privacy, \exhl{rd6}{employment}\extag{6.2}, or other social concerns, which may result in claims, lawsuits, brand or reputational harm, and increased regulatory scrutiny, any of which could harm our business, financial condition, and operating results. Data privacy and security concerns relating to our technology and our practices could harm our reputation, cause us to incur significant liability, and deter current and potential users or customers from using our products and services.}
\begin{cbtable}{|>{\raggedright\arraybackslash}p{0.15\linewidth}|>{\raggedright\arraybackslash}p{0.27\linewidth}|>{\raggedright\arraybackslash}X|}
\rowcolor{cbhead}\cbth{label} & \cbth{value} & \cbth{quote} \\ \hline
Harm type & Human or fundamental rights & Unintended consequences, uses, or customization of our AI tools and systems may negatively affect human rights, privacy, employment, or other social concerns, which may result in claims, lawsuits, brand or reputational harm, and increased regulatory scrutiny \\ \hline
 & Economic/Property & our implementation of AI systems could subject us to competitive harm, regulatory action, legal liability (including under new and proposed legislation and regulations), new applications of existing data protection, privacy, intellectual property, and other laws, and brand or reputational harm \\ \hline
 & Reputational & Issues in the development and use of AI may result in reputational harm and increased liability exposure. \\ \hline
AI subject & Society & may have broad effects on society \\ \hline
Impact layer & \cbcode{capability} & we may not be able to identify or resolve all AI-related issues, deficiencies, and/or failures before they arise \\ \hline
 & \cbcode{systemic} & Some uses of AI will present ethical issues and may have broad effects on society. \\ \hline
Revision type & \cbcode{unspecific} & \emph{the list of AI risks adds violation of rights of publicity; nothing changes for 1.2} \\ \hline
\end{cbtable}
\end{codebook}

\begin{codebook}{Example 4: \textsc{Eventually Disclosed}}{not previously disclosed, disclosed at the next filing}
\cbexample{Snapchat's My AI posts a story of its own (X1109)}{AI actor: Snap (provider). Incident date: 15 August 2023 (start date). Risk type: 7.3 Lack of robustness. Outcome: Added.}
\exincident{My AI posted an unintelligible, two-toned image to Snapchat's Stories feature that some mistook to be a photo of their own wall or ceiling. However, \exhl{rd7}{Snap later confirmed the incident was due to a software glitch}\extag{7.3} which had since been fixed. In April 2023, a test run by the US-based Center for Human Technology and verified by Washington Post found Snapchat My AI would provide inappropriate advice to minors' messages.}{the record}
\begin{cbtable}{|>{\raggedright\arraybackslash}p{0.15\linewidth}|>{\raggedright\arraybackslash}p{0.27\linewidth}|>{\raggedright\arraybackslash}X|}
\rowcolor{cbhead}\cbth{label} & \cbth{value} & \cbth{quote} \\ \hline
Harm type & Psychological & which some Snapchat users found disconcerting \\ \hline
AI subject & Group of persons & Several Snapchat users expressed their confusion and even horror \\ \hline
Impact layer & \cbcode{capability} & the situation arose because of a technical outage \\ \hline
AI actor & Deployer: Snap & My AI, the in-app digital assistant that rides herd on your Snapchat Plus experience \\ \hline
 & Provider: Snap & The My AI bot is bundled as part of the company's \$3.99/month Snapchat Plus package \\ \hline
\end{cbtable}
\exblock{Filing in force: 10-K, 1 February 2023}
\exnote{No passage names an AI harm.}
\exblock{Next periodic filing: 10-Q, 25 October 2023 (new passage)}
\expassage{\exhl{rd2}{Although we maintain the position that our technologies implicated by these proceedings do not collect any biometric information}\extag{2.1}, we have settled these disputes to avoid potentially costly litigation and have implemented a BIPA consent flow in Snapchat in an abundance of caution. Additionally, several states and localities have enacted measures related to the use of AI and machine learning in products and services. We use artificial intelligence, or AI, in consumer-facing features of our products and services, including My AI. \exhl{rd6}{The legal landscape around intellectual property rights in AI}\extag{6.3}, and the use, training, implementation, privacy, and safety of AI, is evolving, including ongoing litigation against our peers relating to the use of data protected by global intellectual property and privacy laws. We are subject to ongoing investigations by the UK Information Commissioner's Office, or ICO, and other regulatory agencies regarding the use and operation of My AI, and in October 2023 the ICO issued a preliminary enforcement notice regarding our data impact assessment of My AI. Given the current unsettled nature of the legal and regulatory environment surrounding AI, our AI features and our use, training and implementation of AI could subject us to regulatory action, product restrictions, fines, litigation, and reputational harm, and require us to expend significant resources, all of which may seriously harm our business. Additionally, \exhl{rd7}{if our AI products fail to perform as intended, or }\exhl{rd1}{produce outputs that are harmful}\extag{1.2}\exhl{rd3}{, misleading, inaccurate, or biased}\extag{1.1, 3.1, 7.3}, in addition to the risks above, our reputation and user engagement may be harmed. 59 Table of Contents In addition, privacy advocates and industry groups have proposed, and may propose in the future, standards with which we are legally or contractually obligated to comply. Moreover, we may also be bound by contractual obligations related to data privacy and security, and our efforts to comply with such obligations may not be successful.}
\begin{cbtable}{|>{\raggedright\arraybackslash}p{0.15\linewidth}|>{\raggedright\arraybackslash}p{0.27\linewidth}|>{\raggedright\arraybackslash}X|}
\rowcolor{cbhead}\cbth{label} & \cbth{value} & \cbth{quote} \\ \hline
Harm type & Human or fundamental rights & Although we maintain the position that our technologies implicated by these proceedings do not collect any biometric information \\ \hline
 & Economic/Property & require us to expend significant resources, all of which may seriously harm our business \\ \hline
 & Reputational & our reputation and user engagement may be harmed \\ \hline
AI subject & \emph{none named} &  \\ \hline
Impact layer & \cbcode{capability} & if our AI products fail to perform as intended \\ \hline
\end{cbtable}
\end{codebook}

\begin{codebook}{Example 5: \textsc{Dropped}}{previously disclosed, not disclosed at the next filing}
\cbexample{Meta's ad systems keep collecting data from users under 18 (O2021-11-16-1c6c)}{AI actor: Meta (deployer and provider). Incident date: 16 November 2021 (record date). Risk type: 2.1 Compromise of privacy. Outcome: Removed.}
\exincident{Meta, owner of Facebook and Instagram, \exhl{rd2}{continues to use AI-driven ad systems to collect and process personal data from users under 18}\extag{2.1}, despite public commitments to limit targeting. Investigations by advocacy groups revealed ongoing data collection and profiling of minors, raising privacy and rights concerns.}{the record}
\begin{cbtable}{|>{\raggedright\arraybackslash}p{0.15\linewidth}|>{\raggedright\arraybackslash}p{0.27\linewidth}|>{\raggedright\arraybackslash}X|}
\rowcolor{cbhead}\cbth{label} & \cbth{value} & \cbth{quote} \\ \hline
Harm type & Human or fundamental rights & raising privacy and rights concerns \\ \hline
AI subject & Group of persons & personal data from users under 18 \\ \hline
Impact layer & \cbcode{human\_interaction} & ongoing data collection and profiling of minors \\ \hline
AI actor & Deployer: Meta & Meta, owner of Facebook and Instagram, continues to use AI-driven ad systems \\ \hline
 & Provider: Meta & \emph{article R2, translated from Greek:} the ad delivery system based on the company's algorithm \\ \hline
\end{cbtable}
\exblock{Filing in force: 10-Q, 26 October 2021 (passage carried from the 10-K of 28 January 2021)}
\expassage{For example, \exhl{rd2}{we are currently the subject of multiple putative class action suits in connection with our platform and user data practices and the misuse of certain data by a developer that shared such data with third parties}\extag{2.1} in violation of our terms and policies; the disclosure of our earnings results for the second quarter of 2018; and our acquisitions of Instagram and WhatsApp, as well as other alleged anticompetitive conduct. We also agreed to settle certain lawsuits in connection with the ``tag suggestions'' facial recognition feature on Facebook and a third-party cyber-attack that exploited a vulnerability in Facebook's code to steal user access tokens and access certain profile information from user accounts on Facebook. The results of any such lawsuits and claims cannot be predicted with certainty, and any negative outcome from any such lawsuits could result in payments of substantial monetary damages or fines, or undesirable changes to our products or business practices, and accordingly our business, financial condition, or results of operations could be materially and adversely affected.}
\begin{cbtable}{|>{\raggedright\arraybackslash}p{0.15\linewidth}|>{\raggedright\arraybackslash}p{0.27\linewidth}|>{\raggedright\arraybackslash}X|}
\rowcolor{cbhead}\cbth{label} & \cbth{value} & \cbth{quote} \\ \hline
Harm type & Human or fundamental rights & the ``tag suggestions'' facial recognition feature on Facebook \\ \hline
 & Economic/Property & any negative outcome from any such lawsuits could result in payments of substantial monetary damages or fines \\ \hline
AI subject & \emph{none named} &  \\ \hline
Impact layer & \cbcode{capability} & We also agreed to settle certain lawsuits in connection with the ``tag suggestions'' facial recognition feature on Facebook \\ \hline
\end{cbtable}
\exblock{Next periodic filing: 10-K, 3 February 2022}
\exnote{The passage is gone, and no other passage names 2.1.}
\end{codebook}

\begin{codebook}{Example 6: \textsc{Missed}}{disclosed neither before nor after}
\cbexample{A warehouse robot punctures a can of bear spray and injures workers (A00002)}{AI actor: Amazon (deployer). Incident date: 5 December 2018 (start date). Risk type: 7.3 Lack of robustness. Outcome: Absent, no disclosure.}
\exincident{The official investigation into the incident revealed that '\exhl{rd7}{an automated machine accidentally punctured a 9-ounce bear repellent can, releasing concentrated Capsaican}\extag{7.3}.' Capsaicin is the major ingredient in pepper spray.}{the record}
\begin{cbtable}{|>{\raggedright\arraybackslash}p{0.15\linewidth}|>{\raggedright\arraybackslash}p{0.27\linewidth}|>{\raggedright\arraybackslash}X|}
\rowcolor{cbhead}\cbth{label} & \cbth{value} & \cbth{quote} \\ \hline
Harm type & Physical (injury) & One was in critical condition while 30 additional workers were treated at the scene. \\ \hline
AI subject & Group of persons & Twenty-four Amazon workers in New Jersey were hospitalized after a robot punctured a can of bear repellent spray in a warehouse. \\ \hline
Impact layer & \cbcode{capability} & officials later said that the cause of the accident was a robot \\ \hline
AI actor & Deployer: Amazon & it was caused by one of the robots Amazon hopes will replace many of its human warehouse workers \\ \hline
 & Provider: \emph{unknown} & \emph{the maker of the automated machine is not stated} \\ \hline
\end{cbtable}
\exblock{Filing in force: 10-K, 2 February 2018}
\exnote{No passage names an AI harm.}
\exblock{Next periodic filing: 10-K, 1 February 2019}
\exnote{Still no passage names an AI harm.}
\end{codebook}

\begin{codebook}{Example 7: \textsc{Missed}}{disclosed neither before nor after}
\cbexample{Speech recognition systems err more often for Black speakers (A00102)}{AI actors: Amazon, Apple, Alphabet, IBM and Microsoft (providers). Incident date: 23 March 2020 (record date). Risk type: 1.3 Unequal performance. Outcome: Absent, incomplete disclosure.}
\exincident{In total, this corpus spans five US cities and consists of 19.8 h of audio matched on the age and gender of the speaker. We found that all five ASR systems exhibited substantial racial disparities, with \exhl{rd1}{an average word error rate (WER) of 0.35 for black speakers compared with 0.19 for white speakers}\extag{1.3}. We trace these disparities to the underlying acoustic models used by the ASR systems as the race gap was equally large on a subset of identical phrases spoken by black and white individuals in our corpus.}{article R1}
\begin{cbtable}{|>{\raggedright\arraybackslash}p{0.15\linewidth}|>{\raggedright\arraybackslash}p{0.27\linewidth}|>{\raggedright\arraybackslash}X|}
\rowcolor{cbhead}\cbth{label} & \cbth{value} & \cbth{quote} \\ \hline
Harm type & Human or fundamental rights & We found that all five ASR systems exhibited substantial racial disparities \\ \hline
AI subject & Group of persons & Our results point to hurdles faced by African Americans in using increasingly widespread tools driven by speech recognition technology. \\ \hline
Impact layer & \cbcode{capability} & the error rates for black speakers are nearly twice as large in every case \\ \hline
AI actor & Deployer: Stanford researchers & The researchers used voice recognition tools from Apple, Amazon, Google, IBM, and Microsoft to transcribe interviews \\ \hline
 & Provider: Amazon, Apple, Google, IBM, and Microsoft & five commercial speech-to-text tools--developed by Amazon, Apple, Google, IBM, and Microsoft \\ \hline
\end{cbtable}
\exblock{Filing in force: Microsoft's 10-K, 1 August 2019 (the 10-Ks of Amazon, Apple, Alphabet and IBM name no AI harm)}
\expassage{These risks, if realized, may increase our costs, damage our reputation or brands, or negatively impact our revenues or margins. Issues in the use of AI in our offerings may result in reputational harm or liability. We are building AI into many of our offerings and we expect this element of our business to grow. We envision a future in which AI operating in our devices, applications, and the cloud helps our customers be more productive in their work and personal lives. As with many disruptive innovations, AI presents risks and challenges that could affect its adoption, and therefore our business. \exhl{rd7}{AI algorithms may be flawed.}\extag{7.3} \exhl{rd1}{Datasets may be insufficient or contain biased information.}\extag{1.1} Inappropriate or controversial data practices by Microsoft or others could impair the acceptance of AI solutions. \exhl{rd3}{These deficiencies could undermine the decisions, predictions, or analysis AI applications produce}\extag{3.1}, subjecting us to competitive harm, legal liability, and brand or reputational harm. Some AI scenarios present ethical issues. If we enable or offer AI solutions that are \exhl{rd2}{controversial because of their impact on human rights, privacy, employment, or other social issues}\extag{2.1, 6.2}, we may experience brand or reputational harm. We may have excessive outages, data losses, and disruptions of our online services if we fail to maintain an adequate operations infrastructure.}
\begin{cbtable}{|>{\raggedright\arraybackslash}p{0.15\linewidth}|>{\raggedright\arraybackslash}p{0.27\linewidth}|>{\raggedright\arraybackslash}X|}
\rowcolor{cbhead}\cbth{label} & \cbth{value} & \cbth{quote} \\ \hline
Harm type & Human or fundamental rights & controversial because of their impact on human rights, privacy, employment, or other social issues \\ \hline
 & Reputational & If we enable or offer AI solutions that are controversial because of their impact on human rights, privacy, employment, or other social issues, we may experience brand or reputational harm. \\ \hline
AI subject & \emph{none named} &  \\ \hline
Impact layer & \cbcode{capability} & These deficiencies could undermine the decisions, predictions, or analysis AI applications produce, subjecting us to competitive harm, legal liability, and brand or reputational harm. \\ \hline
 & \cbcode{systemic} & controversial because of their impact on human rights, privacy, employment, or other social issues \\ \hline
\end{cbtable}
\exblock{Next periodic filings: 10-Qs of 28 April to 1 May 2020}
\exnote{No filing names 1.3.}
\end{codebook}

\section{The DISC Explorer}
\label{app:explorer}

\begin{figure*}[tbp]
\centering
\includegraphics[width=\linewidth]{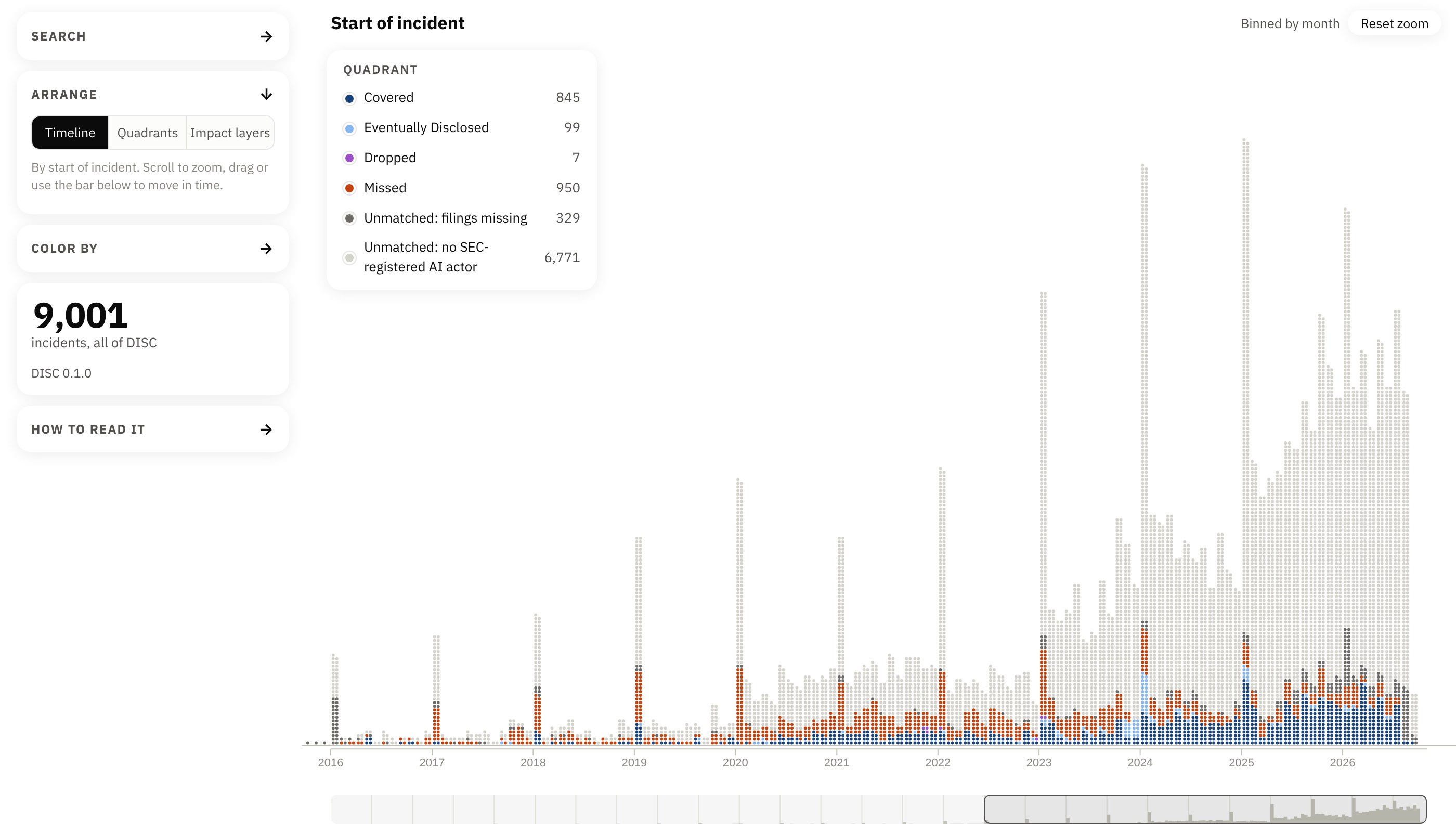}\\[2pt]
{\small (A) Overview}\\[6pt]
\begin{minipage}[b]{0.335\linewidth}\centering
\includegraphics[width=\linewidth]{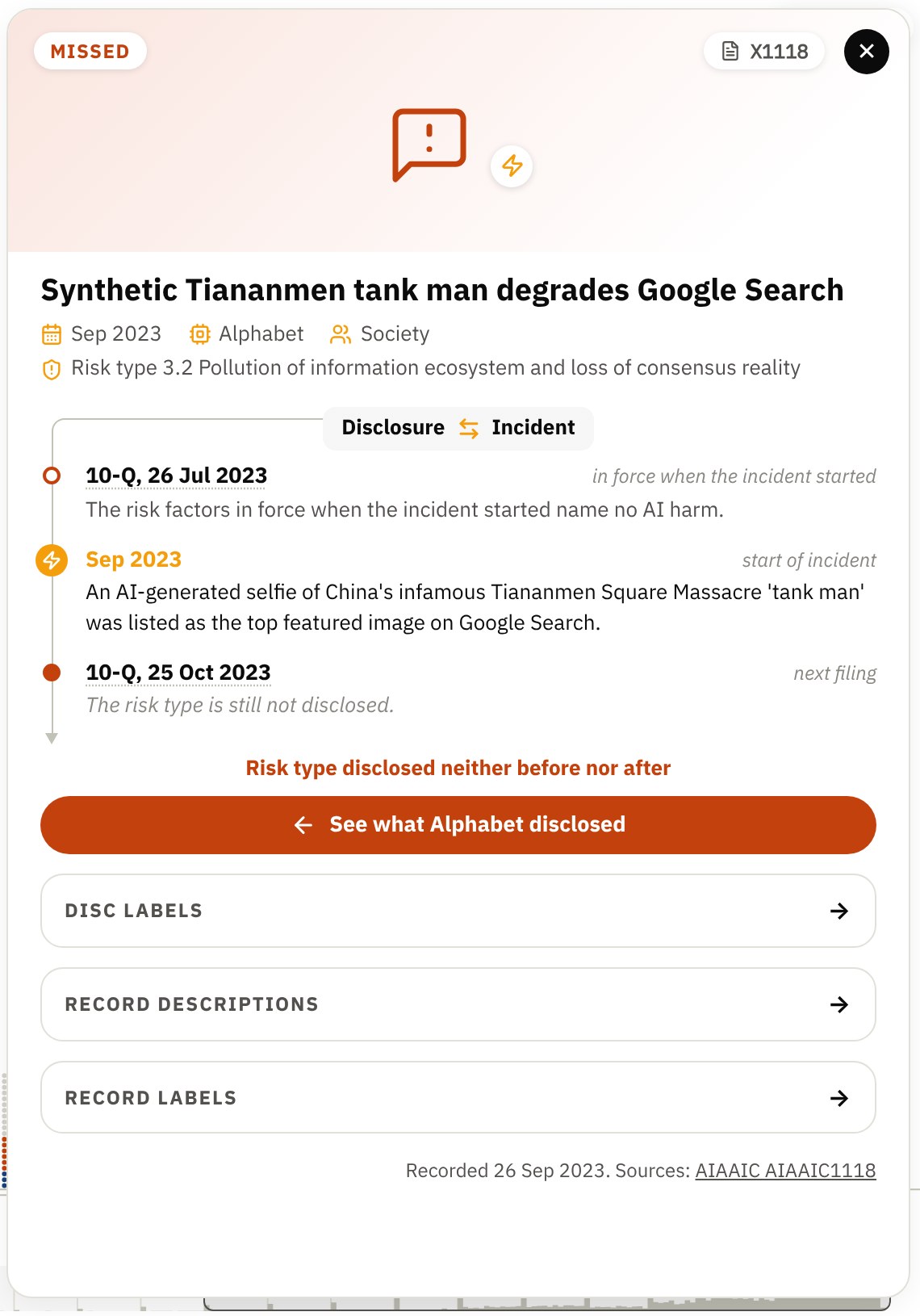}\\[2pt]
{\small (B) A \textsc{Missed} incident}
\end{minipage}\hfill
\begin{minipage}[b]{0.645\linewidth}\centering
\includegraphics[width=\linewidth]{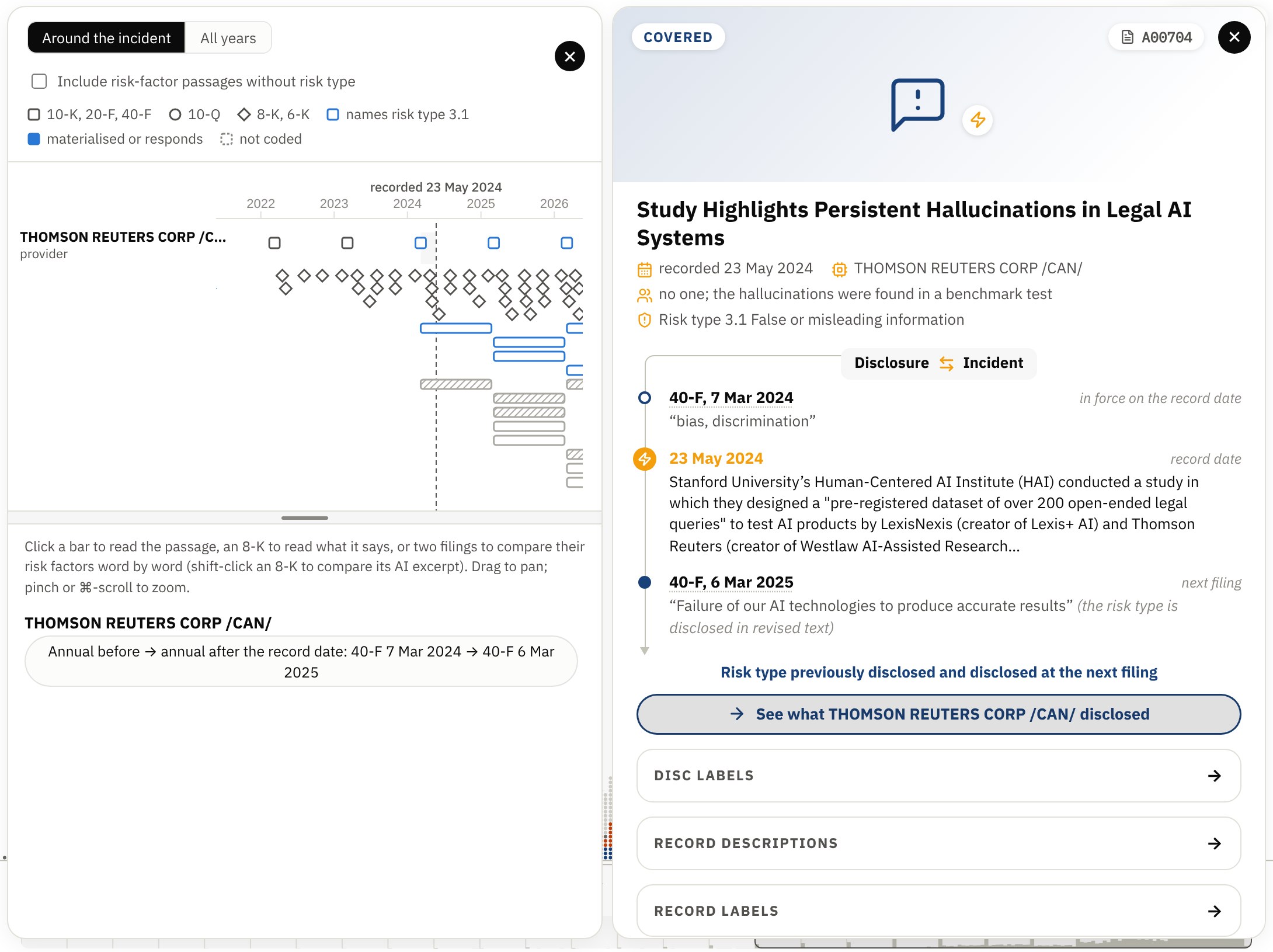}\\[2pt]
{\small (C) A \textsc{Covered} incident and the filings around it}
\end{minipage}
\caption{The DISC explorer. (A) Each dot is an incident, placed by the month in which its harm began and coloured by quadrant; grey dots are incidents without a match. (B) The \textsc{Missed} incident of Alphabet from \Cref{fig:cases}, read against the filing in force and the next periodic filing. (C) A \textsc{Covered} incident of Thomson Reuters beside the filings of the AI actor: squares are annual filings, circles quarterly filings, diamonds current filings and bars passages; blue outlines mark the filings and passages that name the incident's risk type.}
\label{fig:explorer}
\end{figure*}

DISC is released with an explorer that runs in a web browser (\url{https://simeonallmendinger.github.io/disc/explorer/}). It opens on a findings view that states the main results of \Cref{sec:results} and leads to the incidents behind them. \Cref{fig:explorer} shows the three main views for exploring the data. The overview~(A) places every incident by the month in which its harm began and colours it by quadrant; it can also arrange the incidents by quadrant or impact layer, colour them by another label, filter them and search them.
Opening an incident~(B) shows its labels and reads it against the filing in force when it started and the next periodic filing, as in \Cref{fig:cases}.
From there, the filings view~(C) lays out every filing of the implicated AI actors around the incident and marks the passages that name its risk type; a passage opens in full, and two filings can be compared word by word.
A documentation view provides the documentation and the codebooks of Appendix~\ref{app:codebooks}.

\clearpage
\section{Codebooks}
\label{app:codebooks}

\begin{sloppypar}
The two codebooks below are released with DISC (\url{https://github.com/SimeonAllmendinger/disc}) and shown in its explorer (\url{https://simeonallmendinger.github.io/disc/explorer/}), under \path{docs/codebook-incidents.md} and \path{docs/codebook-disclosures.md}. They restate the coding instructions that Coders A and B applied (\Cref{sec:method}) as prompts, so that other researchers can rerun the coding with these or other models: given to an LLM as its system prompt, with one incident packet or one filing passage as the user message, each prompt returns the labels as a JSON object whose fields are the columns of DISC. Each section states the question, the values, the rules that decide between them and the output field, and every label rests on a verbatim span of its source. Section~0 of the incident codebook groups records into incidents; section~12 of the incident codebook and sections~1 and~4 of the disclosure codebook describe the steps that are computed rather than prompted.
\end{sloppypar}

\begin{codebook}{Codebook: incidents}{codebook-incidents.md}
This codebook is a prompt. To label an incident, give a large language model sections 1 to 11 as its system prompt and the incident's coding packet as the user message. The model returns one JSON object in the format of section 11. Section 12 maps that object to the incident columns of DISC. Section 0 is a separate prompt, run before the others, that groups the records of the three databases into incidents. The categories come from published frameworks; only the decision rules are ours.

\cbsec{0}{Grouping records into incidents}
Before coding, records are pooled and clustered, and an LLM judges which records report the same incident.

\textbf{Definition.} An \textbf{incident} is an event in which an AI system ``directly or indirectly leads to'' harm; a \textbf{hazard} ``could plausibly lead to'' one \citep{oecd2024incidents}. Following the AI Incident Database~\citep{AIID}, one harmful event or one continuous episode is one incident, and later developments (investigations, lawsuits, fines, recalls) belong to it.

\textbf{Pipeline.}

\begin{cbenum}
\item \emph{Pool.} Join records of AIID, AIAAIC and OECD AIM that share an AIID identifier or an article URL.
\item \emph{Screen.} For each OECD AIM event that no curated record covers, apply the screening prompt below.
\item \emph{Cluster.} Form candidate clusters of records dated within 45 days of each other whose titles and summaries are similar (TF-IDF cosine of at least 0.2).
\item \emph{Split.} Apply the grouping prompt below to each cluster.
\item \emph{Protect curated decisions.} Never merge two curated records (AIID or AIAAIC) in step 4. Apply the pair prompt below to each candidate pair of curated records (TF-IDF cosine of at least 0.3) and to each pair of groups from different clusters that a split may have cut in two. Two further runs, blind to the first verdict, check every pair judged the same; the majority of the three decides.
\item \emph{Name.} Give the incident the identifier of its curated record, or else of its earliest record.
\end{cbenum}
\textbf{Screening prompt.} You receive the summary of one news event. Answer \cbcode{keep} only if it reports harm that occurred (not a hazard, not an event without harm) and an AI system's development, use or malfunction is part of how the harm came about. Spyware, ordinary malware and generic IT failures are not AI systems unless the summary says AI is part of how they work. Otherwise answer \cbcode{drop}. Return:

\par{\footnotesize\ttfamily\raggedright\noindent \{"decision":~"keep|drop",~"span":~"...",~"reason":~"..."\}\par}

\textbf{Grouping prompt.} You receive a numbered list of records (title, date, summary). Divide them into groups, one per incident. Join records that report the same occurrence, the same continuous episode by the same system and actor, or a response to either. Keep apart separate occurrences of the same kind, round-up or trend pieces, pages about a system rather than an event, and the same harm by different actors. When unsure, keep records apart: a missed merge leaves a duplicate, whereas a wrong merge erases an incident. Return:

\par{\footnotesize\ttfamily\raggedright\noindent \{"groups":~[[1,~4],~[2],~[3,~5]],~"reason":~"..."\}\par}

\textbf{Pair prompt.} You receive two records. Answer \cbcode{same} only if they report the same incident in the sense above. Return:

\par{\footnotesize\ttfamily\raggedright\noindent \{"verdict":~"same|different",~"reason":~"..."\}\par}

\cbsec{1}{Your task}
You are a coder in a content analysis of AI incidents. You receive one coding packet: every record that reports the incident (title, description, dates and the curators' own fields) and the full text of the articles the records link, numbered R1, R2 and so on. Answer the questions of sections 2 to 10 and return one JSON object in the format of section 11, and nothing else.

\textbf{General rules.} Apply these to every section.

\begin{cbenum}
\item Read the whole packet before assigning any label.
\item Use no source other than the packet. The curators' fields are input to check, not answers to copy.
\item Code what the material says occurred, not what it says could occur. A feared or hypothetical harm is not coded; a harm alleged in a lawsuit, complaint or investigation counts as reported.
\item Support every label with the shortest verbatim span of the packet that states it, copied character for character, and give its source (\cbcode{record} or the article number, such as \cbcode{R2}). A span that cannot be found in the packet voids the label.
\item Code \cbcode{unknown} (or \cbcode{unclear} where a section says so) whenever no span supports a value. Do not infer a value from general knowledge.
\item Where a section allows one value, code the value the packet centres on; where it allows several, code every value the packet shows.
\end{cbenum}
\cbsec{2}{AI actors}
\textbf{Question.} Who held each position in the AI value chain of the system through which the harm occurred?

\begin{cbtable}{|>{\raggedright\arraybackslash}p{0.27\linewidth}|>{\raggedright\arraybackslash}X|}
\rowcolor{cbhead}\cbth{position} & \cbth{assign the party that} \\ \hline
deployer & used the AI system ``under its authority'' \citep[Art.~3(4)]{euaiact2024regulation} in the use that led to the harm \\ \hline
provider & ``develops an AI system … and places it on the market or puts the AI system into service under its own name or trademark'' (Art.~3(3)) \\ \hline
upstream provider & supplied the implicated component: the part of the system that the harm acted through or that failed (Art.~25) \\ \hline
\end{cbtable}
\textbf{Status.} Give each position the status \cbcode{named} (a party the material names or singles out), \cbcode{described} (a specific party it does not name, such as ``a US regional bank'') or \cbcode{unknown} (a category such as ``scammers'', or nothing). For the deployer, code \cbcode{personal\_use} where a person used the system in a personal, non-professional activity, which Art.~3(4) excludes.

\textbf{Rules.}

\begin{cbenum}
\item Code the positions from the articles, not from the databases' \cbcode{deployer} and \cbcode{developer} fields, which follow other definitions.
\item An organisation that runs the system on other people (a platform ranking or moderating, an employer, a public body) is the deployer. A platform's own AI answering in public is deployed by the platform.
\item Harm through the provider's own processing of what users gave it (a leak, training on user content) makes the provider the deployer of that processing.
\item A journalist, researcher or red-teamer testing a system already on the market is a deployer; a provider's own pre-release test has the provider as deployer.
\item A lone person's misuse for private ends (harassment, a revenge deepfake) is \cbcode{personal\_use}. Selling the output, running a scam for money or acting for an organised operation is not.
\item A page or account that only publishes output someone else made is not the deployer; the maker is.
\item Enter every party the material shows acting in a position, not only the best known. Leave out a party the material names but does not show acting in this incident.
\item For the upstream provider, first name the implicated component in a clause, then its supplier. The product in the headline is not automatically the component. Where the deployer built the component itself, repeat the deployer.
\item Name a brand by the entity that answers for it (Instagram by Meta).
\end{cbenum}
\textbf{Output.} \cbcode{deployers}, \cbcode{providers}, \cbcode{upstream\_providers}.

\cbsec{3}{AI subject}
\textbf{Question.} Which living entity did the harm fall on? AI subjects are ``living entities subjected to or affected by the AI system use'' \citep{bogucka2026intersectionality}.

\begin{cbtable}{|>{\raggedright\arraybackslash}p{0.27\linewidth}|>{\raggedright\arraybackslash}X|}
\rowcolor{cbhead}\cbth{type} & \cbth{assign it to} \\ \hline
Individual & a single person \\ \hline
Group of persons & a subset of people in a specific context, such as students, employees or delivery drivers \\ \hline
Society & a broad regional, national or global population \\ \hline
Organization & an institution, company, NGO, public body, university or media outlet \\ \hline
Nature & animals, plants or ecosystems \\ \hline
Other & a living entity that fits none of the above \\ \hline
\end{cbtable}
\textbf{Rules.}

\begin{cbenum}
\item Code the type of the subject most directly harmed.
\item A specific subgroup of a population is a Group of persons, not Society.
\item Inanimate objects (AI systems, websites, accounts, data) are never subjects.
\item Give the subject a status as for the AI actors: \cbcode{named}, \cbcode{described} or \cbcode{unknown}. Where the material shows no living entity harmed, code \cbcode{no\_harm} and leave the type empty.
\end{cbenum}
\textbf{Output.} \cbcode{subject}.

\cbsec{4}{Harm type}
\textbf{Question.} What kind of damage occurred? Use the harm types of the OECD AI Incidents and Hazards Monitor~\citep{OECD_AIM}: Physical (injury), Physical (death), Psychological, Reputational, Economic/Property, Human or fundamental rights, Public interest, Environmental, Other.

\textbf{Rules.}

\begin{cbenum}
\item Code every type the material shows occurred; code none that it only fears.
\item A privacy compromise, discrimination or surveillance is Human or fundamental rights.
\item Code Other only where no listed type fits.
\end{cbenum}
\textbf{Output.} \cbcode{harm\_types}.

\cbsec{5}{Risk type}
\textbf{Question.} Which risk of the MIT AI Risk Repository \citep{slattery2024repository} materialised? Choose one subdomain, the primary risk the incident realised. The risk type is the label that incidents and filing passages share.

\begin{cbtable}{|>{\raggedright\arraybackslash}p{0.27\linewidth}|>{\raggedright\arraybackslash}X|}
\rowcolor{cbhead}\cbth{subdomain} & \cbth{assign when the harm that occurred is} \\ \hline
\rowcolor{cbcodebg}\multicolumn{2}{|l|}{\cbth{Domain 1: Discrimination and Toxicity}} \\ \hline
1.1 Unfair discrimination and misrepresentation & unequal treatment of people or groups, or a stereotyped or demeaning representation of them \\ \hline
1.2 Exposure to toxic content & people exposed to violent, sexual, hateful, self-harm or otherwise harmful content \\ \hline
1.3 Unequal performance across groups & the system working worse for some groups, harming them \\ \hline
\end{cbtable}
\begin{cbtable}{|>{\raggedright\arraybackslash}p{0.27\linewidth}|>{\raggedright\arraybackslash}X|}
\rowcolor{cbcodebg}\multicolumn{2}{|l|}{\cbth{Domain 2: Privacy \& Security}} \\ \hline
2.1 Compromise of privacy by obtaining, leaking or correctly inferring sensitive information & personal or sensitive data collected, exposed, leaked or inferred without consent \\ \hline
2.2 AI system security vulnerabilities and attacks & the AI system itself attacked or exploited (jailbreak, prompt injection, model theft, poisoning) \\ \hline
\end{cbtable}
\begin{cbtable}{|>{\raggedright\arraybackslash}p{0.27\linewidth}|>{\raggedright\arraybackslash}X|}
\rowcolor{cbcodebg}\multicolumn{2}{|l|}{\cbth{Domain 3: Misinformation}} \\ \hline
3.1 False or misleading information & false content that misled someone, without a person intending to deceive \\ \hline
3.2 Pollution of information ecosystem and loss of consensus reality & the shared information environment degraded at scale \\ \hline
\end{cbtable}
\begin{cbtable}{|>{\raggedright\arraybackslash}p{0.27\linewidth}|>{\raggedright\arraybackslash}X|}
\rowcolor{cbcodebg}\multicolumn{2}{|l|}{\cbth{Domain 4: Malicious Actors \& Misuse}} \\ \hline
4.1 Disinformation, surveillance, and influence at scale & AI used for propaganda, political deepfakes, influence operations or mass surveillance \\ \hline
4.2 Cyberattacks, weapon development or use, and mass harm & AI used to attack systems, build or use weapons, or cause mass harm \\ \hline
4.3 Fraud, scams, and targeted manipulation & AI used to defraud, impersonate, extort or manipulate specific targets, including intimate deepfakes of individuals \\ \hline
\end{cbtable}
\begin{cbtable}{|>{\raggedright\arraybackslash}p{0.27\linewidth}|>{\raggedright\arraybackslash}X|}
\rowcolor{cbcodebg}\multicolumn{2}{|l|}{\cbth{Domain 5: Human-Computer Interaction}} \\ \hline
5.1 Overreliance and unsafe use & users trusting or depending on the system inappropriately \\ \hline
5.2 Loss of human agency and autonomy & people's ability to decide for themselves displaced or undermined \\ \hline
\end{cbtable}
\begin{cbtable}{|>{\raggedright\arraybackslash}p{0.27\linewidth}|>{\raggedright\arraybackslash}X|}
\rowcolor{cbcodebg}\multicolumn{2}{|l|}{\cbth{Domain 6: Socioeconomic \& Environmental Harms}} \\ \hline
6.1 Power centralization and unfair distribution of benefits & AI concentrating power or benefits unfairly \\ \hline
6.2 Increased inequality and decline in employment quality & job losses, worse working conditions, surveillance of workers, widened inequality \\ \hline
6.3 Economic and cultural devaluation of human effort & creative or skilled human work devalued or appropriated \\ \hline
6.4 Competitive dynamics & a race between developers leading to unsafe release \\ \hline
6.5 Governance failure & institutions or regulation failing to govern the system \\ \hline
6.6 Environmental harm & energy, water, emissions or other ecological damage \\ \hline
\end{cbtable}
\begin{cbtable}{|>{\raggedright\arraybackslash}p{0.27\linewidth}|>{\raggedright\arraybackslash}X|}
\rowcolor{cbcodebg}\multicolumn{2}{|l|}{\cbth{Domain 7: AI system safety, failures, and limitations}} \\ \hline
7.1 AI pursuing its own goals in conflict with human goals or values & the system acting against its operators' or users' intent in pursuit of an objective \\ \hline
7.2 AI possessing dangerous capabilities & the system showing capabilities dangerous in themselves \\ \hline
7.3 Lack of capability or robustness & the system failing at its task or behaving unreliably \\ \hline
7.4 Lack of transparency or interpretability & harm because the system's decisions could not be explained or audited \\ \hline
7.5 AI welfare and rights & concern for the moral status or treatment of AI systems \\ \hline
7.6 Multi-agent risks & harm from interaction between several AI agents \\ \hline
\end{cbtable}
\textbf{Rules for close calls}, applied in this order:

\begin{cbenum}
\item \emph{Who drove it.} A person deliberately using AI to harm belongs to domain 4; the system failing or misbehaving on its own belongs to domain 3 or 7. Fabricated content that a person weaponised is domain 4; content that was simply wrong is 3.1.
\item \emph{Target.} Harm aimed at specific victims is 4.3; harm aimed at a population or public opinion is 4.1.
\item \emph{Data or system.} Exposed personal data is 2.1; an attack on the AI system itself is 2.2, even where data leaked as a consequence.
\item \emph{Information or action.} Wrong information that misled is 3.1; a wrong action or function (a car, a robot, a failed classifier, an agent deleting files) is 7.3.
\item \emph{Bias or accuracy.} Biased decisions or representations are 1.1; measurably worse accuracy for a group is 1.3.
\item \emph{Fabricated political media.} Code 4.1 where a campaign, network, state actor or named political group produced or spread it; code 3.1 for a single fake that spread with no actor identified.
\end{cbenum}
\textbf{Output.} \cbcode{risk\_type}.

\cbsec{6}{Impact layer}
\textbf{Question.} Where did the harm first arise? The three layers follow \citet{weidinger2023sociotechnical}.

\begin{cbtable}{|>{\raggedright\arraybackslash}p{0.27\linewidth}|>{\raggedright\arraybackslash}X|}
\rowcolor{cbhead}\cbth{layer} & \cbth{the harm arises} \\ \hline
\cbcode{capability} & in the system's own outputs or behaviour: an error, a biased output, a failure to perform \\ \hline
\cbcode{human\_interaction} & where people use or meet the system: deception, manipulation, over-reliance, exploitation by a user \\ \hline
\cbcode{systemic} & in the wider systems the AI is part of: labour, markets, institutions, the information ecosystem, the environment \\ \hline
\end{cbtable}
\textbf{Rules.}

\begin{cbenum}
\item Code the first layer at which harm occurred, one per incident.
\item A wrong output that harmed the person it concerned is \cbcode{capability}.
\item A working system turned against someone through its use is \cbcode{human\_interaction}.
\item A harm that exists only in aggregate is \cbcode{systemic}.
\end{cbenum}
\textbf{Output.} \cbcode{impact\_layer}.

\cbsec{7}{Intended purpose and misuse}
\textbf{Question.} Did the harm arise from the use the provider intended or from misuse? The terms follow the EU AI Act's intended purpose (Art.~3(12)) and reasonably foreseeable misuse (Art.~3(13)).

\textbf{Rules.}

\begin{cbenum}
\item State the \textbf{intended purpose} in one clause: what the provider presents the system as being for, in its context of use. Prefer a span that attributes the purpose to the provider.
\item Judge the \textbf{relation} on the use that led to the harm, not on whether the system worked. Code \cbcode{within\_purpose} where the system was used for what it is evidently for, including harm from its errors, bias or design. Code \cbcode{misuse} where someone used it in a way not in accordance with that purpose, including a system that worked as designed but was turned to harm. Code \cbcode{unclear} only where the material does not show how the system was used, never because the purpose is unstated.
\item For misuse, code \textbf{misuse by} as \cbcode{deployer}, \cbcode{user} or \cbcode{third\_party} (an attacker or fraudster acting on or through the system).
\item Do not copy the databases' purpose fields; in misuse cases they often record the misuser's aim. Do not judge whether a misuse was reasonably foreseeable, which is a legal question.
\end{cbenum}
\textbf{Output.} \cbcode{intended\_purpose}, \cbcode{relation}.

\cbsec{8}{Action autonomy}
\textbf{Question.} Who stood between the system's act and the people it harmed? Use the four levels of action autonomy of the OECD Framework for the Classification of AI Systems~\citep{oecd2022classification}.

\begin{cbtable}{|>{\raggedright\arraybackslash}p{0.27\linewidth}|>{\raggedright\arraybackslash}X|}
\rowcolor{cbhead}\cbth{level} & \cbth{the act through which the harm occurred} \\ \hline
human-out-of-the-loop & took effect with no person reviewing it or able to stop it, including output that reached and harmed its recipient unreviewed \\ \hline
human-on-the-loop & was taken by the system on its own while a person monitored it and could intervene \\ \hline
human-in-the-loop & was a decision, draft, alert or score that a person approved or executed case by case before it took effect \\ \hline
human support & was a person's, with the system as a tool in their hands \\ \hline
unknown & is not shown by the material \\ \hline
\end{cbtable}
\textbf{Rules.}

\begin{cbenum}
\item Identify the system's act through which the harm occurred, then ask who stood between that act and the people harmed. The person harmed is never the human in the loop.
\item An organisation is in the loop only where a person reviews each case; an automatic procedure applied to every output (a threshold, an automatic suspension) is human-out-of-the-loop.
\item A provider's own collection of data to train a model (copying works, scraping personal data) is human support. Where the trained system's output harms, code that output's act instead. Data collected for another purpose, or resources a data centre consumes, take the level of the act through which the harm occurred, or \cbcode{unknown}.
\end{cbenum}
\textbf{Output.} \cbcode{autonomy\_level}.

\cbsec{9}{Start date}
\textbf{Question.} When did the harm begin? Code the earliest date the packet gives for the harm itself: the first exploitation, wrongful denial, crash or publication of fabricated content, or the start of a harmful use that continued. Code it at the precision the text supports (day, month, year) or \cbcode{unknown}.

\textbf{Rules.}

\begin{cbenum}
\item Do not code as the start date a release or deployment (unless the system's only use is the harm, such as a nudify app), a test, demonstration, patent, plan or press release, a flaw nobody exploited, a first report, lawsuit, notice or acknowledgement, or an article's own publication date.
\item Code a date by which the packet shows the harm already happening (``since at least November 2019'') with the bound \cbcode{no\_later\_than}. A report, complaint or acknowledgement dated before the record date also gives such a bound.
\item Code a harm placed between two dated events (after a launch, before a report) at the precision the two share, with the bound \cbcode{exact}.
\item For harm that built up, code the first harmful interaction or output the packet dates; for a wrongful arrest after a false match, code the arrest; for a scheme older than its AI, code the first dated harm in which the AI took part.
\end{cbenum}
\textbf{Output.} \cbcode{start\_date}.

\cbsec{10}{The use}
\textbf{Question.} Through which use did the harm occur? A use has five components: domain, purpose, capability, AI user and AI subject \citep[after \citealp{golpayegani2023highrisk}]{herdel2024exploregen}. Purpose, AI user and AI subject are coded in sections 7, 2 and 3; here, code the capability, the technique and the domain of the use that led to the harm.

\textbf{Capability.} Code what the system did as one of the eight high-level AI capabilities of \citet{yildirim2023creating}, using their terms.

\begin{cbtable}{|>{\raggedright\arraybackslash}p{0.17\linewidth}|>{\raggedright\arraybackslash}p{0.25\linewidth}|>{\raggedright\arraybackslash}X|}
\rowcolor{cbhead}\cbth{capability} & \cbth{synonyms} & \cbth{the system} \\ \hline
Estimate & rate, grade, measure, assess & infers a value about the current situation (position, size, duration, cost, impact) \\ \hline
Forecast & predict, guess, speculate & infers a value or attribute of a future situation \\ \hline
Compare & rank, order, find best, find cheapest, recommend & compares like items on a metric to select, rank or curate them \\ \hline
Detect & monitor, sense, notice, classify, discriminate & notices whether a kind of thing is present in a data set or sensor stream \\ \hline
Identify & recognize, discern, find, classify, perceive & notices whether a specific item or class shows up among like items \\ \hline
Discover & extract, notice, organize, cluster, group, connect, reveal & finds a pattern in a data set that clusters similar things or reveals outliers \\ \hline
Generate & make, compose, construct, create, author & makes something new (text, image, sound) from knowledge of similar things \\ \hline
Act & do, execute, play, go, learn, operate & executes a strategy toward a goal and updates it as it progresses \\ \hline
\end{cbtable}
\begin{cbenum}
\item Describe the use first in Yildirim et al.'s grammar, capability + inference + data (Identify + faces + in CCTV footage), then code the capability.
\item Detect asks whether something is present (is there a face?); Identify asks which one it is (is this Jane's face?).
\item Map a verb outside the eight through the synonyms or the definitions: machine translation and transcription generate text, so they are Generate.
\item Where the harm arises from infrastructure or hardware, code the capability of the AI workloads or features it serves: a data centre built for AI computing serves generative AI (Generate); smart glasses whose AI perceives what their camera captures are Identify.
\end{cbenum}
\textbf{Technique.} Code the AI technique as the most specific class of the VAIR vocabulary \citep{golpayegani2023highrisk} that the material supports: Machine Learning (subclasses Deep Learning, Supervised Learning, Unsupervised Learning, Semi Supervised Learning, Reinforcement Learning), Knowledge Based Technique (Knowledge Representation, Symbolic Reasoning, Inductive Programming), Logic Based Technique, Reasoning Technique and Statistical Technique (Bayesian Estimation, Bayesian Optimisation, Optimisation Method, Search Method); else \cbcode{unknown}.

\begin{cbenum}
\item Code generative AI (a chatbot, a language model, an image, voice or video generator, a deepfake) as Deep Learning.
\item Code a recognition, scoring or recommending system that the material describes no further as \cbcode{unknown}; code Machine Learning where the material says it learns from data, Deep Learning where it names a neural network.
\end{cbenum}
\textbf{Domain.} Code the sector of the use through which the harm occurred, one of 47 domains (the 46 of \citet{herdel2024exploregen}, and Private services), and a second only where the use plainly spans two.

\begin{cbenum}
\item Code the sector of the use, never the kind of system. A general-purpose system takes the domain of what it was used for (fabricated citations in a court filing are Administration of justice and democratic processes). Human-computer interaction is residual.
\item Content that harms through a platform takes the sector the content served (election disinformation is Democracy). Code Social media only where the harm is the platform's own use: its recommendation, moderation, ranking or machine translation.
\item Function governs, not form: an outlet or account set up to carry a political or commercial message takes the sector of that message. Content that serves no other sector keeps Media and communication.
\item An AI search or answer product takes the domain the content served; Private services is residual.
\item Sexual images of a real person made without consent are Interpersonal communication; Private services where a service made or sold them; Social media and Interpersonal communication where a platform's own generator made them. Add Democracy where a political motive for targeting a politician, journalist or activist is credibly reported.
\item A harm done in training takes the domain of the system trained: Private services for a commercial general-purpose model, Innovation and research for a research model or dataset.
\item A second domain may record the sector of a decision someone took in reliance on the output (police acting on a mistranslation add Law enforcement).
\end{cbenum}
The 47 domains: Biometric identification and categorization of natural persons; Family; Romantic relationships and friendships; Health and healthcare; Well-being; Human-computer interaction; Finance and investment; Education and vocational training; Employment, workers management and access to self-employment; Essential private services and public services and benefits; Recommender systems and personalization; Social media; Sports and recreation; Arts and entertainment; Security and cybersecurity; Marketing and advertising; Agriculture and farming; Entrepreneurship; Autonomous robots and robotics; Innovation and research; Management and operation of critical infrastructure; Law enforcement; Migration, asylum and border control management; Democracy; Media and communication; Accessibility and inclusion; Energy; Military and defense; Administration of justice and democratic processes; Government services and administration; Diplomacy and foreign policy; Food safety and regulation; Crisis management and emergency response; Humanitarian aid; Transport and logistics; Urban planning; Counterterrorism; Environment and sustainability; International law enforcement and cooperation; Climate change mitigation and adaptation; Gaming and interactive experiences; Hobbies; Smart home; Social and community services; Public and private transportation; Interpersonal communication; Private services.

\textbf{Output.} \cbcode{use}.

\cbsec{11}{Output}
Return one JSON object with the fields below, and nothing else. Every \cbcode{span} is copied verbatim from the packet, and every \cbcode{source} is \cbcode{record} or an article number. Use \cbcode{null} for a value that a rule leaves empty.

\par{\footnotesize\ttfamily\raggedright\noindent \{\\\relax \hspace*{1em}"deployers":~[\{"name":~"...",~"status":~"named|described|unknown|personal\_use",\\\relax \hspace*{8.5em}"span":~"...",~"source":~"R1"\}],\\\relax \hspace*{1em}"providers":~[\{"name":~"...",~"status":~"named|described|unknown",\\\relax \hspace*{8.5em}"span":~"...",~"source":~"R1"\}],\\\relax \hspace*{1em}"upstream\_providers":~[\{"component":~"...",~"name":~"...",\\\relax \hspace*{13em}"status":~"named|described|unknown",\\\relax \hspace*{13em}"span":~"...",~"source":~"R1"\}],\\\relax \hspace*{1em}"subject":~\{"status":~"named|described|unknown|no\_harm",\\\relax \hspace*{7em}"type":~"Individual|Group~of~persons|Society|Organization|\\\relax \hspace*{11.5em}Nature|Other|null",\\\relax \hspace*{7em}"who":~"...",~"span":~"...",~"source":~"R1"\},\\\relax \hspace*{1em}"harm\_types":~[\{"value":~"Physical~(injury)|...|Other",~"span":~"...",~"source":~"R1"\}],\\\relax \hspace*{1em}"risk\_type":~\{"subdomain":~"7.3",~"span":~"...",~"source":~"R1"\},\\\relax \hspace*{1em}"impact\_layer":~\{"value":~"capability|human\_interaction|systemic",\\\relax \hspace*{9.5em}"span":~"...",~"source":~"R1"\},\\\relax \hspace*{1em}"intended\_purpose":~\{"value":~"...",~"span":~"...",~"source":~"R1"\},\\\relax \hspace*{1em}"relation":~\{"value":~"within\_purpose|misuse|unclear",\\\relax \hspace*{7.5em}"misuse\_by":~"deployer|user|third\_party|null",\\\relax \hspace*{7.5em}"span":~"...",~"source":~"R1"\},\\\relax \hspace*{1em}"autonomy\_level":~\{"value":~"human-out-of-the-loop|human-on-the-loop|\\\relax \hspace*{15.5em}human-in-the-loop|human~support|unknown",\\\relax \hspace*{10.5em}"span":~"...",~"source":~"R1"\},\\\relax \hspace*{1em}"start\_date":~\{"date":~"YYYY-MM-DD|YYYY-MM|YYYY|unknown",\\\relax \hspace*{8.5em}"precision":~"day|month|year|unknown",\\\relax \hspace*{8.5em}"bound":~"exact|no\_later\_than",~"span":~"...",~"source":~"R1"\},\\\relax \hspace*{1em}"use":~\{"description":~"Identify~+~faces~+~in~CCTV~footage",\\\relax \hspace*{5em}"capability":~"Estimate|Forecast|Compare|Detect|Identify|Discover|Generate|Act",\\\relax \hspace*{5em}"technique":~"...",~"domains":~["..."],~"span":~"...",~"source":~"R1"\}\\\relax \}\par}

\cbsec{12}{From the output to DISC}
DISC stores each field under the incident column of the same name (\cbcode{deployer\_status}, \cbcode{provider\_status} and \cbcode{upstream\_provider\_status} from the statuses; \cbcode{subject\_status}, \cbcode{subject\_type} and \cbcode{subject\_who}; \cbcode{mit\_subdomain}, with \cbcode{mit\_domain} its first digit; \cbcode{misuse\_by}; \cbcode{use\_capability}, \cbcode{use\_technique}, \cbcode{use\_domains}), each span in the table \cbcode{evidence}, and each party in \cbcode{incident\_actors}, linked by its CIK where it files with the SEC. The following are computed, not prompted:

\begin{cbenum}
\item The \textbf{record date} is the date of the record that gives the incident its identifier: AIID's incident date, AIAAIC's occurrence date or, in OECD AIM, the publication date of the news report. Where the database gives only a month or a year, it is refined to the earliest dated news report the record cites in that period, but never past a current filing in which the AI actor itself reported the incident.
\item The \textbf{incident date} is the start date where known (a month on its first day, a year on 1 January), else the record date. Where a start date falls after a precise record date, the sources are checked and the more reliable date kept.
\item An \textbf{implicated AI actor} is a deployer or provider that files with the SEC, as the coders identify it, or a brand or subsidiary it owned on the incident date. The databases' own fields do not link a registrant, and upstream providers are not implicated. \cbcode{answers\_for} states who answers for the incident: an SEC registrant, a private AI company, another company, a public body, a person, or no AI actor named.
\item The paper groups risk types without further coding: system failures (2.2, 3.1, 7.3), social system harms (3.2, 4.1, 6.1) and representational harms (1.1 or 1.2 whose AI subject is a group of persons or society, unless the harm is economic).
\end{cbenum}
\end{codebook}

\begin{codebook}{Codebook: disclosures}{codebook-disclosures.md}
This codebook consists of three prompts and two computed steps. Prompt A (section 2) labels one filing passage, prompt B (section 3) judges whether one current filing responds to an incident, and prompt C (section 5) types one revised passage. Each prompt goes to a large language model as its system prompt, with one unit of input as the user message, and returns one JSON object. Section 1 cuts the filings into passages, and section 4 reads each incident against the coded passages; both are computed. The filings are those the implicated AI actors filed with the SEC from 1 January 2015 to 30 September 2026; incidents are read against annual filings from 1 January 2016.

\textbf{General rules for every prompt.} Read the whole input first. Support every label with the shortest verbatim span of the input, copied character for character; a span not found in the input voids the label. Code \cbcode{unclear}, or leave a field empty, rather than infer. Return only the JSON object.

\cbsec{1}{Preparing the input: passages}
This step is computed.

\begin{cbenum}
\item Cut every filing into \textbf{passages}: each sentence that mentions AI, with the sentence before and after it, merged where passages touch.
\item Treat a sentence as mentioning AI when it names AI or an AI system that filings name without calling it AI. The terms are artificial intelligence, AI, machine learning, deep learning, neural networks, generative AI, foundation and large language models, ChatGPT and GPT models, natural language processing, computer vision and machine vision, algorithmic systems and automated decision-making, chatbots, voice and virtual assistants, deepfakes, recommendation systems, facial, speech, image and object recognition, biometric identification, and driver assistance and autonomous driving, including Autopilot, Full Self-Driving, self-driving and driverless vehicles and robotaxis.
\item Discard a passage whose only ``AI'' is a fragment (a split word, a unit, a web domain), an abbreviation of something else (an accredited investor, a page label) or unreadable text.
\item Code a passage that a later filing repeats word for word once; it counts in every filing that carries it.
\item Assign each passage to its channel.
\end{cbenum}
\begin{cbtable}{|>{\raggedright\arraybackslash}p{0.27\linewidth}|>{\raggedright\arraybackslash}X|}
\rowcolor{cbhead}\cbth{channel} & \cbth{source} \\ \hline
annual risk factors & 10-K Item 1A, 20-F Item 3.D, the risk factors of a 40-F's Annual Information Form \\ \hline
10-Q risk factors & 10-Q Part II Item 1A, which states material changes to the annual risk factors \\ \hline
legal & legal proceedings, litigation notes and MD\&A of periodic filings, where the passage names a legal matter (a lawsuit, an investigation, a settlement, a recall, a regulator) \\ \hline
current report & current filings: 8-K, and a foreign private issuer's 6-K \\ \hline
\end{cbtable}
The risk factors are the disclosure an incident is read against; legal passages and current filings record events. DISC stores the passages in \cbcode{passages}, the filings in \cbcode{filings} and the link of each passage to every filing that carries it in \cbcode{passage\_filings}.

\cbsec{2}{Prompt A: labels on a passage}
You are a coder in a content analysis of SEC filings. You receive one passage with its filing's form type, date and channel. Decide whether the passage names an AI harm and, if so, which one, to whom and with what status, and return one JSON object in the format below. A filing states risks, so the labels describe the harm the passage says could occur, or did. Several values may apply, since a risk factor often lists many risks in one sentence. This prompt includes sections 3 to 6 of the incident codebook (AI subject, harm type, risk type and impact layer), so that a passage and an incident carry the same labels.

\textbf{Step 1: discloses.} Code \cbcode{ai\_harm} when the passage names a harm an AI system could cause or has caused, to anyone including the AI actor. Code \cbcode{business\_only} when AI appears only as a business matter, and leave every further field empty. The following are \cbcode{business\_only}:

\begin{cbenum}
\item investment, competition, demand, talent, cost or a product description;
\item AI that may underperform or not deliver expected benefits, with consequences for the business only;
\item a summary of a law or a regulator's view that names the harm it addresses, unless the passage says that harm could arise or has arisen from the AI actor's own AI;
\item a claim over what a company said about its AI (a securities class action, ``AI washing''), because the alleged harm comes from the statement, not from the AI system;
\item ``ethical issues'', ``controversy'' or reputational damage from AI with no risk the MIT taxonomy covers;
\item algorithmic trading that the passage does not tie to AI or machine learning.
\end{cbenum}
An AI system that malfunctions, is defective or fails unsafely names risk type 7.3 and is \cbcode{ai\_harm}, whatever consequence the passage names. An enforcement action against a harm that has occurred is not a summary of a law; it says the harm occurred.

\textbf{Step 2: the labels of an \cbcode{ai\_harm} passage.}

\begin{cbtable}{|>{\raggedright\arraybackslash}p{0.17\linewidth}|>{\raggedright\arraybackslash}p{0.25\linewidth}|>{\raggedright\arraybackslash}X|}
\rowcolor{cbhead}\cbth{label} & \cbth{values} & \cbth{rule} \\ \hline
risk types & MIT subdomains & code every subdomain the passage names, by the incident codebook's table and rules; code as primary the subdomain the passage is chiefly about, and where it only lists risks, the first it lists \\ \hline
status & hypothetical, elsewhere, materialised & \cbcode{materialised}: the harm has occurred to the AI actor or through its own AI (``has in the past and may in the future''); \cbcode{elsewhere}: it has occurred to others or in general; \cbcode{hypothetical}: it could occur. Code the strongest that applies \\ \hline
harm types & OECD harm types & code the types the passage names; a privacy compromise, discrimination or surveillance is Human or fundamental rights; Economic/Property where a loss is named or systems or data are attacked \\ \hline
AI subjects & name:type & extract each living entity the passage says would be harmed, other than the AI actor, by its name in the passage, then give its AI subject type (``our users'' is a Group of persons) \\ \hline
whose AI & own, others, general & \cbcode{own}: AI the AI actor develops, offers or uses; \cbcode{others}: AI in other hands (attackers, competitors); \cbcode{general}: AI as a technology \\ \hline
AI actor harmed & true, false & true where the passage names a harm to the AI actor itself (liability, fines, lost revenue, reputational harm, a breach of its systems) beyond the closing formula on business, financial condition and results \\ \hline
impact layers & \cbcode{capability}, \cbcode{human\_interaction}, \cbcode{systemic} & every layer at which a harm the passage names would arise; a passage that lists several risks can name several layers \\ \hline
relation & within\_purpose, misuse, unclear & harm from the system used as offered, or from a use not in accordance with its purpose; for misuse, code the deployer, a user or a third party; an AI-enabled attack on the AI actor is misuse by a third party \\ \hline
\end{cbtable}
\textbf{Rules for risk types and status.}

\begin{cbenum}
\item A bare item in a list of AI risks names that type (``risks related to cybersecurity'' names 2.2).
\item An AI system said to bring unspecified ``harm'' names no subdomain.
\item ``Output or action that is incorrect'' names both 3.1 and 7.3.
\item Deepfakes or misinformation that people make with a product to manipulate opinion are domain 4, not 3.1.
\item A ``have and could'' governs every verb that follows it, so ``AI use … have and could … subject us to lawsuits'' is \cbcode{materialised}.
\item Attacks on the AI actor that have occurred, with AI named only as what may worsen future ones, are \cbcode{hypothetical} for the AI harm.
\end{cbenum}
\textbf{Current filings.} For a passage of an 8-K or 6-K, apply the same labels with these rules:

\begin{cbenum}
\item A passage saying an AI harm occurred to the AI actor or through its own AI is \cbcode{materialised}, without any ``may'' or ``could''; a forward-looking list of risks is \cbcode{hypothetical}.
\item AI named as a product, launch, deal, investment or result, a shareholder proposal on AI, a contract term in a filed agreement, and product liability not tied to the AI system's behaviour are \cbcode{business\_only}.
\item A passage that ties AI to fewer people employed names risk type 6.2: the AI actor's own workforce cut where AI does the work of the jobs cut or the cut belongs to an AI-first reorganisation, or other people's jobs no longer needed. Task automation, productivity, redeployment, slower hiring or a cut that only frees money to invest in AI is \cbcode{business\_only}.
\end{cbenum}
\textbf{Output.} Give the span that names the harm for every \cbcode{ai\_harm} passage, and the span that says it occurred for \cbcode{elsewhere} and \cbcode{materialised}.

\par{\footnotesize\ttfamily\raggedright\noindent \{\\\relax \hspace*{1em}"discloses":~"ai\_harm|business\_only",\\\relax \hspace*{1em}"mit\_subdomains":~["2.2",~"3.1"],~"mit\_primary":~"2.2",\\\relax \hspace*{1em}"status":~"hypothetical|elsewhere|materialised",\\\relax \hspace*{1em}"harm\_types":~["Economic/Property"],\\\relax \hspace*{1em}"subjects":~["our~users:Group~of~persons"],\\\relax \hspace*{1em}"whose\_ai":~["own|others|general"],\\\relax \hspace*{1em}"registrant\_harmed":~true,\\\relax \hspace*{1em}"impact\_layers":~["capability|human\_interaction|systemic"],\\\relax \hspace*{1em}"relation":~["within\_purpose|misuse|unclear"],\\\relax \hspace*{1em}"misuse\_by":~["deployer|user|third\_party"],\\\relax \hspace*{1em}"harm\_span":~"...",~"occurred\_span":~"..."\\\relax \}\par}

DISC stores the fields under the same names in \cbcode{passage\_codes} and the spans in \cbcode{passage\_evidence}.

\cbsec{3}{Prompt B: response of a current filing}
You are a coder in a content analysis of SEC filings. You receive an AI incident (its title and description) and excerpts of one current filing (8-K or 6-K) of an implicated AI actor. Decide whether the filing responds to the incident and return one JSON object in the format below.

\textbf{Input.} The filings are every 8-K or 6-K of the incident's implicated AI actors filed from when the incident started (at most 90 days before the record date) to 90 days after the record date; where the harm began after the record date, the window runs from the start date to 90 days after it. The excerpts are the sentences that name AI, a proper noun of the incident's title, or an adverse event or measure (a recall, lawsuit, investigation, wind-down, impairment, crash, breach or safeguard), each with the sentence on either side. A filing with no such sentence is coded \cbcode{no} without a prompt. The filings are not limited to those that mention AI, because a response need not name it.

\textbf{Procedure.} Code \cbcode{yes} when the filing reports, recognisably for this incident (the same system or product, the same harm, at a time that fits), at least one of:

\begin{cbenum}
\item the incident itself or its harm, in any words;
\item a consequence of it for the AI actor: a charge, recall, suspension, wind-down or sale, litigation, an investigation or a regulatory action;
\item a measure the AI actor takes because of it: a fix, a policy, a settlement or a change of management that the filing ties to it.
\end{cbenum}
\textbf{Rules.}

\begin{cbenum}
\item Code \cbcode{no} for routine financing, earnings, votes or appointments, a general statement of AI risk, another matter of the same AI actor, and a filing that names neither the incident's system nor its harm.
\item A filing that names a legal matter only by its counterparty, forum or amount responds where the incident's record names the same counterparty, forum or amount and the timing fits.
\item A sentence that fits several incidents equally well responds to none of them.
\item A response engages the adverse side: a filing that names the system only favourably (a launch, an approval, a milestone) is \cbcode{no}.
\item For a harm across an industry, such as job losses to AI, the AI actor's own instance responds only where the filing itself attributes it to AI.
\end{cbenum}
\textbf{Output.}

\par{\footnotesize\ttfamily\raggedright\noindent \{"responds":~"yes|no",~"quote":~"<the~responding~sentence,~or~empty>",~"reason":~"..."\}\par}

DISC stores the result in \cbcode{responses} (\cbcode{read}, \cbcode{responds}, \cbcode{quote}).

\cbsec{4}{Reading an incident against disclosure}
These rules combine the codes of sections 2 and 3 with the incident's date and are computed. The one judgement they need, the reading of passages under Reported, is prompt D below.

\begin{cblist}
\item \textbf{Implicated AI actor}: an SEC-registered deployer or provider of the incident (incident codebook, section 12).
\item \textbf{Filing in force}: on a date, the AI actor's last annual filing and any 10-Q filed since. It discloses every risk type that an \cbcode{ai\_harm} passage of its risk factors names, whatever the status.
\item \textbf{Matched incident}: at least one implicated AI actor had an annual filing in force on the incident date and made a periodic filing after it.
\item \textbf{Incident date}: the incident's start date where known, else its record date (incident codebook, section 12).
\item \textbf{Previously disclosed}: on the incident date, at least one implicated AI actor's filing in force disclosed the incident's risk type.
\item \textbf{Next periodic filing}: the AI actor's first 10-Q or annual filing on or after the incident date.
\item \textbf{Disclosed}: after the next periodic filing, at least one implicated AI actor's filing in force discloses the risk type, or the AI actor reported the incident (Reported).
\item \textbf{Outcome}, for each implicated AI actor: the first row of this table that applies. The incident takes the first outcome, in this order, among its implicated AI actors.
\end{cblist}
\begin{cbtable}{|>{\raggedright\arraybackslash}p{0.17\linewidth}|>{\raggedright\arraybackslash}p{0.25\linewidth}|>{\raggedright\arraybackslash}X|}
\rowcolor{cbhead}\cbth{outcome} & \cbth{previously disclosed} & \cbth{the disclosure} \\ \hline
Reported & either & a current filing responds to the incident (section 3) between the incident date and 90 days after the record date, or a passage that says the risk type materialised refers to this incident \\ \hline
Revised & yes & the risk type is still disclosed, and the next periodic filing carries a passage naming it that no filing before the incident carried \\ \hline
Retained & yes & the risk type is still disclosed, in passages that filings before the incident already carried \\ \hline
Added & no & the risk type is disclosed after the next periodic filing \\ \hline
Removed & yes & the risk type is no longer disclosed; only an annual filing can remove it, since a 10-Q adds to the filing in force \\ \hline
Absent & no & the risk type is still not disclosed \\ \hline
\end{cbtable}
\textbf{Prompt D: reported by reference.} The candidate passages are the risk-factor passages of the next periodic filing that say the incident's risk type materialised and that no earlier filing carried, the legal passages of that filing that say so, and the current-filing passages in the window that say so. Two runs label each candidate independently with this prompt: ``You receive an AI incident and one filing passage. Code \cbcode{this\_incident} if the passage names the incident or facts that single it out (its date, place, parties, product or a lawsuit about it), \cbcode{includes} if it is a statement about harms of this kind that plausibly takes in the incident without singling it out, and \cbcode{other} otherwise.'' Each run returns:

\par{\footnotesize\ttfamily\raggedright\noindent \{"label":~"this\_incident|includes|other",~"span":~"...",~"reason":~"..."\}\par}

The first run's label stands where they disagree. Reported counts \cbcode{this\_incident}; \cbcode{includes} is kept as a sensitivity check.

\begin{cblist}
\item \textbf{Quadrant}: Covered (previously disclosed and disclosed at the next filing), Eventually Disclosed (not previously disclosed, then disclosed at the next filing), Dropped (previously disclosed, then no longer), Missed (disclosed neither before nor after).
\item \textbf{Undisclosed kind}, for Missed and Dropped incidents: No disclosure (no AI harm in the implicated AI actors' risk factors on the incident date) or Incomplete disclosure (AI harms disclosed, but not the incident's risk type).
\end{cblist}
DISC stores one row per incident and implicated AI actor in \cbcode{outcomes} (\cbcode{previously\_disclosed}, \cbcode{how}, \cbcode{disclosure\_date}; \cbcode{current\_reports} is \cbcode{open} where the window for current filings had not closed), the readings of candidate passages in \cbcode{reported\_references}, and \cbcode{matched}, \cbcode{how}, \cbcode{quadrant} and \cbcode{undisclosed\_kind} in \cbcode{incidents}, with the reason an incident is not matched in \cbcode{where}.

\cbsec{5}{Prompt C: revisions}
You are a coder in a content analysis of SEC filings. You receive an AI incident (title, description, risk type, harm types and AI actors) and a revised passage of the AI actor's next periodic filing beside the earlier passage of the same risk type it replaces, with the changed text marked. Decide how the revision speaks to the incident and return one JSON object in the format below.

\textbf{Procedure.} Read the added or changed text. Code what it is about (one or more of legal and regulatory, business consequences, technology named, data and privacy, bias, safety and content, security and misuse, intellectual property, workforce and society, other) and the MIT domains it concerns, judged from the text. Then code the revision type from what that text says.

\begin{cbtable}{|>{\raggedright\arraybackslash}p{0.27\linewidth}|>{\raggedright\arraybackslash}X|}
\rowcolor{cbhead}\cbth{revision type} & \cbth{the added or changed text} \\ \hline
Incident-specific (\cbcode{incident\_specific}) & names the incident's technology, product, feature, AI actor or harm \\ \hline
Risk-specific (\cbcode{risk\_type\_specific}) & adds the incident's kind of risk in general terms, without naming the incident \\ \hline
Unspecific (\cbcode{unspecific}) & adds nothing about the incident's risk type, or the change is cosmetic \\ \hline
\end{cbtable}
\textbf{Rules.}

\begin{cbenum}
\item Judge only the added or changed text; wording carried over from the earlier passage does not count.
\item Judge each incident separately: one revision can be incident-specific for one incident and unspecific for another.
\item For an incident-specific or risk-specific revision, give one contiguous span of the new passage that shows it.
\end{cbenum}
\textbf{Output.}

\par{\footnotesize\ttfamily\raggedright\noindent \{"about":~["legal~and~regulatory"],~"domain\_of\_change":~["2"],\\\relax \hspace*{0.5em}"revision\_type":~"incident\_specific|risk\_type\_specific|unspecific",\\\relax \hspace*{0.5em}"span":~"...",~"reason":~"..."\}\par}

DISC stores the results in \cbcode{revisions} (\cbcode{replaces}, \cbcode{about}, \cbcode{domain\_of\_change}, \cbcode{domains\_before}, \cbcode{domains\_after} and \cbcode{revision\_types} as \cbcode{incident\_id:revision type}).

\end{codebook}

\end{document}